\pdfoutput=1
\documentclass[runningheads]{llncs}

\usepackage{eccv}

\usepackage{eccvabbrv}
\usepackage{multirow}
\usepackage{adjustbox}
\usepackage{booktabs}
\usepackage{algorithm}
\usepackage[table]{xcolor}
\usepackage{colortbl}
\definecolor{emerald}{rgb}{0.31, 0.78, 0.47}
\definecolor{hansayellow}{rgb}{0.91, 0.84, 0.42}
\newcommand{\namesec}{Sec.}
\newcommand{\namefig}{Fig.}
\newcommand{\nametab}{Table}
\newcommand{\namemethod}{\textit{HALO}} 

\newcommand{\equalcontrib}{\textsuperscript{*}}

\newcommand{\namealg}{Algorithm}

\usepackage{graphicx}
\usepackage{booktabs}
\usepackage{wrapfig}

\usepackage{algpseudocode}
\usepackage{amsmath}
\usepackage{setspace}

\usepackage[accsupp]{axessibility}  % Improves PDF readability for those with disabilities.

\usepackage{hyperref}
\usepackage{orcidlink}

\begin{document}

\title{Controlling Motion Transfer in Diffusion Transformers via Attention Heads} 

\titlerunning{Controlling Motion Transfer}


\author{Sunyoung Jung\inst{1}\equalcontrib \orcidlink{0009-0004-6915-2675} \and
Jiwoo Park\inst{1,2}\equalcontrib \orcidlink{0009-0005-0538-5725} \and
Yoonseok Choi\inst{1}\orcidlink{0009-0004-4807-1884} \and
Kyobin Choo\inst{1}\orcidlink{0000-0003-2856-402X} \and
Ming-Hsuan Yang\inst{3}\orcidlink{0000-0003-4848-2304} \and
Seong Jae Hwang\inst{1}\textsuperscript{\dag}\orcidlink{0000-0002-3713-5553}
}

\authorrunning{S. Jung et al.}

\institute{$^{1}$Yonsei University \quad $^{2}$LG Electronics
 \quad $^{3}$University of California, Merced\\
% \email{\{sunyoungj,wldn1677,yoonseokchoi,chu,seongjae\}@yonsei.ac.kr \\ 
% mhyang@ucmerced.edu} \\
Project page: \href{https://sunyj-hxppy.github.io/halo/}{\texttt{https://sunyj-hxppy.github.io/halo}}
}

% Save the original TOC-writing command.
% \let\savedaddcontentsline\addcontentsline

% % Disable TOC entries for the main paper.
% \renewcommand{\addcontentsline}[3]{}

\maketitle

\begingroup
\renewcommand{\thefootnote}{}
\footnotetext{
\hspace{-1.8em}\textsuperscript{*} Equal contribution.
\quad
\textsuperscript{\textdagger} Corresponding author.
}
\endgroup

\begin{center}
    \centering
    \captionsetup{type=figure}
    \includegraphics[width=\linewidth]{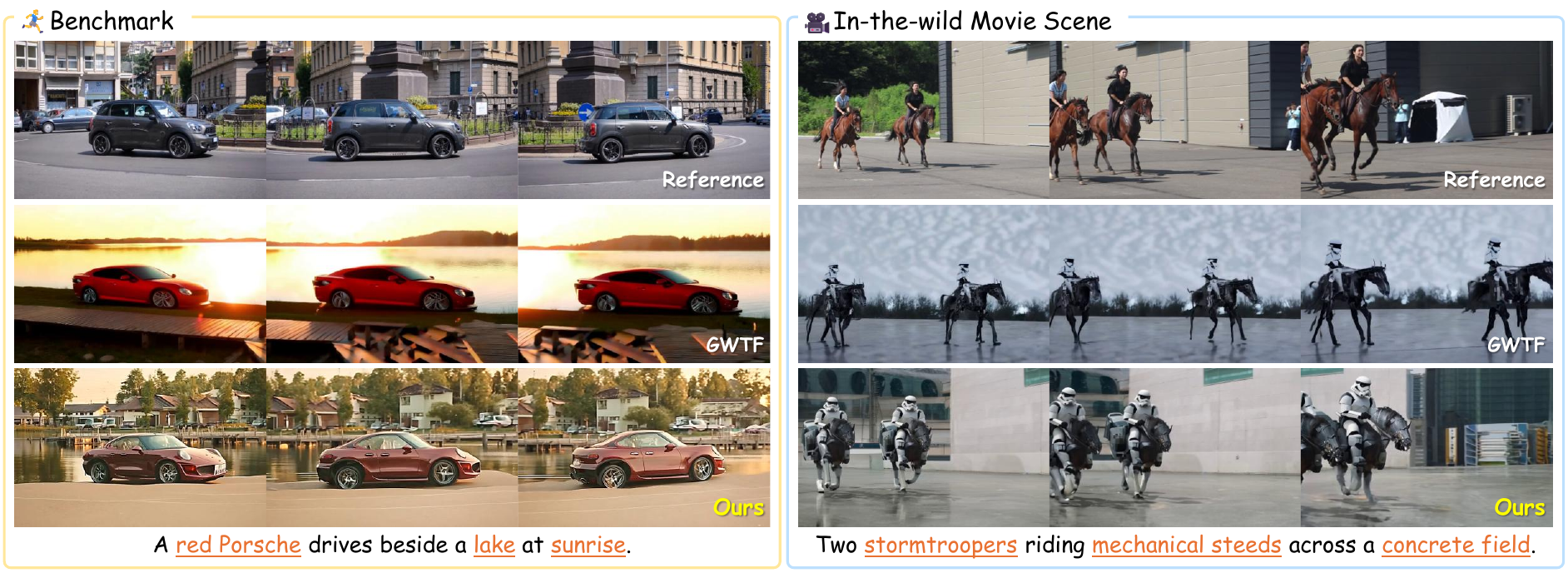}
    \vspace{-6mm}
    \caption{Overview.
    We present \namemethod{}, a head-aware controllable motion transfer framework for video Diffusion Transformers, which identifies motion- and structure-specialized attention heads within the model. Leveraging these findings, \namemethod{} generates videos that follow the target prompt while remaining motion- and structure-aligned with reference videos, achieving accurate motion transfer.
    % \namemethod{} generates videos that faithfully reflect the target prompt while remaining both motion- and structure-aligned with reference videos, achieving precise and coherent motion transfer. 
    }
    \label{fig:fig1}
    \vspace{-2mm}
\end{center}

\begin{abstract}
%\vskip -1mm
Diffusion Transformers (DiTs) have advanced video generation with high-quality, temporally coherent results. 
However, extending them to motion transfer, which requires following reference motion while aligning with a target prompt, remains challenging due to limited understanding of motion and structure representations within DiTs. 
We analyze video DiTs at the attention-head level and identify distinct heads specialized for motion and spatial structure. 
Based on this insight, we propose a head-aware controllable motion transfer framework that requires no parameter updates. 
Our method refines motion cues from motion-specialized heads via semantic correspondence guidance and preserves structure through selective feature injection. 
% This head-level control enables accurate motion transfer and provides a foundation for controllable video generation with DiTs. 
This head-level control not only enables accurate motion transfer but also provides an interpretable foundation for controllable video generation with DiTs.
% Code will be publicly released.
\vspace{-4mm}

% This new understanding reveals the inner mechanisms of video DiTs, enabling accurate motion transfer and establishing a foundation for controllable video generation. 
\end{abstract}
% Among their applications, motion transfer remains a fundamental challenge, as achieving faithful motion and consistent structure requires a precise understanding of how video DiTs internally encode motion and spatial cues.
% In this work, we aim to deepen the understanding of the specialized roles within DiTs to enable more effective motion transfer.
% We are the first to analyze video DiTs at the attention-head level, revealing distinct subsets of heads that specialize in modeling motion and structure.
% Our analysis reveals that distinct subsets of heads inherently specialize in modeling motion and structure.
% This head-level understanding provides a principled foundation for controllable and semantically consistent motion transfer with diffusion transformers.
% Motion transfer, which generates a video that follows the motion of a reference while aligning with the semantics of a target prompt, has shown significant progress with Diffusion Transformers (DiTs). However, the limited understanding of how motion and structure are represented within DiTs makes motion transfer challenging to ensure motion consistency and semantic coherence aligned manner.
% This understanding has been hindered by the complexity of their multi-head attention mechanisms.     
\vspace{-8pt}
\vspace{-4mm}
\section{Introduction}
\label{sec:intro}
Motion transfer in video generation synthesizes a video that follows the motion of a reference video while adhering to a target prompt. The primary objectives are (1) motion fidelity, ensuring temporal adherence to the reference motion, and (2) structural alignment, maintaining spatial layout of the reference~\cite{smm,ropecraft}. Achieving these goals requires modeling of spatio-temporal dependencies, an area in which recent video Diffusion Transformers (DiTs)~\cite{cogvideox,wan,hunyuanvideo,xing2024make} have shown strong capability. Given their ability to capture spatial structure and temporal dynamics, DiTs have become a natural choice for motion transfer~\cite{ditflow, ropecraft, gowithflow}.
 
Existing DiT-based motion transfer approaches, such as noise warping~\cite{gowithflow}, rotary positional embedding manipulation~\cite{ropecraft}, and cross-frame attention optimization~\cite{ditflow}, offer varying degrees of motion controllability.
However, these methods focus on manipulating motion representations without an understanding of how motion and structure are encoded within DiTs.
This lack of understanding is due to the distributed functionality of DiTs, which makes their internal mechanisms challenging to analyze~\cite{ditctrl, stableflow, scaling}. 
Consequently, generated videos often exhibit seemingly plausible motion yet inaccurate trajectories or misaligned object structures relative to the reference video.
As illustrated in \namefig{}~\ref{fig:fig1}, Go-with-the-flow (GWTF)~\cite{gowithflow}, a state-of-the-art method, exhibits motion deviations, including a car moving straight instead of turning and two stormtroopers with misaligned positions compared to the reference.
% As a result, the internal mechanisms by which DiTs capture motion dynamics and spatial structure remain underexplored.

Therefore, we conduct a detailed analysis of video DiTs, focusing on the attention heads to answer the fundamental question:
\textit{How are motion and structural cues internally encoded in video DiTs?}
To identify the heads specialized in modeling motion, we introduce the first head-level analysis based on displacement maps. 
The displacement map~\cite{ditflow} encodes motion as patch-wise coordinate differences between frames, making it effective for analyzing motion properties of heads. 
% Using cross-frame attention~\cite{ditflow,difftrack}, we compute patch coordinates between frames, referred to as displacements, which encode motion, as shown in \namefig{}~\ref{fig:fig2}. 
% Using cross-frame attention, which captures frame-wise correlations~\cite{ditflow,difftrack}, we compute inter-frame displacements representing motion patterns, as shown in \namefig{}~\ref{fig:fig2}.
% We then analyze these displacement maps under three head configurations—temporal, spatial, and all—to identify heads specialized in modeling motion.
For structural cues, we compute the visual token attention-map entropy, which represents the uncertainty in patch dependency. 
By utilizing the entropy, we select heads that reliably capture structural information.
% visual token attention map이 패치들간의 uncertainty를 나타내기 떄문에 이를 활용해서 map의 entropy로 structure 담당하는 head를 찾는다. 

% By investigating attention patterns of each head, we uncover distinct subsets of heads exhibiting specialized roles for motion flow and structure.

Our analysis identifies two distinct attention head subsets within video DiTs:
(1) \textit{Motion-Specific Heads.} 
% The attention maps of particular heads exhibit strong visual token correspondences across frames, enabling the capture of temporally coherent motion cues on the displacement maps.
A subset of heads exhibits strong patch correspondences across frames, capturing temporally coherent motion cues in the displacement maps.
These heads present clear displacement maps that accurately reflect the motion, demonstrating their superior capability to represent coherent motion flow.
% Compared to other heads, motion-specific heads demonstrate superior capability in representing coherent motion flow.
% Therefore, their attention maps clearly reveal inter-frame motion patterns that reflect consistent motion trajectories.
%, whereas the others predominantly encode static context and appearance.
(2) \textit{Structure-Specialized Heads.} Another subset of heads encodes structural information.
The attention maps of the heads show low entropy, characterized by sharply diagonal patterns. 
% Their visual token attention maps exhibit low-entropy, sharply diagonal patterns, indicating stable patch-to-patch relationships.
This attention pattern yields attention features with concentrated structural information, confirming that attention map entropy serves as an indicator of the structural content. 
% This indicates that those heads capture stable patch-to-patch attention across frames, therby encoding structural information. 

% These patterns correlate with the feature space, where structural information is concentrated, confirming that these heads capture spatial information.
% Visualization of their attention maps reveals periodic diagonal patterns with low attention-map entropy, indicating stable spatial patch-to-patch relationships.

Building on these findings, we propose \namemethod{}, a head-driven semantic-structu
\noindent
ral motion transfer framework that simultaneously enforces motion fidelity and structural alignment.
By leveraging the properties of attention heads in video DiTs, our approach first constructs inter-frame displacement maps using only motion-specific heads to guide motion optimization.
These displacement maps capture the motion dynamics but lack semantic information. 
\begin{figure*}[!t] % r: 우측 배치, 0.5\textwidth: 본문 너비의 절반 차지
  \begin{center}
    \vspace{-2pt} % 상단 여백 조절 (필요시)
    \includegraphics[width=\textwidth]{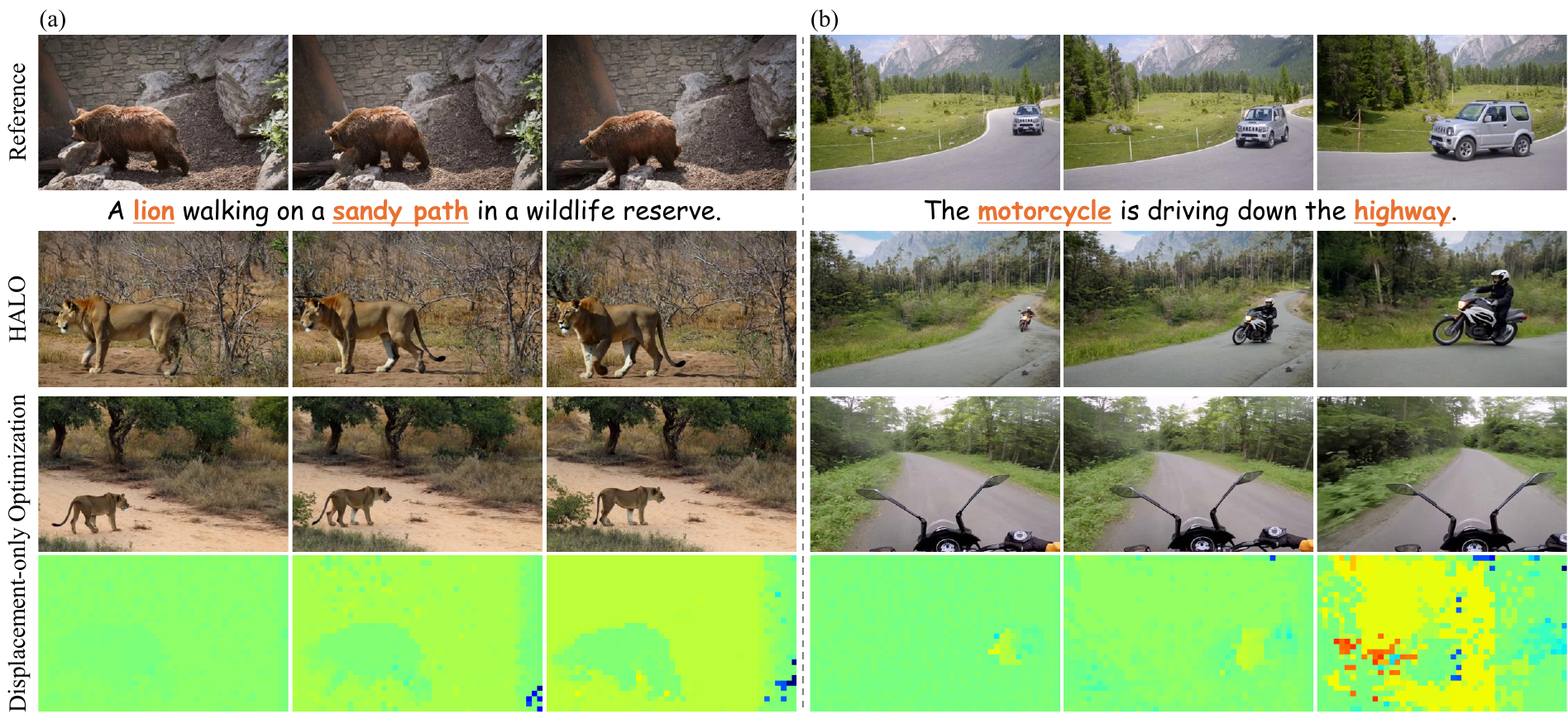}
    \vspace{-10pt} % 이미지와 캡션 사이 여백
    \caption{Limitations of displacement-based motion transfer. Comparison between \namemethod{} and displacement-only optimization~\cite{ditflow}. (a) Lack of semantic alignment causes motion errors. (b) Missing structural preservation leads to spatial misalignment. \namemethod{} ensures consistent motion and spatial fidelity.}
    \label{fig:fig2}
    \vspace{-30pt} % 하단 여백 조절
  \end{center}
\end{figure*}
This often leads to transferring motion into semantically irrelevant regions, causing inconsistent or reversed object motion, as shown in \namefig{}~\ref{fig:fig2}(a).
To overcome this issue, we introduce semantic guidance modules that align motion cues with semantic similarities derived from diffusion features, known to encode rich object-level semantics~\cite{cove,dift}.
This refinement produces semantically aligned displacement maps, enabling temporally consistent and semantically coherent motion transfer. 

While the refined displacement map provides motion flow information, it remains insufficient for preserving the spatial layout of the reference video.
Consequently, the generated results often exhibit structural misalignment with the reference, as illustrated in \namefig{}~\ref{fig:fig2}(b).
To overcome this limitation, we leverage our second insight that low-entropy attention heads primarily encode structural information.
Specifically, we introduce selective structural feature injection by incorporating attention features from the reference video through these low-entropy heads.
This head selection reinforces structural information while suppressing noisy and diffuse cues, yielding motion transfer that is structurally aligned with the reference.

% \begin{figure}[t!]
%     \centering
%     \includegraphics[width=\columnwidth]{Figure/fig2_final.jpg}
%     \caption{Limitations of displacement-based motion transfer.
%     Comparison between \namemethod{} and displacement-only optimization~\cite{ditflow}, along with the corresponding displacement maps. 
%     (a) Lack of semantic alignment causes motion errors (e.g., the lion walks backward). 
%     (b) Missing structural preservation leads to the object becoming spatially misaligned (e.g., the motorcycle). Conversely, \namemethod{} ensures consistent motion and spatial fidelity.
%     }
%     \label{fig:fig2}
%     \vspace{-5pt}
% \end{figure}

The main contributions of this work are:
\begin{itemize}
    \vspace{-10pt}
    \item We present a \textbf{head-level functional analysis} of video DiTs, revealing motion-specific and structure-specialized attention heads and validating them as \textbf{control primitives} for controllable video generation.
    \item We propose \textbf{\namemethod{}}, a \textit{head-aware controllable motion transfer} framework that enhances motion fidelity through \textbf{semantic-aware displacement optimization} and preserves spatial consistency via \textbf{selective feature injection} from structurally informative heads.
    \item We perform extensive evaluations on standard motion transfer benchmarks and our Movie Scene Dataset, demonstrating that it achieves \textbf{better motion coherence and structural alignment}.
    % \item We perform evaluations on a motion transfer benchmark, demonstrating that our method consistently achieves \textbf{better motion coherence and structural alignment}.
    % For more applicability, we validate the results on a movie scene dataset.
\end{itemize}

\section{Related Work}
%\subsection{Text-to-Video Generation}
\noindent \textbf{Text-to-Video Generation.}
Early Text-to-Video (T2V) generation was driven by U-Net-based diffusion models~\cite{videocrafter2,sora,zeroscope}, which extended image diffusion with temporal attention or 3D convolutions to capture spatio-temporal dynamics~\cite{lumiere,animatediff,svd}. 
More recently, DiTs have emerged as the dominant paradigm~\cite{wan,cogvideox,ltx,hunyuanvideo,gentron}, significantly improving temporal consistency and visual quality. 
Motivated by these advances, we build our framework upon a video DiT backbone.

%\subsection{Motion Transfer}  
\vspace{1mm}
\noindent \textbf{Motion Transfer.}
Motion transfer aims to generate a target video by extracting and reflecting only the motion information from a reference video~\cite{transfer1, smm, conmo, sma}. The main challenge of this task lies in effectively decoupling the motion and appearance information from the reference~\cite{moft,vmc}. 

Early studies have addressed this challenge by extracting motion embedding from the temporal modules of U-Net~\cite{motionclone, motioninv, motiondirector}. 
Recently, the strong generative capability of DiTs has motivated their adoption in the motion transfer field.
For instance, GWTF~\cite{gowithflow} explicitly warps noise to manipulate the motion, while RoPECraft~\cite{ropecraft} handles motion by warping the RoPE embeddings using optical flow. Furthermore, DiTFlow~\cite{ditflow} derives displacement from cross-frame attention to guide patch movement. 
While these prior works have explored various ways to handle motion information in DiTs, there has been a lack of direct analysis of how motion and structure are encoded within DiTs' internal representations.
% they lack a direct analysis of how motion and structure are encoded within DiTs' internal representations.

% To bridge this analytical gap, our work analyzes DiT internal representations to disentangle and extract features that purely represent motion and structure for motion fidelity. 
% Using these extracted features, we perform training-free optimization for motion transfer without updating the model parameters.
To bridge this analytical gap, our work analyzes DiT internal representations to identify motion-specific and structure-specialized heads. 
By leveraging these heads, we achieve training-free motion transfer and extend our approach to controllable video generation.
% Unlike DiTFlow, which derives displacement from raw cross-frame attention, HALO builds semantic-aware motion representations from motion-specific heads and complements them with selective structural feature injection.
%leverage the motion information implicitly embedded within the DiT attention. 
% We analyze that significant motion information is already encoded in the DiT's specific attention heads and proposes a motion transfer framework by utilizing this insight compared to existing works.

%\subsection{Attention in Video DiTs}
\vspace{1mm}
\noindent \textbf{Attention in Video DiTs.}
U-Net-based video diffusion models~\cite{videocrafter2,zeroscope} explicitly separate spatial and temporal attention, enabling motion extraction from temporal modules~\cite{vmc,motionclone}. 
In contrast, DiTs~\cite{cogvideox,wan} adopt unified attention that jointly models spatial and temporal information. 
While this design enhances expressiveness, it obscures the distinction between motion and structure, complicating motion-specific control~\cite{ditflow,understanding}. 
In contrast, we posit that this motion information is independently encoded within the unified structure of attention in the DiTs.
Consequently, we focus on analyzing this unified attention structure at the level of individual attention heads to identify the distinct encoding of motion and structure in terms of motion transfer.

\label{sec:related}
\section{Analyzing Attention Heads of Video DiTs} 
\label{sec:analysis}
% \begin{wrapfigure}{r}{0.48\columnwidth} % L: Left, 너비는 컬럼의 약 절반
%     \vspace{-25pt}
%     \centering
%     \includegraphics[width=\linewidth]{Figure/fig3_final.jpg}
%     \vspace{-15pt}
    % \caption{Head Configuration Comparison.
    %     (a) Attention maps exhibit distinct patterns: \textit{temporal heads} capture cross-frame diagonals, while \textit{spatial heads} maintain intra-frame locality. 
    %     (b) Quantitative evaluation using directional alignment and correlation shows that temporal heads align more closely with reference motion. 
    %     (c) Displacement maps further confirm that temporal heads more accurately capture motion flow.
    %     }
%     \vspace{-5pt}
%     \label{fig:fig3}
% \end{wrapfigure}

% \begin{wrapfigure}{R}{0.48\columnwidth} % R: Right
%     \centering
%     \vspace{-25pt}
%     \includegraphics[width=\linewidth]{Figure/figure4.pdf}
%     \caption{
%         Analysis of the relationship between the attention map and the structural cues encoded in hidden features. 
%         A strong correlation is observed: heads with lower attention entropy exhibit lower feature entropy, indicating higher structural fidelity.
%         Example: L20 H8 denotes the 8th attention head in the 20th layer.
%         }
%     \vspace{-30pt}
%     \label{fig:fig4}
% \end{wrapfigure}

\begin{figure}[t!]
    \centering
    % --- 왼쪽: Histogram (Fig 6) ---
    \begin{minipage}{0.49\columnwidth}
        \centering
        \includegraphics[width=\linewidth]{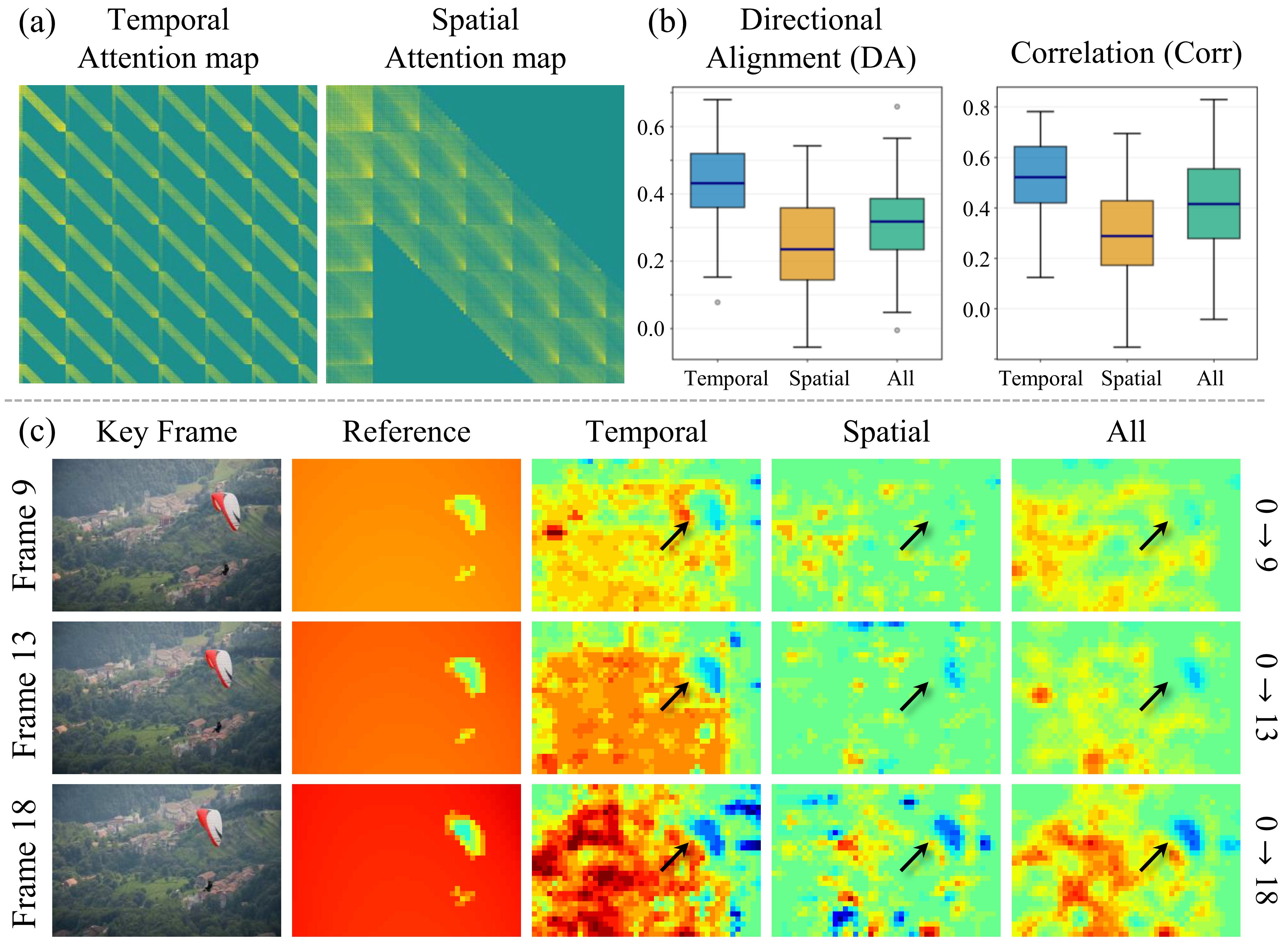}
\caption{Head Configuration Comparison.
(a) Attention maps show distinct patterns: \textit{temporal heads} capture cross-frame diagonals, while \textit{spatial heads} maintain intra-frame locality. 
(b) Quantitative evaluation using directional alignment and correlation shows that temporal heads align more closely with reference motion. 
(c) Displacement maps further confirm temporal heads more accurately capture motion flow.
}
        \label{fig:fig3}
    \end{minipage}
    \hfill % 두 이미지 사이 간격 확보
    % --- 오른쪽: Saliency/Entropy (Fig 7) ---
    \begin{minipage}{0.45\columnwidth}
        \centering
        \includegraphics[width=\linewidth]{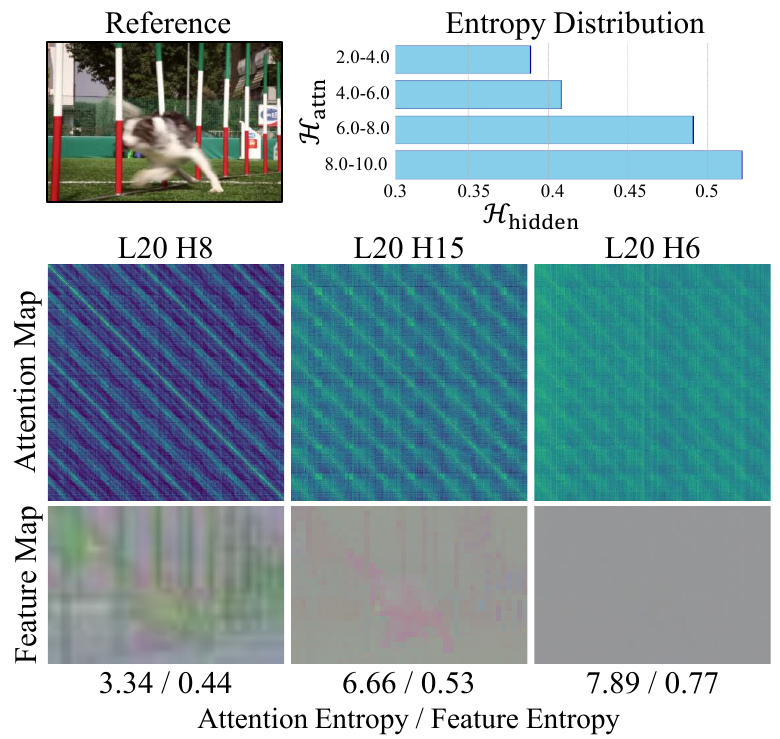}
        \vspace{-3mm}
\caption{
Relationship between attention maps and structural cues in hidden features. 
Heads with lower attention entropy show lower feature entropy, indicating stronger structural fidelity. 
Example: L20 H8 denotes the 8th attention head in the 20th layer.
}
        \label{fig:fig4}
    \end{minipage}
    \vspace{-10pt} % 하단 본문과의 간격 조절
\end{figure}

We focus on understanding how motion and structural information are represented within video DiTs~\cite{cogvideox}.
Here, we present a head-level analysis of attention heads to uncover their distinct functional roles in modeling motion and spatial structure, offering a mechanistic perspective overlooked in prior studies~\cite{kim2025seg4diff, sparsevideogen, ahn2025fine}.

\vspace{2pt}
\noindent\textbf{Analysis 1: Motion-Specific Heads.}
We aim to identify the motion-specific properties encoded within individual attention heads.
Building on SparseGen~\cite{sparsevideogen}, which identifies temporal and spatial head patterns for efficient video generation (see \namefig{}~\ref{fig:fig3}(a)), 
we first classify heads based on their similarity to pattern masks (e.g., cross-frame diagonal for temporal, sparse localized patterns for spatial).
% This selection is performed on-the-fly at inference time and requires no dataset-specific tuning. 
Based on this, we construct head-specific displacement maps from cross-frame attention~\cite{difftrack,ditflow}.
Specifically, cross-frame attention is computed using the queries $Q$ and keys $K$ derived from the latent features $z_f \in \mathbb{R}^{head \times N \times d}$ as:
%\small
\begin{equation}
\mathbf{M}^{h}_{f,f'} = \mathrm{softmax}\left(\frac{Q^{h}_f \cdot K^{h \top}_{f'}}{\sqrt{d}}\right),
%\tag{1}
\label{eq:1}
\end{equation}
%\normalsize
where $h$ denotes the attention head, $N=H \times W$ is the number of spatial tokens, and $d$ is the feature dimension. 
Here, $f$ and $f'$ represent the frame indices. 
We then derive a head-specific displacement map $\mathcal{D}^{h}_{f,f'} \in \mathbb{R}^{H \times W \times 2}$, which captures the patch-wise motion between frames $(f, f')$, defined as:
%\small
%\begin{align}
\begin{equation}
\mathrm{I}^{h}_{f,f'} = \underset{f'}{\operatorname{argmax}} \left( \mathbf{M}^{h}_{f,f'} \right), 
%\tag{2}
%\label{eq:2} 
\quad
\mathcal{D}^{h}_{f,f'}(i) = \mathbf{g}\big(\mathrm{I}^{h}_{f,f'}(i)\big) - \mathbf{g}(i),
%\tag{3}
\label{eq:3}
%\end{align}
\end{equation}
%\normalsize
where $\mathbf{g}(\cdot)$ converts a 1D patch index into 2D spatial coordinates and $i$ denotes the patch index.  

Using displacement maps, we analyze cross-frame motion patterns across attention heads to quantify motion capture effectiveness, as shown in \namefig{}~\ref{fig:fig3}(c). Temporal heads capture motion variations more accurately than spatial or combined heads, showing stronger alignment with object motion (blue) and background movement (red).
To validate these observations, we measure the similarity between predicted displacement maps and ground-truth optical flow~\cite{raft} through two quantitative metrics:
(1) Directional Alignment (DA), calculating cosine similarity between motion directions, and
(2) Correlation (Corr), using Pearson correlation for global pattern consistency.

As illustrated in \namefig{}~\ref{fig:fig3}(b), temporal heads consistently achieve higher DA and Corr scores. These results confirm that while spatial heads attend to intra-frame patch relationships, temporal heads specialize in cross-frame dependencies, effectively encoding motion-related information within video DiTs.

\vspace{2pt}
\noindent\textbf{Analysis 2: Structure-Specialized Heads.} 
To identify attention heads that encode structural information, we analyze the visual token attention maps $A^{h}$ corresponding to each head.
% Attention maps across different heads exhibit distinct structural patterns, as shown in \namefig{}~\ref{}. 
Our analysis reveals distinct patterns across different heads:
% Those exhibiting sharply aligned diagonal patterns indicate stable token-to-token alignments across frames, reflecting consistent spatial structure, whereas irregular or diffuse patterns suggest weak structural correlations, as shown in \namefig{}~\ref{fig:fig4}. 
Heads exhibiting well-defined diagonal alignment patterns with low entropy (e.g., L20 H8) stand in contrast to those with irregular or diffuse attention distributions with high entropy (e.g., L20 H6), as shown in \namefig{}~\ref{fig:fig4}.
% These variations suggest that certain heads inherently encode spatial structural relationships, whereas others primarily capture non-structural or context-dependent associations. 
% We connect this phenomenon to the structural information, as the entropy quantifies the uncertainty in capturing the relationships between visual tokens.
Since the entropy serves to quantify the uncertainty in capturing the relationships between visual tokens, low entropy implies the presence of structural information.

To validate this, we compute the attention entropy~\cite{attentropy} and the spatial entropy of corresponding attention features~\cite{spatialentropy,spatialentropy2,Kang}. 
Given a head $h$ with attention map $A^{h}$, the attention entropy $\mathcal{H}^{h}_{\text{attn}}$ is defined as
$\mathcal{H}^{h}_{\text{attn}} = -\sum_{k=1}^{N} A^{h}_k \log A^{h}_k$, 
where lower values indicate a more concentrated attention distribution.
% , implying that the head focuses on spatially coherent regions. 
To capture spatial properties in attention features, we compute the spatial entropy of PCA-transformed feature embeddings.

As illustrated in \namefig{}~\ref{fig:fig4}, lower entropy heads exhibit clear diagonal attention patterns and yield feature maps maintaining structural layouts.
% signifying stable patch-to-patch alignments across consecutive frames.
% Consequently, the hidden features of these heads yield feature maps exhibiting strong structural patterns.
% These heads predominantly correspond to coherent object regions whose token representations maintain consistent appearance over time, indicating that low-entropy attention serves as a structural prior of the model.
% Consequently, the attention outputs guide hidden features to emphasize spatially consistent regions, resulting in sharper and semantically localized representations.
\namefig{}~\ref{fig:fig4} also provides quantitative evidence of a strong correlation between lower attention entropy $\mathcal{H}_{\text{attn}}$ and lower hidden feature spatial entropy $\mathcal{H}_{\text{hidden}}$.
Therefore, these diagonal attention patterns indicate strong spatial patch correlations, validating that such heads are specialized in encoding structural information.

% \begin{figure}[t!]
%     \centering
%     \includegraphics[width=\columnwidth]{Figure/fig4_final.jpg}
%     \caption{
%     Analysis of the relationship between the attention map and the structural cues encoded in hidden features. 
%     A strong correlation is observed: heads with lower attention entropy exhibit lower feature entropy, indicating higher structural fidelity.
%     Example: L20 H8 denotes the 8th attention head in the 20th layer.
%     }
%     \vspace{-8pt}
%     \label{fig:fig4}
% \end{figure}
\section{Method}

\begin{figure*}[t!]
    \centering
    \includegraphics[width=\textwidth]{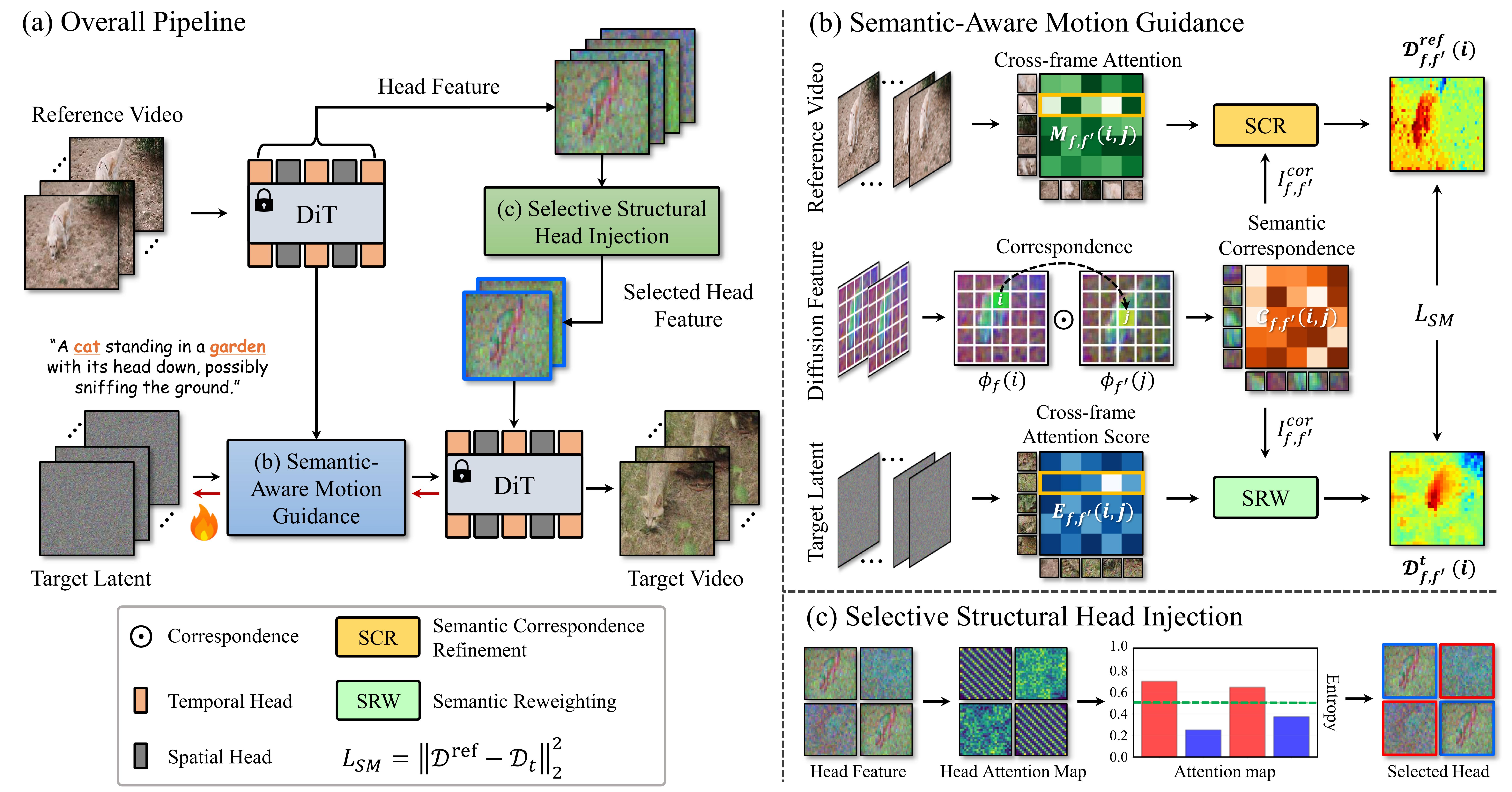}
    \caption{Overview of \namemethod{}. (a) From a reference video, we extract displacement maps and head features. Displacements from motion-specific heads guide motion by optimizing the latent representation, while selected head features from the reference are injected to preserve structure during generation. (b) To enhance motion guidance, semantic correspondence derived from diffusion features refines the displacement map, ensuring semantically aligned motion flow. (c) For structural guidance, we inject value features from heads selected via entropy analysis, targeting heads that encode essential spatial information.}
    \label{fig:fig5}
    \vspace{-13pt}
\end{figure*}

We introduce \textit{\namemethod{}}, a head-aware controllable motion transfer framework that jointly achieves \textit{motion fidelity} and \textit{structural alignment} with a reference video (see \namefig{}~\ref{fig:fig5}(a)). Building on \namesec{}~\ref{sec:analysis}, we propose (i) semantic-aware motion optimization (\namesec{}~\ref{sec:motion}) and (ii) structure-guided feature injection (\namesec{}~\ref{sec:structure}), illustrated in \namefig{}~\ref{fig:fig5}(b)–(c).

\vspace{-7pt}
\subsection{Semantic-Aware Motion Guidance}
\label{sec:motion}
For motion transfer, we construct cross-frame displacement maps within DiTs. Our analysis (Analysis~1, \namesec{}~\ref{sec:analysis}) shows that motion-specific heads $\mathcal{M}$ capture motion more effectively. We therefore aggregate cross-frame attention $\mathbf{M}^{h}_{f,f'}$ for $h \in \mathcal{M}$, identified via temporal pattern masks to extract a displacement map.

While cross-frame attention provides patch-level correlation cues, it can overemphasize visually similar yet semantically unrelated regions, producing mismatched displacements and degraded motion alignment (e.g., camera-following movements that disregard objects and incorrect object trajectories; see Supp.~\namesec{}~B).
To mitigate this, we introduce a semantic-aware motion guidance that refines displacements using semantic similarities from semantically rich diffusion features~\cite{dift, cove, unsupervised}, as shown in \namefig{}~\ref{fig:fig5}(b).

\begin{figure*}[t!]
    \centering
    \includegraphics[width=\textwidth]{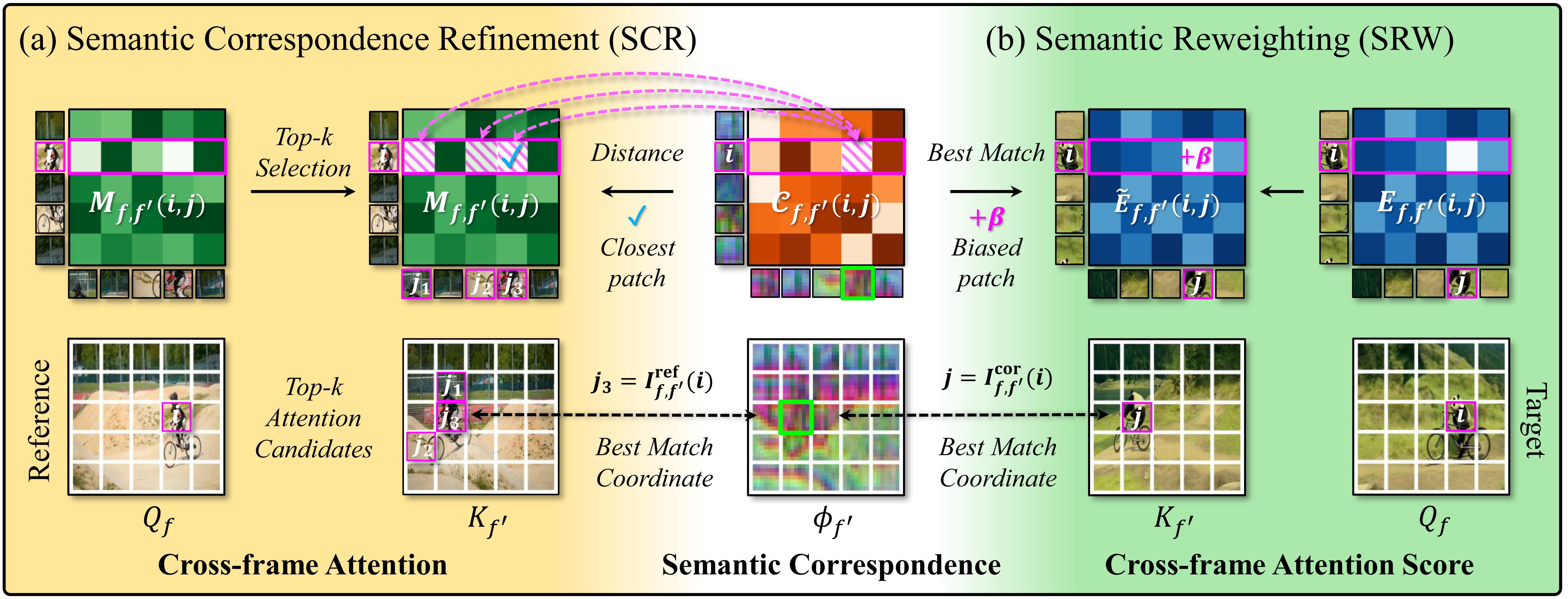}
    \vspace{-2mm}
    \caption{
    Details of Semantic Correspondence Refinement (SCR) and Semantic Reweighting (SRW). (a) $\mathrm{I}^{\text{ref}}_{f,f'}$ is obtained by choosing, among top-$k$ attention candidates, the patch closest to the semantic best match $\mathrm{I}^{\text{cor}}_{f,f'}$. (b) SRW manipulates target cross-frame attention by adding a correspondence-based bias at $\mathrm{I}^{\text{cor}}_{f,f'}$.
    }
    \label{fig:fig6}
    \vspace{-4mm}
\end{figure*}

\begin{figure}[tp]
%    \vspace{-20pt}   
\centering
\includegraphics[width=\columnwidth]{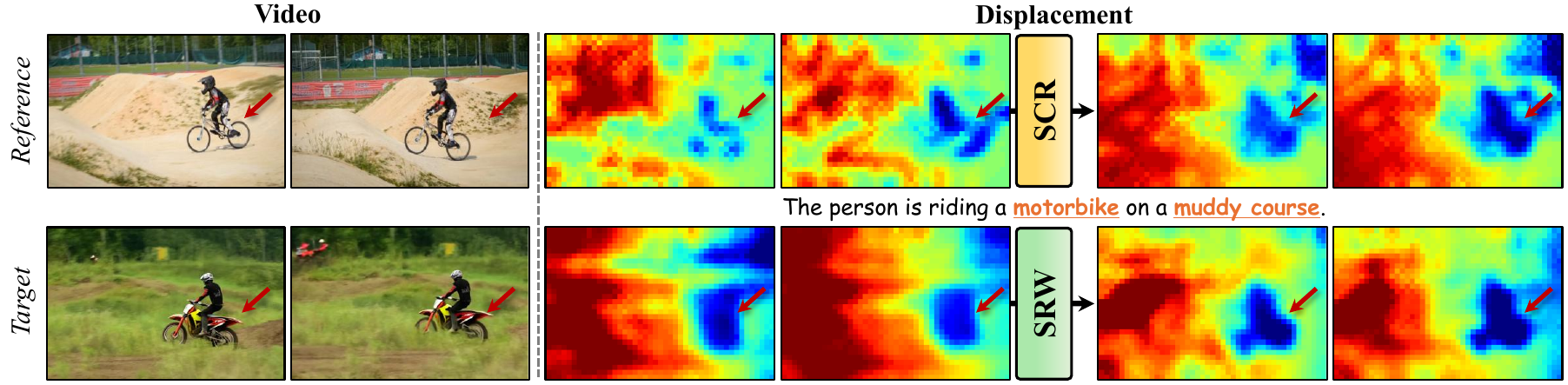}
\vspace{-2mm}
    \caption{Effect of SCR and SRW on displacement maps. The refinements improve robustness to fine-grained object motion and better preserve object shape.}
    \label{fig:fig7}
    \vspace{-4mm}
%\end{wrapfigure}
\end{figure}

\vspace{1mm}
\noindent\textbf{Semantic Correspondence Refinement.}
To refine the reference displacement map, we first select the top-$k$ attention candidates from cross-frame attention rather than relying on a single maximum, as illustrated in \namefig{}~\ref{fig:fig6}(a). 
We then construct semantic correspondence $\mathcal{C}$ using pairwise cosine similarity over diffusion features $\phi$:
%\small
%\begin{align*}
\begin{equation}
\mathcal{C}_{f, {f'}} = \frac{\phi_f\cdot \phi_{f'}}{\Vert{\phi_f}\Vert_2 \Vert{\phi_{f'}}\Vert_2},
%\tag{4}
%\end{align*}
\end{equation}
\normalsize
% and define the most similar semantic indices $\mathrm{I}^{\text{cor}}_{f,f'} \in \mathbb{R}^{N}$ between frames
% We compute distances between $\mathrm{I}^{\text{cor}}$ and the top-$k$ attention candidates and choose the one with the smallest distance as the reference best-match coordinate $\mathrm{I}^{\text{ref}}_{f,f'}$.
For each patch index, we compute the distances between the most similar semantic indices $\mathrm{I}^{\text{cor}}_{f,f'} \in \mathbb{R}^{N}$ between frames and the top-$k$ attention candidates.
% Among these candidates, we compare their spatial coordinates with the corresponding semantic match given by $\mathrm{I}^{\text{cor}}_{f,f'}$ and choose the closest one as the reference best-match coordinate $\mathrm{I}^{\text{ref}}_{f,f'}$.
The nearest candidate is then selected as the reference best-match coordinate $\mathrm{I}^{\text{ref}}_{f,f'}$.
% This selection preserves the reliability of attention-based correspondences while aligning the displacement estimate with semantic similarity.
The reference inter-frame displacement map $\mathcal{D}^{\text{ref}}$ is then constructed from $\mathrm{I}^{\text{ref}}_{f,f'}$ following Eq.~\ref{eq:3} (first row in \namefig{}~\ref{fig:fig7}).

\vspace{1mm}
\noindent\textbf{Motion Optimization.}
Given $\mathcal{D}^{\text{ref}}$, we optimize $z_T$ to align generated motion with the reference. We apply Semantic Reweighting (SRW) to adjust cross-frame attention via a correspondence-guided bias derived from $\mathcal{C}$ (\namefig{}~\ref{fig:fig6}(b)). Using $\mathrm{I}^{\text{cor}}$, we form the bias matrix $\mathcal{B}_{f,f'}$:
%\small
%\begin{equation}
\begin{align} 
\mathcal{B}_{f,f'}(i,j) =
\begin{cases}
\beta, & \text{if } j = \mathrm{I}^{\text{cor}}_{f,f'}(i), \\[3pt]
0, & \text{otherwise,}
\end{cases}
%\tag{5}
\label{eq5}
\end{align}
%\end{equation}
\normalsize
where $\beta$ controls bias strength. The refined cross-frame attention scores are $\tilde{E}_{f,f'} = E_{f,f'} + \mathcal{B}_{f,f'}$, and refined attention is $\tilde{M}_{f,f'}=\mathrm{softmax}(\tilde{E}_{f,f'})$. We then estimate displacement as an expectation over coordinate differences, 
%\small
%\begin{align}
\begin{equation}
\Delta x_{f,f'} = \sum_{i,j} (x_j - x_i)\tilde{M}_{f,f'}(i,j), \quad
\Delta y_{f,f'} = \sum_{i,j} (y_j - y_i)\tilde{M}_{f,f'}(i,j),
%\tag{6}
%\end{align}
\end{equation}
\normalsize
and stack them in $\mathcal{D}_{t} = [\Delta x_{f,f'}, \Delta y_{f,f'}]$ (second row in \namefig{}~\ref{fig:fig7}). 
% \begin{figure*}[t!]
%     \centering
%     \includegraphics[width=\textwidth]{Figure/fig7.png}
%     \vspace{-18pt}
%     \caption{Effect of SCR and SRW on displacement maps. The refinements improve robustness to fine-grained object motion and better preserve object shape.}
%     \label{fig:fig7}
%     \vspace{-7pt}
% \end{figure*}
Finally, to optimize the latent for motion alignment, we minimize the semantic motion loss $L_{SM}$, defined as the L2 distance between the reference displacement map $\mathcal{D}_{\text{ref}}$ and the target displacement map $\mathcal{D}_{t}$. 
% This loss aligns patch-wise displacement vectors across frame pairs, encouraging the generated video to follow the reference motion under SRW-guided semantic correspondence.
Incorporating semantic cues into displacement estimation unifies motion alignment with semantic coherence, yielding consistent motion transfer.
Notably, this strategy represents the first attempt to incorporate semantic correspondences within motion representations, facilitating high-fidelity, semantically consistent video generation.

\subsection{Selective Structural Head Injection}
\label{sec:structure}
Although displacement maps capture directional motion, they lack sufficient intra-frame structure, which can cause spatial misalignment with the reference (\namefig{}~\ref{fig:fig2}, right). 
To address this, we propose a structural guidance mechanism for preserving the spatial integrity of the reference. 
We inject reference value features during generation. Naïvely injecting all head features, however, introduces artifacts such as noise and identity leakage~\cite{motionbyqueries} (see Supp.~\namesec{}~C.4, Supp.~\namefig{}~6).

We therefore select structurally informative heads via entropy-based analysis (Analysis~2, \namesec{}~\ref{sec:analysis}). 
Low-entropy heads exhibit strong diagonal attention and produce spatially coherent features; we inject their value features into the corresponding heads, supplying structure-aware guidance without noise or leakage (\namefig{}~\ref{fig:fig5}(c)). 
This selective injection complements motion optimization and enforces spatial alignment with the reference.

% \begin{figure}[t!]
%     \centering
%     \includegraphics[width=0.5\columnwidth]{Figure/fig7_final.jpg}
%     \caption{Effect of SCR and SRW on displacement maps. The refinements improve robustness to fine-grained object motion and better preserve object shape.}
%     \label{fig:fig7}
%     \vspace{-10pt}
% \end{figure}

\label{sec:method}

\section{Experiments}
\label{sec:experiment}

\noindent\textbf{Implementation Details.}
Following prior works, we evaluate our approach using a standard motion transfer benchmark and settings~\cite{smm, ropecraft, det}.
Across all experiments, we adopt the video diffusion transformer CogVideoX~\cite{cogvideox} as the base model, using 50 denoising steps, 12 optimization steps $T_{\text{opt}}$, and a guidance scale of 7, consistent with common practice. 
For additional results with the video DiT model, Wan~\cite{wan}, please refer to Supp.\ Sec.\ D.
The bias strength $\beta$ for semantic reweighting is fixed to 0.1, and the entropy range $\tau$ for selective structure specialized-head injection is set to $\tau < 7$, based on 7 being the median entropy value observed in our analysis (validated in Section~\ref{sec:ablation_exp}).  
Hyperparameter configurations (e.g., $\beta$, $T_{\text{opt}}$, $\tau$, top-$k$) and \namemethod{} optimization algorithm are detailed in Supp.\ Sec.\ C.1 and C.2, respectively.
% Additional results, including those using an alternative video DiT backbone~\cite{wan} and various hyperparameter settings, are included in the supplementary material.

\vspace{1mm}
\noindent\textbf{Baselines.}
We compare against representative motion-transfer approaches, including U-Net-based methods (MOFT~\cite{moft}, ConMO~\cite{conmo}, and MotionClone~\cite{motionclone}) and recent DiT-based methods (DiTFlow~\cite{ditflow}, RoPECraft~\cite{ropecraft}, and GWTF~\cite{gowithflow}).

\vspace{1mm}
\noindent\textbf{Movie Scene Dataset.}
Motion transfer has emerged as a pivotal task in video production and digital cinematography, where it is increasingly demanded to replicate sophisticated film-style shot compositions and intricate post-production effects~\cite{ma2025controllable}. 
However, existing benchmarks~\cite{motionclone, det} are largely derived from generic internet videos (e.g., DAVIS/WebVid), which do not fully reflect these professional requirements. 
% To evaluate robustness under this distribution shift
Therefore, we curate a Movie Scene Dataset specifically for the motion transfer task (see Supp.\ Sec.\ E for details). 
% This dataset enables a more practical assessment of models in complex, real-world cinematic scenarios.
Complementing existing benchmarks, this dataset provides a production-oriented evaluation setting to assess whether methods can achieve faithful motion transfer under real-world cinematic conditions.
% Rather than functioning as a large-scale benchmark, this dataset complements existing benchmarks by providing a production-oriented evaluation setting with cinematic scenarios, including controlled camera trajectories, film-shot layouts, and complex object interactions.

\vspace{-9pt}

\subsection{Main Results}

\begin{figure*}[tp]
    \centering
    \includegraphics[width=\textwidth]{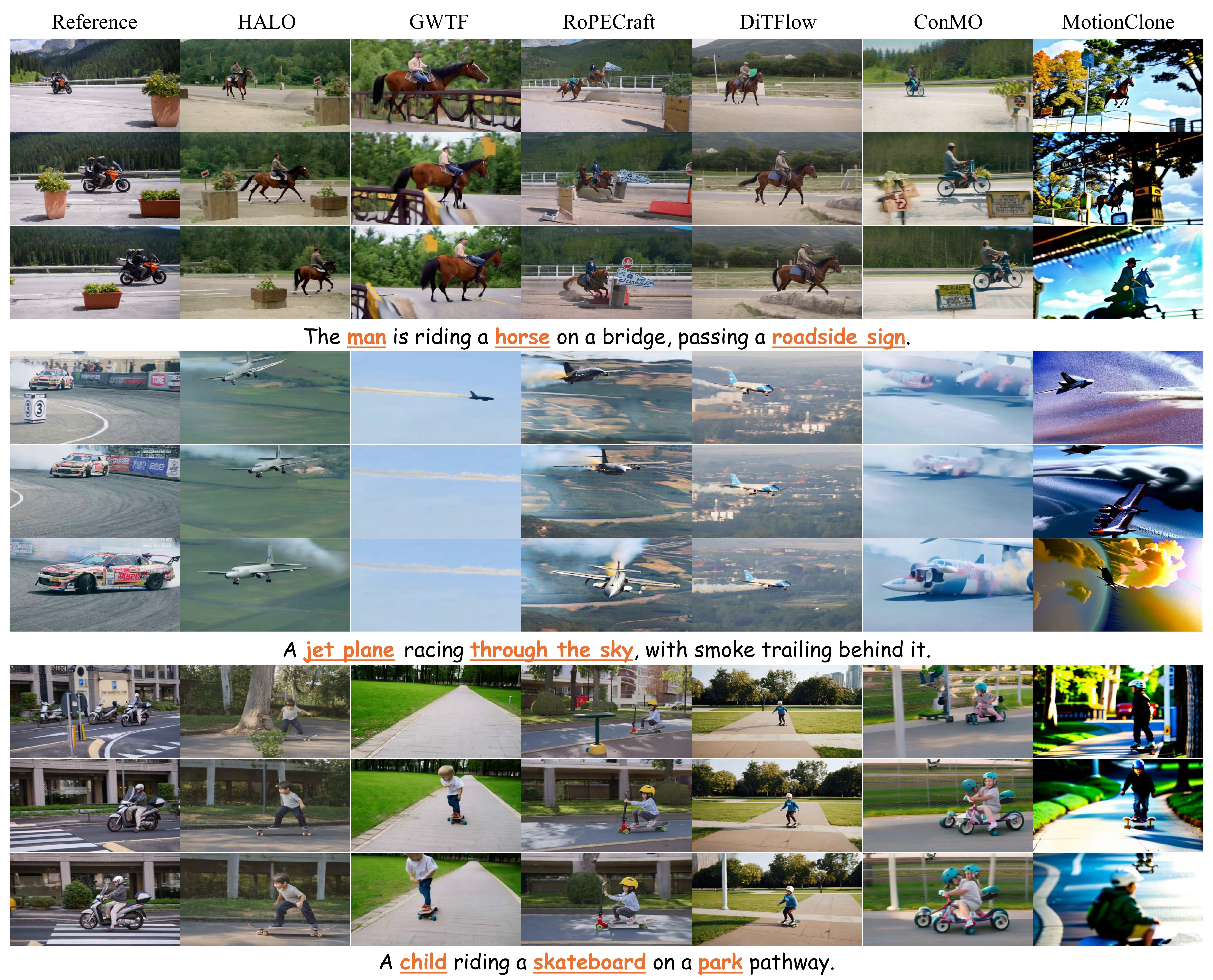}
    \vspace{-18pt}
    \caption{Qualitative comparison between U-Net- and DiT-based baselines and \namemethod{}.}
    \label{fig:fig8}
    \vspace{-8pt}
\end{figure*}

\begin{figure*}[t!]
    \centering
    \includegraphics[width=\textwidth]{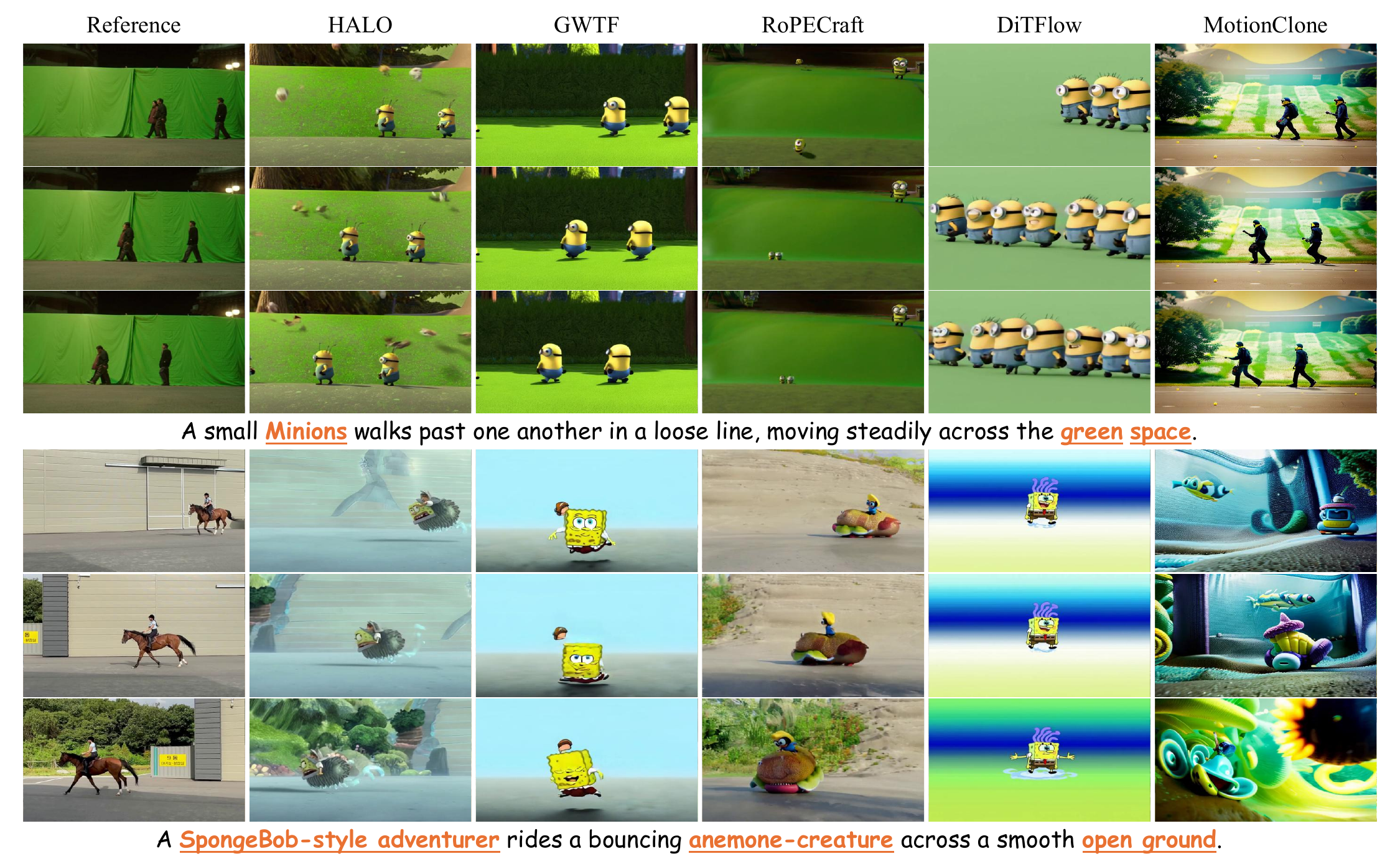}
    \vspace{-15pt}
    \caption{Qualitative results in the movie scene dataset.}
    \label{fig:fig_movie}
    \vspace{-8pt}
\end{figure*}

\noindent\textbf{Qualitative Results.}
\namefig{}~\ref{fig:fig8} presents qualitative comparisons with state-of-the-art models. 
Baseline methods often exhibit camera drift, generation failures, or identity leakage (e.g., generating a motorcycle instead of a horse). They also produce semantically inconsistent motion, such as misaligned orientations between correlated objects (e.g., car vs.\ airplane). In contrast, our method produces semantically coherent motion aligned with the reference while preserving subject identity and structural integrity.  
% \namefig{}~\ref{fig:fig_movie} presents qualitative comparisons on the Movie Scene Dataset, where our method maintains faithful motion alignment while preserving subject identity under cinematic shot composition and post-production effects.
Furthermore, in \namefig{}~\ref{fig:fig_movie}, \namemethod{} exhibits remarkable robustness on the Movie Scene Dataset.
Even when guided by complex cinematic prompts, \namemethod{} faithfully preserves the target's identity integrity while following the reference dynamics.

% % \vspace{-5pt}
% \begin{figure*}[t!]
%     \centering
%     \includegraphics[width=\textwidth]{Figure/movie_baseline.pdf}
%     \vspace{-15pt}
%     \caption{Qualitative results in the movie scene dataset.}
%     \label{fig:fig_movie}
%     \vspace{-8pt}
% \end{figure*}

% \begin{figure*}[p] % [p]는 해당 피겨들을 별도의 한 페이지에 모으라는 옵션입니다. [t!]도 가능합니다.
%     \centering
    
%     % --- 첫 번째 그림 ---
%     \begin{minipage}{\textwidth}
%         \centering
%         \includegraphics[width=\textwidth]{Figure/fig8_final.jpg}
%         \vspace{-10pt}
%         \caption{Qualitative comparison between U-Net- and DiT-based baselines and \namemethod{}.}
%         \label{fig:fig8}
%     \end{minipage}

%     \vspace{15pt} % 두 그림 사이의 간격 조절

%     % --- 두 번째 그림 ---
%     \begin{minipage}{\textwidth}
%         \centering
%         \includegraphics[width=\textwidth]{Figure/movie_baseline.pdf}
%         \vspace{-10pt}
%         \caption{Qualitative results in the movie scene dataset.}
%         \label{fig:fig_movie}
%     \end{minipage}
    
% \end{figure*}

\vspace{1mm}
\noindent\textbf{Quantitative Results.}
We evaluate performance using CLIP score (CLIP)~\cite{clipscore} for text–video alignment, Temporal Consistency (TC)~\cite{dino} for frame coherence, and Motion Fidelity (MF)~\cite{smm} and FTD~\cite{ropecraft} for reference-motion alignment.  
As shown in \nametab{}~\ref{tab:main_results}, \textit{\namemethod{}} achieves notable improvements in CLIP, MF, and FTD, demonstrating that head-level analysis within DiTs enables motion transfer with strong motion fidelity and structural alignment.
% Further quantitative evidence from our Movie Scene Dataset is included in the supplementary material, validating \namemethod{}'s robustness in challenging cinematic scenarios.

Since motion transfer requires following the reference motion while adhering to the target text prompt, CLIP and MF must be jointly evaluated. 
Prior works exhibit a trade-off between these metrics, as shown in Table~\ref{tab:main_results}. 
In contrast, our method achieves balanced performance, maintaining high CLIP and MF scores.
% \namefig{}~\ref{fig:quantitative_analysis}.  
This indicates that our head-analysis-based formulation transfers reference motion effectively while suppressing irrelevant semantic information.
Although \namemethod{} reports a slightly lower TC than DiTFlow and GWTF in \nametab{}~\ref{tab:main_results}, TC tends to favor more static outputs. 
Further analysis of the CLIP--MF balance and the relationship between motion dynamics and temporal consistency is provided in Supp.\ Sec.\ F.2.
We also report the runtime and peak GPU memory in Supp.\ Sec.\ F.3, showing that \namemethod{} incurs only marginal overhead relative to the displacement optimization framework.
% takes 1125\,s and peaks at 38.2\,GB with only marginal overhead from our module.

% \vspace{-5pt}

\vspace{1mm}
\noindent\textbf{User Study.}
% We conduct a user study with 20 participants to assess editing accuracy, temporal consistency, and motion accuracy (see Supp.\ Sec.\ G for details). 
% As reported in \nametab{}~\ref{tab:user_study}, \namemethod{} achieves the highest mean score across all criteria, indicating consistent user preference.
To ensure an assessment aligned with human preference, we conduct a user study to evaluate \namemethod{} across three criteria: editing accuracy, temporal consistency, and motion accuracy. 
Twenty participants rated each result on a 5-point Likert scale (1: lowest, 5: highest), with scores subsequently normalized to a 0--100 range. 
As reported in \nametab{}~\ref{tab:user_study}, \namemethod{} achieves the highest mean score across all criteria, demonstrating a consistent preference among participants. 

% \begin{figure}[t!]
%     \centering
%     \vspace{-1pt}
%     \includegraphics[width=\columnwidth]{Figure/fig9_final.jpg}
%     \vspace{-10pt}
%     \caption{Objective comparison across motion-transfer methods in terms of Motion Fidelity (MF) and CLIP Score (CLIP). Baseline methods exhibit a trade-off between MF and CLIP, whereas \namemethod{} (Ours) achieves balanced performance.}
%     \vspace{-29pt}
%     \label{fig:quantitative_analysis}
% \end{figure}

% \vspace{2pt}

\subsection{Ablation Study}
\label{sec:ablation}

We perform ablation studies under conditions identical to the main experiments, using displacement optimization~\cite{ditflow} as the baseline.  

\begin{table*}[t]
    \centering
    
    % --- 왼쪽: Main Quantitative Results ---
    \begin{minipage}[t]{0.6\textwidth}
        \centering
        \caption{Quantitative results with state-of-the-art motion transfer methods. Best and second results are represented with \textbf{bold} and \underline{underlined}.}
        \label{tab:main_results}
        \vspace{-4pt}
        \begin{adjustbox}{max width=\linewidth}
            \setlength{\tabcolsep}{5pt}
            \renewcommand{\arraystretch}{1.05}
            \begin{tabular}{ll c cccc}
                \specialrule{0.8pt}{1pt}{1pt}
                \multicolumn{2}{l}{Model} & & CLIP $\uparrow$ & TC $\uparrow$ & MF $\uparrow$ & FTD $\downarrow$ \\ 
                \specialrule{0.5pt}{1pt}{1pt}
                \multirow{3}{*}{U-Net-based} & MoFT~\cite{moft} & \footnotesize{\textcolor{gray}{NeurIPS'24}} & 30.9 & 85.8 & 34.8 & 23.0 \\
                & ConMO~\cite{conmo} & \footnotesize{\textcolor{gray}{CVPR'25}} & 29.8 & 85.3 & 52.0 & \textbf{17.4} \\
                & MotionClone~\cite{motionclone} & \footnotesize{\textcolor{gray}{ICLR'25}} & 30.4 & 78.6 & 55.6 & 19.8 \\
                \specialrule{0.5pt}{0.5pt}{0.5pt}
                \multirow{4}{*}{DiT-based} & DiTFlow~\cite{ditflow} & \footnotesize{\textcolor{gray}{CVPR'25}} & 31.0 & \textbf{89.5} & 59.6 & 23.0 \\
                & RoPECraft~\cite{ropecraft} & \footnotesize{\textcolor{gray}{NeurIPS'25}} & 30.3 & 85.9 & 58.2 & 19.6 \\
                & GWTF~\cite{gowithflow} & \footnotesize{\textcolor{gray}{CVPR'25}} & \underline{31.6} & \underline{88.2} & \underline{62.5} & 21.6 \\
                \rowcolor{emerald!20} \cellcolor{white} & \namemethod{} & & \textbf{31.7} & 87.5 & \textbf{66.2} & \underline{19.4} \\
                \specialrule{0.8pt}{1pt}{1pt}
            \end{tabular}
        \end{adjustbox}
    \end{minipage}
    \hfill % 두 테이블 사이 간격
    % --- 오른쪽: Movie Scene Dataset Results ---
    \begin{minipage}[t]{0.38\textwidth}
        \centering
        \caption{Quantitative results in the Movie Scene dataset (methods requiring video masks are excluded).}
        \label{tab:supp_movie}
        \vspace{-7pt}
        \begin{adjustbox}{max width=\linewidth}
            \setlength{\tabcolsep}{5pt}
            \renewcommand{\arraystretch}{1.05}
            \begin{tabular}{l ccc}
                \specialrule{0.8pt}{1pt}{1pt}
                {Model}  & CLIP $\uparrow$ & TC $\uparrow$ & MF $\uparrow$ \\ 
                \specialrule{0.5pt}{1pt}{1pt}
                MotionClone~\cite{motionclone}  & 27.1 & 74.5 & 47.7 \\
                DiTFlow~\cite{ditflow} & 29.7 & \textbf{90.4} & 48.4 \\
                RoPECraft~\cite{ropecraft} & 28.3 & 88.2 & \underline{49.1} \\
                GWTF~\cite{gowithflow}  & \underline{30.2} & 87.6 & 46.6 \\
                \rowcolor{emerald!20} \namemethod{}  & \textbf{30.5} & \underline{88.9} & \textbf{52.6} \\
                \specialrule{0.8pt}{1pt}{1pt}
            \end{tabular}
        \end{adjustbox}
    \end{minipage}

\end{table*}

\begin{table*}[t] % 전체 너비 사용을 위해 table* 권장
    \centering
    
    % 왼쪽: Ablation Results
    \begin{minipage}[t]{0.58\textwidth}
        \centering
        \captionof{table}{Ablation results of proposed components on performance. “Semantic” denotes the semantic-aware motion guidance, and “Injection” denotes the selective structural head injection.}
        \label{tab:ablation}
        \vspace{-4pt}
        \resizebox{\linewidth}{!}{%
            \begin{tabular}{ccccccc}
            \specialrule{0.8pt}{1pt}{1pt}
            Exp.\# & Semantic & Injection & CLIP $\uparrow$ & TC $\uparrow$ & MF $\uparrow$ & FTD $\downarrow$ \\
            \specialrule{0.5pt}{1pt}{1pt}
            1 & & & 31.0 & 89.5 & 59.6 & 23.0 \\
            2 &  & \checkmark & 30.6 & 88.3 & 61.8 & 20.5  \\
            3 & \checkmark & & 30.9 & 88.5 & 61.3 & 20.9   \\
            \rowcolor{emerald!20} 4 (\namemethod{}) & \checkmark & \checkmark & 31.7 & 87.5 & 66.2 & 19.4 \\
            \specialrule{0.8pt}{1pt}{1pt}
            \end{tabular}
        }
    \end{minipage}
    \hfill % 두 테이블 사이 간격
    % 오른쪽: User Preference / Comparison (기존 wraptable 내용)
    \begin{minipage}[t]{0.38\textwidth}
    \centering
    \captionof{table}{User preference study on motion transfer. The results represent the preference rate across three dimensions.} % 적절한 캡션 추가
    \label{tab:user_study}
    \vspace{-3.5pt}
    \resizebox{\linewidth}{!}{%
        \begin{tabular}{lccc} % 4개 컬럼으로 변경
        \specialrule{0.8pt}{1pt}{1pt}
        Model & Edit Acc. & TC & Motion Acc. \\
        \specialrule{0.5pt}{1pt}{1pt}
        DiTFlow   & 84.1 & 83.1 & 69.0 \\
        RoPECraft & 72.4 & 75.6 & 77.6 \\
        GWTF      & 84.0 & 77.1 & 72.8 \\
        \rowcolor{emerald!20} \textit{HALO} & \textbf{86.5} & \textbf{87.8} & \textbf{93.3} \\
        \specialrule{0.8pt}{1pt}{1pt}
        \end{tabular}
    }
    \end{minipage}
    
    \vspace{-10pt} % 본문과의 간격 조정
\end{table*}

\begin{figure}[t!]
    \centering
    \vspace{-2.2pt}
    \includegraphics[width=\columnwidth]{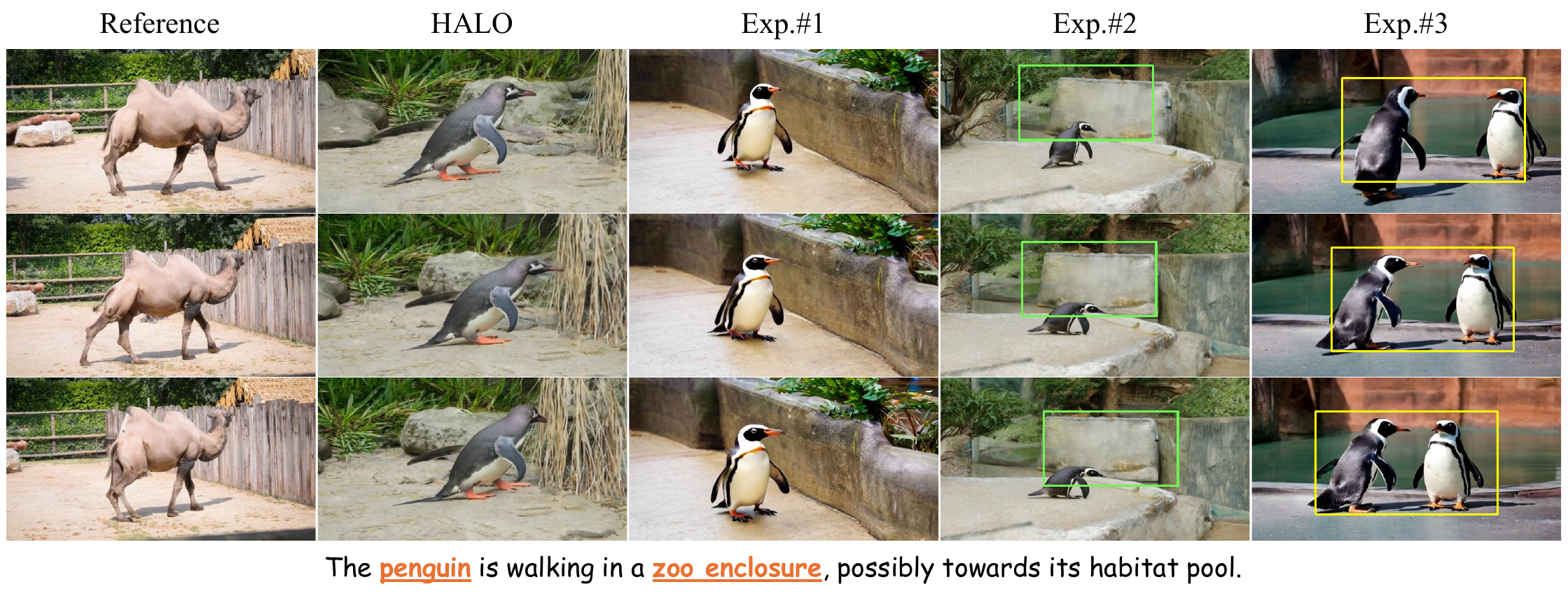}
    \caption{Qualitative results of our method across ablation studies. Exp.\# corresponds to \nametab{}~\ref{tab:ablation}.}
    \vspace{-4mm}
    \label{fig:ablation}
\end{figure}

\vspace{2pt}
\noindent\textbf{Effect of Semantic-Aware Motion Guidance.}
To evaluate the role of semantic correspondence refinement in displacement computation, we directly compute displacement maps from raw cross-frame attention.  
As shown in \namefig{}~\ref{fig:ablation} (green box), the absence of semantic refinement causes motion to the background rather than the actual moving object. 
This demonstrates that cross-frame attention alone captures visually similar but semantically unrelated regions, leading to motion misalignment. 
Semantic correspondence refinement is therefore crucial for achieving semantically consistent motion transfer.

\vspace{1mm}
\noindent\textbf{Effect of Selective Structure-Specialized Head Injection.}
Removing selective structure-specialized head injection significantly degrades motion fidelity (Exp.\#3 in \nametab{}~\ref{tab:ablation}).  
As shown in \namefig{}~\ref{fig:ablation} (yellow box), this omission results in structural distortions and duplicated objects (e.g., two penguins).  
These findings indicate that injecting low-entropy structural features helps maintain spatial consistency, making explicit structural guidance essential for accurate motion transfer.

\vspace{-7pt}

\subsection{Additional Validations}
\label{sec:ablation_exp}

\noindent\textbf{Complex Motion.}
% There has been a growing emphasis on the robustness of motion transfer models when handling complex motions.
% To address this, we evaluate the robustness of \namemethod{} under challenging scenarios. 
Robust motion transfer remains challenging under rapid movements and long-range trajectories. 
To evaluate the robustness of \namemethod{} in such settings, we utilize a benchmark stratified into three difficulty levels (Easy/Medium/Hard) based on trajectory complexity. 
We assess performance using prompt alignment (CLIP) and motion fidelity (MF), while omitting MoFT due to its lack of competitiveness in these scenarios.
% notably, MoFT is excluded from this comparison due to its significantly lower performance.

\begin{figure}[t!]
    \centering
    \vspace{-5pt}
\includegraphics[width=0.99\columnwidth]{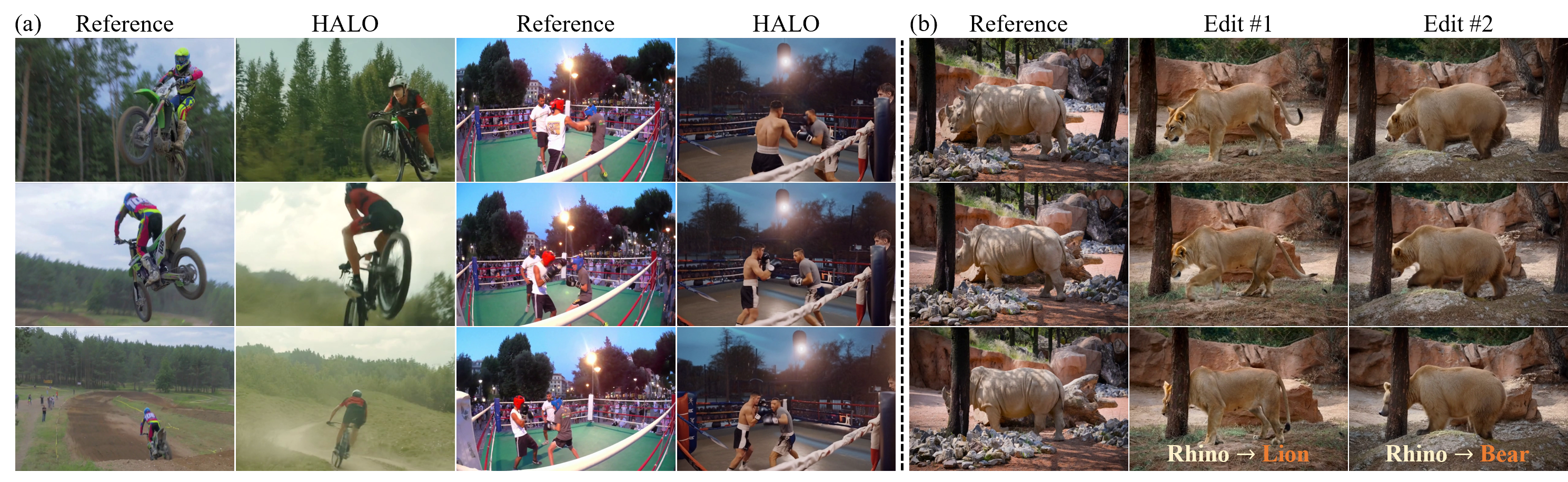}
    \caption{Qualitative results on complex motion and video editing. (a) Generated samples illustrating the \namemethod{}'s capability in handling complex motion dynamics. (b) Application to video editing using the structure-specialized head.}
    \vspace{-4mm}
    \label{fig:video_edit}
\end{figure}

As summarized in \nametab{}~\ref{tab:difficulty}, \namemethod{} consistently achieves superior performance across all levels with minimal performance decay, indicating high stability. 
% While MF decreases as difficulty increases, \namemethod{} consistently achieves the best MF across all levels, indicating robust motion transfer under complex trajectories.
This robustness is further evidenced qualitatively in \namefig{}~\ref{fig:video_edit}(a); our method handles extreme cases effectively, such as occlusions in a motorcycle jump and the high-velocity dynamics of boxing. 
Overall, these results validate that our approach remains stable across a wide spectrum of complex motions.

\vspace{1mm}
\noindent\textbf{Validation of Motion-Specific Heads.}
To verify motion-head selection, \nametab{}~\ref{tab:analysis_a} compares motion-specific, spatial, and all-head configurations.  
Spatial heads alone yield significantly lower MF and FTD, indicating a limited motion-related signal for motion transfer.  
Although aggregating all heads offers a slight improvement, it still underperforms motion-specific heads. 
% This confirms that spatial heads encode less motion information and that indiscriminate aggregation dilutes motion cues.
These results suggest that incorporating spatial heads can dilute relevant motion cues; in contrast, isolating motion-specific heads better captures dynamic correlations essential for motion transfer.
% In contrast, motion-specific heads effectively capture dynamic correlations, leading to superior motion transfer.  
Qualitative comparisons are included in Supp.\ Sec.\ C.5.

% To demonstrate the generalizability of the motion head across diverse applications beyond motion transfer, we conduct additional experiments focused on motion dynamics. 
To demonstrate the broader applicability of motion-specific heads as a motion-control mechanism, we conduct additional experiments focusing on motion dynamics. 
Building on Layer Perturbation Guidance~\cite{difftrack}, we extend this approach to our motion-specific head by introducing head perturbation during inference.
Evaluation using VBench metrics~\cite{vbench} on 100 video samples reveals that integrating this method with CogVideoX~\cite{cogvideox} improves image quality by $+14\%$, aesthetics by $+8\%$, and motion dynamics by $+15\%$.
These results suggest that our motion-head analysis provides a lightweight yet effective mechanism to guide motion in DiT-based architectures.

\begin{table*}[t]
    \centering
    % \renewcommand{\arraystretch}{1.0} % 필요시 행 간격 조절
    
    % ========== 왼쪽: Table 4 (Analysis Ablation) ==========
    \begin{minipage}[t]{0.5\textwidth}
    \centering
    \captionof{table}{Performance of motion-transfer methods across different motion difficulty levels categorized based on the complexity of motion trajectories.}
    \label{tab:difficulty}
    \vspace{-6pt}
    \resizebox{\linewidth}{!}{%
        \begin{tabular}{l cc cc cc}
        \specialrule{0.8pt}{1pt}{1pt}
        \multirow{2}{*}{Model} & \multicolumn{2}{c}{Hard} & \multicolumn{2}{c}{Medium} & \multicolumn{2}{c}{Easy} \\
        \cmidrule(lr){2-3} \cmidrule(lr){4-5} \cmidrule(lr){6-7}
        & CLIP $\uparrow$ & MF $\uparrow$ & CLIP $\uparrow$ & MF $\uparrow$ & CLIP $\uparrow$ & MF $\uparrow$ \\
        \specialrule{0.5pt}{1pt}{1pt}
        MotionClone & 26.8 & 39.0 & 24.8 & 37.0 & 27.8 & 39.0 \\
        ConMo       & 30.7 & 51.0 & 31.0 & 54.0 & 28.3 & 64.0 \\
        DiTFlow     & 32.1 & 54.0 & 32.5 & 67.0 & 32.1 & 66.0 \\
        RoPECraft   & 27.1 & 55.7 & 24.7 & 69.9 & 24.1 & 59.5 \\
        GWTF        & \textbf{33.6} & 56.0 & 32.7 & 69.0 & 30.8 & 74.0 \\
        \rowcolor{emerald!20} \textit{HALO} & 33.2 & \textbf{59.2} & \textbf{33.8} & \textbf{72.3} & \textbf{31.4} & \textbf{75.5} \\
        \specialrule{0.8pt}{1pt}{1pt}
        \end{tabular}
    }
    \end{minipage}
    \hfill
    \begin{minipage}[t]{0.45\textwidth}
        \centering
        \captionof{table}{Analysis of head configuration and selective injection entropy threshold. (a) Different attention heads. (b) Validation of selective structural head injection entropy range $\tau$.}
        \label{tab:analysis_ablation}
        \vspace{-15pt}
        
        % (a) Attention Head
        \begin{minipage}[t]{0.5\linewidth}
            \centering
            \subcaption{Attention Head}
            \label{tab:analysis_a}
            \resizebox{0.9\linewidth}{!}{%
                % \begin{tabular}{c|cc}
                \begin{tabular}{l cc}
                \specialrule{0.8pt}{1pt}{1pt}
                Head & MF $\uparrow$ & FTD $\downarrow$ \\
                \specialrule{0.5pt}{1pt}{1pt}
                All & 63.1 & 21.6 \\
                Spatial & 53.6 & 22.7 \\
                \rowcolor{emerald!20}\textit{Motion} & 66.2 & 19.4 \\
                \specialrule{0.8pt}{1pt}{1pt}
                \end{tabular}
            }
        \end{minipage}
        \hfill
        % (b) Entropy Range
        \begin{minipage}[t]{0.45\linewidth}
            \centering
            \subcaption{Entropy Range}
            \label{tab:analysis_b}
            \resizebox{0.9\linewidth}{!}{%
                % \begin{tabular}{c|cc}
                \begin{tabular}{l cc}
                \specialrule{0.8pt}{1pt}{1pt}
                $\tau$ & MF $\uparrow$ & FTD $\downarrow$ \\
                \specialrule{0.5pt}{1pt}{1pt}
                $7<$ & 54.1 & 23.5 \\
                \rowcolor{emerald!20}$<7$ & 66.2 & 19.4 \\
                \specialrule{0.8pt}{1pt}{1pt}
                \end{tabular}
            }
        \end{minipage}
   
    % ========== 오른쪽: Table 5 (User Study) ==========
    \end{minipage}
\end{table*}

% \hfill
% \begin{minipage}[t]{0.33\linewidth}
% \centering
% \subcaption{Bias Strength}
% \label{tab:analysis_b}
% \setlength{\tabcolsep}{11pt}
% \resizebox{\linewidth}{!}{%
% \begin{tabular}{c|cc}
% \specialrule{0.8pt}{1pt}{1pt}
% $\tau$ & MF $\uparrow$ & FTD $\downarrow$ \\
% \specialrule{1pt}{1pt}{1pt}
% $7<$ & 54.1 & 23.5 \\
% \rowcolor{emerald!20}$<7$ & 66.2 & 19.4 \\
% % 40\% & \textbf{0.813} & \textbf{0.314} \\
% % 80\% & 0.803 & 0.311 \\
% % 100\% & 0.799 & 0.312 \\
% \specialrule{0.8pt}{1pt}{1pt}
% \end{tabular}
% }
% \end{minipage}%
% \hfill
% % ========== (c) Optimization steps (세 번째 미니페이지) ==========
% \begin{minipage}[t]{0.3\linewidth}
% \centering
% \subcaption{Optimization steps}
% \label{tab:ablation_c}
% \begin{tabular}{c|cc}
% \specialrule{0.8pt}{1pt}{1pt}
% $K_{\text{opt}}$ & MF $\uparrow$ & IQ $\uparrow$ \\
% \specialrule{0.5pt}{1pt}{1pt}
% 1 & 0.769 & 0.318 \\
% 5 & 0.797 & 0.313 \\
% 10 & \textbf{0.803} & \textbf{0.313} \\
% \specialrule{0.8pt}{1pt}{1pt}
% \end{tabular}
% \end{minipage}

\vspace{1mm}
\noindent\textbf{Validation of Structure-Specialized Heads.}
\nametab{}~\ref{tab:analysis_b} presents results under different entropy thresholds for structure-specialized head selection.  
Based on our analysis showing a median entropy of 7, we adopt $\tau{=}7$ as the threshold. 
Injecting features from high-entropy heads ($\tau{>}7$) yields lower MF and FTD scores, indicating that such heads contain weaker, more diffuse structural cues.  
This validates the relationship between attention-map entropy and structural information in the context of motion transfer.  
Additional experiments over a wider entropy range and qualitative examples are provided in Supp.\ Sec.\ C.1.

To further assess the applicability of structure-specialized heads, we extend their utility to video editing, a task where preserving structural integrity is crucial. 
Specifically, we perform the attention value injection within these selected structure-specialized heads.
As shown in \namefig{}~\ref{fig:video_edit}(b), our approach successfully modifies subject identity (e.g., rhino$\rightarrow$lion/bear) while maintaining the original layout and structural details. 
These results demonstrate that structure-specialized heads serve as a precise mechanism for structure-preserving video editing, extending their effectiveness beyond motion transfer. 

\vspace{-1mm}
% \noindent\textbf{Scalability and Limitations.}
% We further examine the scalability of \namemethod{} under longer and higher-resolution generation settings.
% Since SCR and SRW operate on latent-space attention and displacement representations, the additional computation remains moderate compared to the overall denoising process.
% The selected structure-specialized heads also maintain consistent structural patterns over extended sequences, supporting long-horizon structural preservation.
% Nevertheless, \namemethod{} is better suited to object-level and medium-scale motion than subtle non-rigid local deformations, such as fine facial expressions, because its motion representation is based on patch-wise displacement. 
% Qualitative results for long-sequence generation, high-resolution generation, and limitation cases are provided in Supp.\ Sec.\ ??.

\vspace{-7pt}

\section{Conclusion}
In this work, we demonstrate for the first time that video DiTs contain distinct motion-specific and structure-specialized heads responsible for temporal dynamics and spatial organization. 
Based on this insight, we present \namemethod{}, a training-free motion transfer framework that leverages these heads for high motion fidelity and structural alignment. 
% By integrating semantic correspondence with motion specific representations and introducing a head-wise injection strategy, our method achieves motion-consistent and structure-aligned transfer. 
Our method integrates semantic correspondences with motion representations and proposes a head-wise injection to enable motion-consistent, structure-aligned transfer.
Extensive experiments validate that \namemethod{} outperforms prior approaches, highlighting the effectiveness of head-level analysis.
Furthermore, by introducing our Movie Scene Dataset, we provide a new direction for future research in practical motion transfer.

\vspace{1mm}
\noindent\textbf{Future Work.}
Our findings reveal that video DiTs inherently disentangle motion and structure across attention heads, opening new directions for controllable video generation. 
Future research may explore explicit control over motion direction and intensity for fine-grained editing.

\label{sec:conclusion}

\section*{Acknowledgments}
This work was supported in part by the IITP RS-2024-00457882 (AI Research Hub Project), IITP 2020-II201361, NRF RS-2024-00345806, and NRF RS-2023-002620, while the authors are affiliated with the Department of Artificial Intelligence (S.J., J.P, S.J.H.) and the Department of Computer Science (K.C.). This work was supported by the AI Seoul Tech Research Support Program of the Seoul Future Foundation.
We would like to express our sincere gratitude to RASCA FX for providing access to the movie dataset used in this work. We also thank the RASCA FX team for their valuable support and collaboration throughout this project.

% \clearpage


% ---------------------------------------------------------------
% Supplementary Material
% ---------------------------------------------------------------

\clearpage
\appendix

% Avoid duplicate hyperref anchors between main paper sections and appendix sections.
\renewcommand{\theHsection}{supp.\Alph{section}}
\renewcommand{\theHsubsection}{supp.\Alph{section}.\arabic{subsection}}
\renewcommand{\theHsubsubsection}{supp.\Alph{section}.\arabic{subsection}.\arabic{subsubsection}}

\begin{center}
    {\LARGE \bfseries Controlling Motion Transfer in Diffusion Transformers via Attention Heads \par}
    \vspace{2mm}
    {\Large \bfseries Supplementary Material \par}
    \vspace{4mm}

    {\large
    Sunyoung Jung$^{1*}$ \quad
    Jiwoo Park$^{1,2*}$ \quad
    Yoonseok Choi$^{1}$ \quad
    Kyobin Choo$^{1}$ \quad
    Ming-Hsuan Yang$^{3}$ \quad
    Seong Jae Hwang$^{1}$
    \par}

    \vspace{1.5mm}
    {\small
    $^{1}$Yonsei University \quad
    $^{2}$LG Electronics \quad
    $^{3}$University of California, Merced
    \par}
\end{center}

\vspace{2mm}

\begin{center}
    {\Large \bfseries Contents \par}
\end{center}

\vspace{-2mm}

\noindent
A \quad Code \& Presentation \dotfill \pageref{sec:supp-code}\\
B \quad Details of Problems in Previous Methods \dotfill \pageref{sec:supp-problems}\\
C \quad Details of \namemethod{} \dotfill \pageref{sec:supp-details}\\
D \quad Application to the Video DiT Model: Wan \dotfill \pageref{sec:wan}\\
E \quad Movie Scene Dataset \dotfill \pageref{sec:supp-movie}\\
F \quad Additional Experiment Results \dotfill \pageref{sec:supp-additional}\\
G \quad Discussion \dotfill \pageref{sec:supp-discussion}

% \clearpage

\section{Code \& Presentation}
% The code for \namemethod{} can be found in the \texttt{code.zip} file submitted as supplementary material. 
The code for \namemethod{} will be released to the public.
In addition, we have prepared a presentation video that introduces \namemethod{} in a clear and accessible manner, illustrating qualitative results and representative video demonstrations. This video is included as the file \texttt{presentation.mp4}.
\label{sec:supp-code}

\section{Details of Problems in Previous Methods}
\label{sec:supp-problems}
This section discusses the limitations of existing DiT-based motion transfer approaches~\cite{gowithflow, ropecraft, conmo}, which largely arise from the lack of understanding of the functional roles of attention heads. We further examine challenges observed in cross-frame attention-based methods~\cite{ditflow}. Although displacement maps can capture motion dynamics, they do not encode semantic correspondence, often resulting in motion being transferred to irrelevant regions and causing inconsistent or even reversed object motion. 
% These shortcomings highlight the necessity of integrating semantic correspondence refinement within our framework.

%----------------------------
\begin{figure}[t!]
    \centering
    % --- 왼쪽: Histogram (Fig 5) ---
    \begin{minipage}[t]{0.49\columnwidth}
        \centering
        \includegraphics[width=\linewidth]{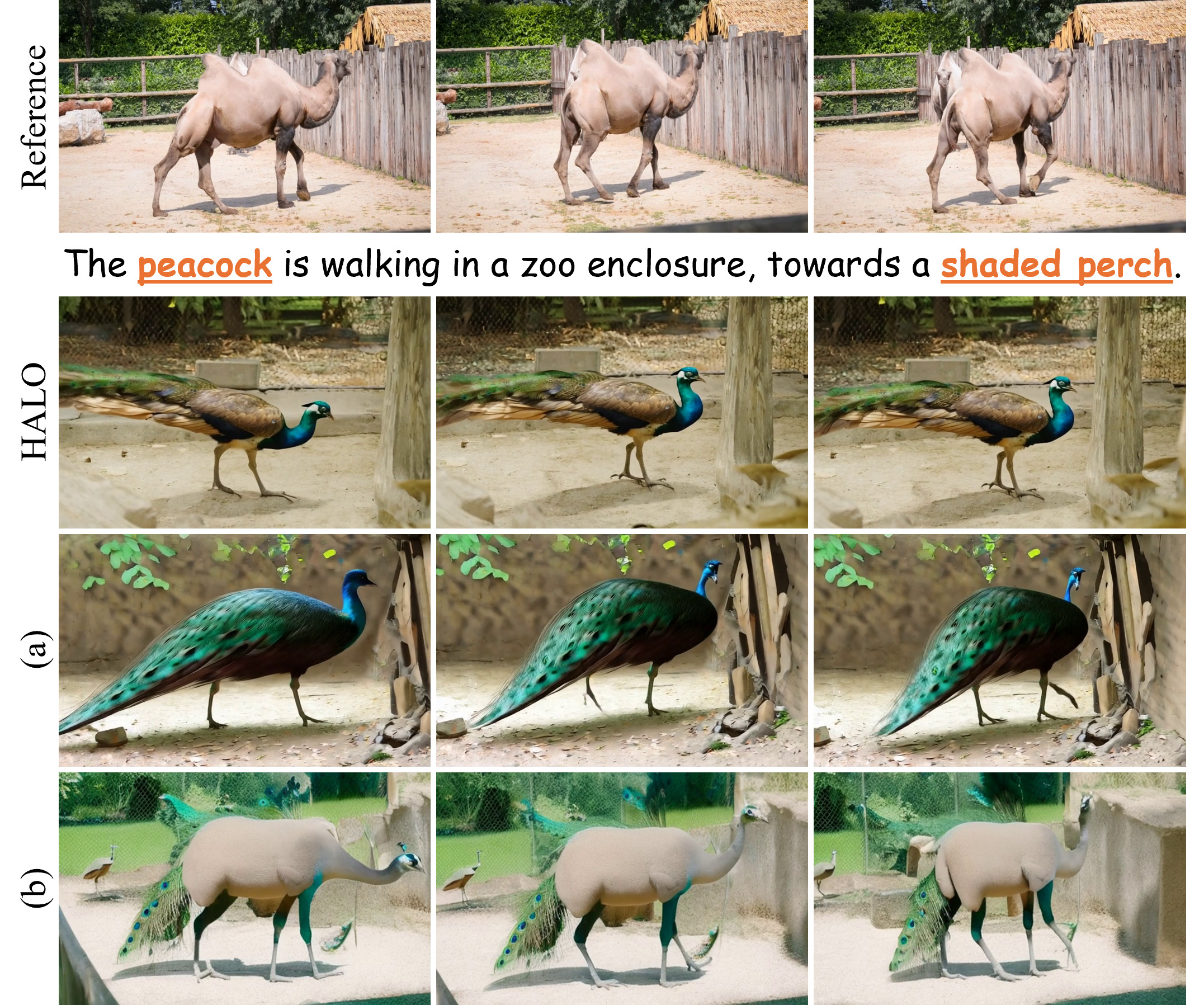}
        \caption{
        Limitations of existing DiT-based motion transfer methods.
        (a) The generated subject exhibits structural distortion and misalignment, failing to preserve the spatial layout of the reference.
        (b) Identity cues from the reference video leak into the generated subject, causing identity corruption and appearance inconsistency across frames.
        }
        \label{fig:problem1}
    \end{minipage}
    \hfill % 두 이미지 사이 간격 확보
    % --- 오른쪽: Saliency/Entropy (Fig 7) ---
    \begin{minipage}[t]{0.48\columnwidth}
        \centering
        \includegraphics[width=\linewidth]{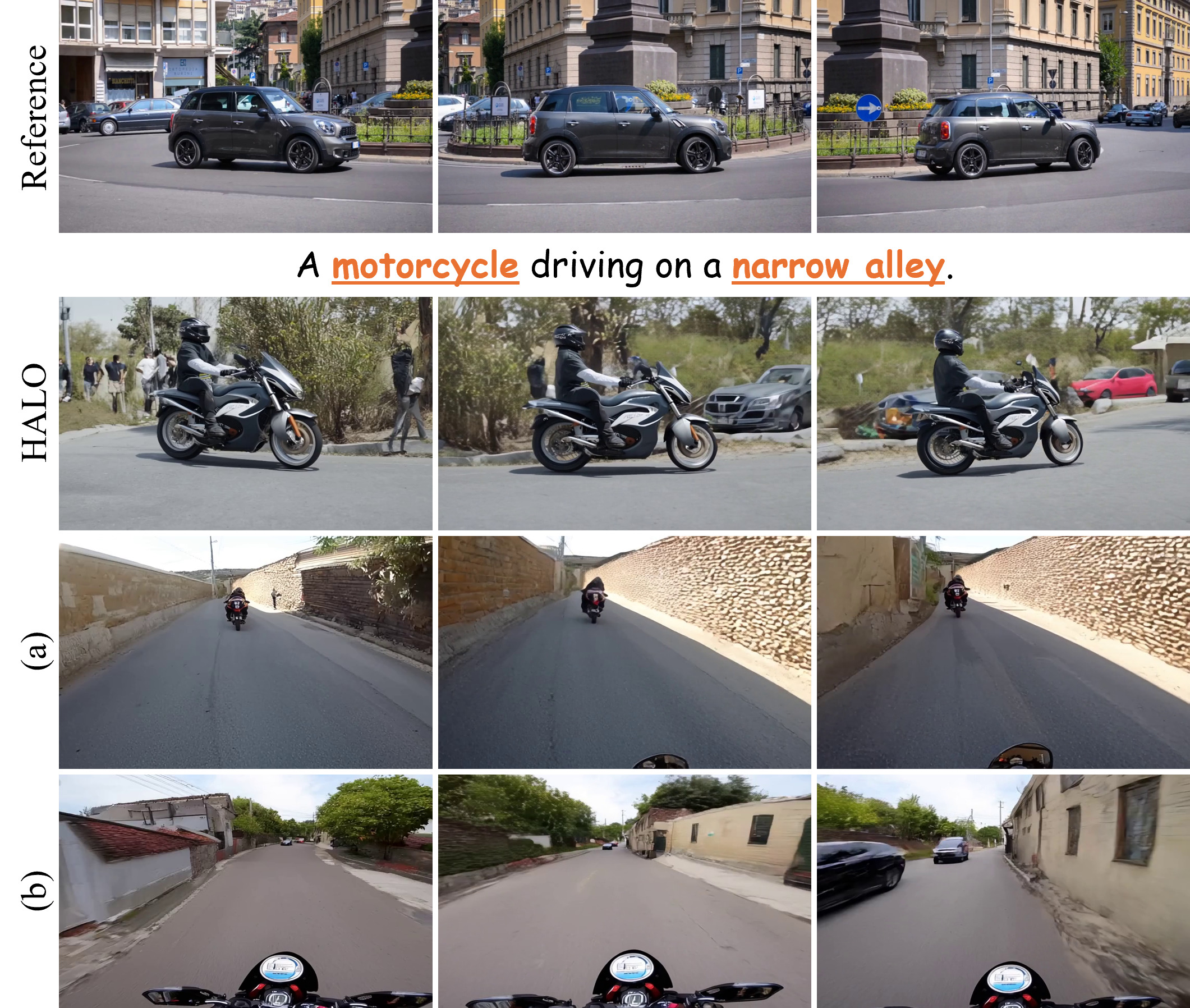}
        \caption{
        Problems of cross-frame attention-based motion transfer methods.
        (a) The generated video reproduces the reference camera motion, but the object fails to follow the reference object's trajectory, leading to degraded motion fidelity.
        (b) The generated video omits the moving object, and its motion is incorrectly mapped onto the background.
        }
        \label{fig:problem2}
    \end{minipage}
    \vspace{-10pt} % 하단 본문과의 간격 조절
\end{figure}
%----------------------------

\vspace{1pt}
\noindent\textbf{Results of DiT-based Methods.}
We analyze the outputs of prior DiT-based motion transfer approaches and identify several characteristic failure patterns. As illustrated in \namefig{}\ref{fig:problem1}(a), existing methods often produce misaligned or distorted object structures: although the peacock roughly follows the reference motion trajectory, it suffers from severe morphological inconsistency. 
Furthermore, \namefig{}\ref{fig:problem1}(b) reveals identity leakage from the reference video. In motion transfer, it is essential to preserve the semantics specified by the target prompt while aligning with the reference motion. However, previous methods inherit appearance cues from the reference, indicating insufficient disentanglement between content and motion representations.
In contrast, \namemethod{} achieves accurate motion transfer, faithfully following the reference motion and adhering to the intended target semantics.

\vspace{1pt}
\noindent\textbf{Results of Cross-Frame Attention-based methods.}
Cross-frame attention-based motion transfer approaches~\cite{ditflow} suffer from a fundamental limitation: because they derive correspondence solely from query-key similarity, the resulting displacement maps tend to reflect coarse attention activation patterns rather than semantic alignment.
As shown in \namefig{}\ref{fig:problem2}(a), even when the global camera motion correctly mirrors the reference (e.g., a right-to-left sweep), the generated object often follows an incoherent spatial trajectory, causing the layout to diverge from the reference. 
This discrepancy arises when spurious attention activations dominate displacement estimation in semantically irrelevant regions.
\namefig{}\ref{fig:problem2}(b) illustrates an additional failure mode: the foreground object disappears while its motion is incorrectly transferred to the background. 
In this case, the intended motion of the foreground subject is improperly transferred to the background, leading to degraded motion alignment.

These shortcomings highlight the necessity of integrating semantic correspondence refinement within the motion transfer pipeline.
Therefore, \namemethod{} achieves precise and coherent synthesis by grounding motion guidance in semantic alignment rather than coarse feature similarity.
To the best of our knowledge, \namemethod{} represents the first attempt to explicitly bridge the gap between high-level semantic correspondence and low-level attention-based motion representations.

\section{Details of \namemethod{}}
\label{sec:supp-details}
We provide the hyperparameter configurations in \namesec{}~\ref{hyperparameter}. 
Our extensive experiments demonstrate that \namemethod{} is highly robust to hyperparameter variations, maintaining consistency across diverse videos and model architectures without the need for specific tuning.
The full algorithmic procedure for \namemethod{} is provided in \namesec{}~\ref{algorithm}.

% We also apply the same hyperparameter setting to a different base model (Wan), suggesting robustness across backbones.

% \input{table/supp_topk}

\subsection{Hyperparameters}
\label{hyperparameter}

\vspace{1pt}
\noindent\textbf{Optimization iteration.}
\begin{table}[t]
    \centering
    % 첫 번째 표 (왼쪽)
    \begin{minipage}[t]{0.48\columnwidth}
        \centering
        \caption{Optimization steps $T_{\text{opt}}$ results.}
        \label{tab:supp_optstep}
        \setlength{\tabcolsep}{10pt}
        \resizebox{\linewidth}{!}
        {
        \begin{tabular}{cccc}
            \specialrule{0.8pt}{1pt}{1pt}
            $T_{\text{opt}}$ & CLIP $\uparrow$ & MF $\uparrow$ & FTD $\downarrow$ \\
            \specialrule{1pt}{1pt}{1pt}
            10  &  30.9  & 50.1 & 26.2 \\
            \rowcolor{emerald!20} 12 & \textbf{31.7} & \textbf{66.2} & \textbf{19.4} \\
            15 &  30.5  & 56.0 & 22.7  \\
            \specialrule{0.8pt}{1pt}{1pt}
        \vspace{-20pt}
        \end{tabular}
        }
    \end{minipage}
    \hfill % 두 표 사이의 간격을 최대한 벌림
    % 두 번째 표 (오른쪽)
    \begin{minipage}[t]{0.48\columnwidth}
        \centering
        \caption{Bias strength $\beta$ analysis results.}
        \label{tab:supp_bias}
        % \vspace{5pt}
        \setlength{\tabcolsep}{10pt}
        \resizebox{\linewidth}{!}
        {
        \begin{tabular}{cccc}
            \specialrule{0.8pt}{1pt}{1pt}
            $\beta$ & CLIP $\uparrow$ & MF $\uparrow$ & FTD $\downarrow$ \\
            \specialrule{1pt}{1pt}{1pt}
            \rowcolor{emerald!20} 0.1  &  \textbf{31.7} & \textbf{66.2} & \textbf{19.4} \\
            0.3 &  30.7  & 53.6 & 23.4 \\
            0.5 &  30.7  & 53.9 & 23.7\\
            \specialrule{0.8pt}{1pt}{1pt}
            \rule{0pt}{2.5ex} % 행 높이 맞춤용 (표 높이가 다를 때 사용)
        \vspace{-20pt}
        \end{tabular}
        }
    \end{minipage}
\end{table}
\begin{table}[t!]
    \centering
    \caption{Hyperparameter sensitivity.}
    
    % ---------------- (a) ----------------
    \begin{minipage}[t]{0.35\columnwidth}
        \centering
        \resizebox{\linewidth}{!}{%
            \begin{tabular}{lcccccc}
            \specialrule{0.8pt}{1pt}{1pt}
             & 4 & 5 & 6 & 7 & 8 & 9 \\
            \specialrule{1pt}{1pt}{1pt}
            MF & 67.0 & 67.2 & 67.1 & 72.3 & 70.2 & 69.1 \\
            \specialrule{0.8pt}{1pt}{1pt}
            \multicolumn{7}{c}{\textbf{(a)} $\tau$ sensitivity} \\
            \end{tabular}
        }
    \end{minipage}
    \hfill
    % ---------------- (b) ----------------
    \begin{minipage}[t]{0.2\columnwidth}
        \centering
        \resizebox{\linewidth}{!}{%
            \begin{tabular}{lccc}
            \specialrule{0.8pt}{1pt}{1pt}
              & 0.01 & 0.05 & 0.1 \\
            \specialrule{1pt}{0.8pt}{0.8pt}
            MF & 66.6 & 68.1 & 69.8 \\
            \specialrule{0.8pt}{1pt}{1pt}
            \multicolumn{4}{c}{\textbf{(b)} $\beta$ sensitivity} \\
            \end{tabular}
        }
    \end{minipage}
    \hfill
    % ---------------- (c) ----------------
    \begin{minipage}[t]{0.22\columnwidth}
        \centering
        \resizebox{\linewidth}{!}{%
            \begin{tabular}{lccc}
           \specialrule{0.8pt}{1pt}{1pt}
              & 3 & 4 & 5 \\
            \specialrule{1pt}{0.8pt}{0.8pt}
            MF & 75.4 & 76.1 & 73.1 \\
            \specialrule{0.8pt}{1pt}{1pt}
            \multicolumn{4}{c}{\textbf{(c)} top-$k$ sensitivity} \\
            \end{tabular}
        }
    \end{minipage}

    \vspace{-2mm}
    \label{tab:hyper_abc}
\end{table}
We set the optimization steps to $T_{\text{opt}}{=}12$. 
Increasing $T_{\text{opt}}$ improves quantitative performance, especially MF, but yields diminishing returns after $12$ iterations (see \nametab{}~\ref{tab:supp_optstep}).
We therefore choose $T_{\text{opt}}{=}12$ to balance accuracy and efficiency.

\vspace{1pt}
\noindent\textbf{Entropy range.}
Beyond the broad range discussed in the main text, \nametab{}~\ref{tab:hyper_abc}(a) provides a fine-grained entropy range $\tau$ analysis on the curated dataset. MF degrades when the value deviates from 7, indicating that an entropy threshold of 7 optimally distinguishes structural information.

\vspace{1pt}
\noindent\textbf{Bias strength.}
Semantic bias strength $\beta$ scales the correspondence-guided bias on displacement maps in Semantic Reweighting (SRW). \nametab{}~\ref{tab:supp_bias} shows that a small bias strength yields the best overall performance, whereas larger values lead to degraded results. 
These results indicate that an excessive semantic bias can suppress motion dynamics in the displacement maps, thereby compromising the motion fidelity.
Based on this observation, we set $\beta{=}0.1$ in all experiments. 
A finer-grained sweep on the curated dataset further shows stable performance within the low-bias range, with $\beta{=}0.1$ achieving the best result, as shown in \nametab{}~\ref{tab:hyper_abc}(b).
% Consequently, our hyperparameter settings are carefully calibrated through empirical evaluations and analysis to strike an optimal balance.

\vspace{1pt}
\noindent\textbf{Top-$k$ Selection.}
To construct the reference displacement map, 
% \vspace{}
\begin{wraptable}{r}{0.5\textwidth} % r: 우측 배치, 0.5: 본문 너비의 절반 차지
    \centering
    \vspace{-25pt} % 표 위쪽 여백 조절
    \caption{Hyperparameter Top-$k$ analysis.}
    \label{tab:supp_topk}
    \vspace{3pt} % 캡션과 표 사이 간격 조절
    \setlength{\tabcolsep}{8pt}
    \resizebox{0.8\linewidth}{!}
    {%
        \begin{tabular}{cccc}
            \specialrule{0.8pt}{1pt}{1pt}
             Top-$k$ & CLIP $\uparrow$ & MF $\uparrow$ & FTD $\downarrow$ \\
                \specialrule{1pt}{1pt}{1pt}
                \rowcolor{emerald!20} 4 & \textbf{31.7} & \textbf{66.2} & \textbf{19.4} \\
                10  &  31.2  & 45.2 & 26.5 \\
                 % &  30.9 & 89.0 & 89.4 & 91.9  \\
                20 &  30.8  & 39.6 & 30.3  \\
                \specialrule{0.8pt}{1pt}{1pt}
        \end{tabular}
    }
    \vspace{-10pt} % 표 아래쪽 여백 조절
\end{wraptable}
we extract the top-$4$ candidates from cross-frame attention and then select the final match by comparing their distances to the semantically best-aligned location. 
To justify the choice of $k{=}4$, we additionally evaluate larger candidates ($k{=}10$, $k{=}20$). 

\begin{figure}[t!]
    \centering
    \includegraphics[width=\columnwidth]{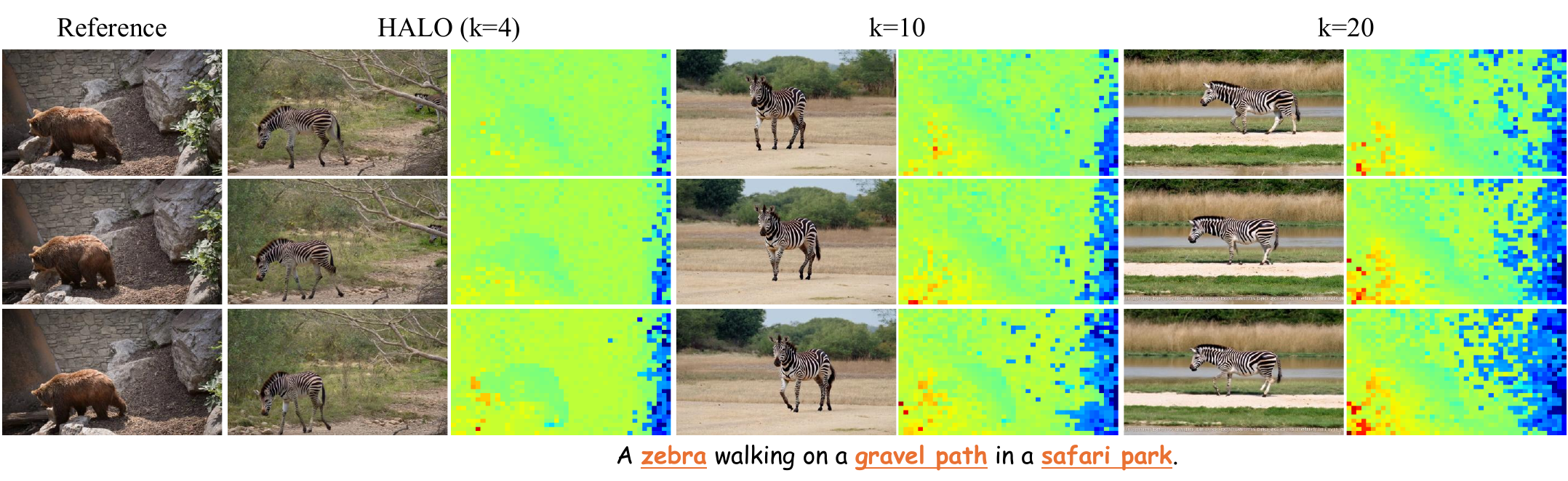}
    \caption{
    Qualitative results with corresponding reference displacement maps $\mathcal{D}_{\text{ref}}$ in hyperparameter Top-$k$ analysis.
    }
    \vspace{-8pt}
    \label{fig:topk}
\end{figure}

As shown in \nametab{}~\ref{tab:supp_topk}, increasing $k$ consistently degrades all metrics, indicating that using the top $4$ candidates yields the most stable and reliable displacement estimation.
This performance drop arises from inaccurate displacement predictions caused by irrelevant cross-frame attention candidates. 
As illustrated in \namefig{}\ref{fig:topk}, while \namemethod{} with $k{=}4$ produces clean, object-aligned displacement maps, larger $k$ settings ($k{=}10,20$) generate noisy and spatially inconsistent maps.
These inaccuracies propagate directly into motion transfer, ultimately leading to weaker motion alignment.
Additionally, evaluating the performance sensitivity within a narrower top-$k$ range (3, 4, and 5) yields consistent results (\nametab{}~\ref{tab:hyper_abc}(c)), demonstrating low variance and robustness to this specific setting.

% \begin{wraptable}{r}{0.5\textwidth} % r: 우측 배치, 0.5: 본문 너비의 절반 차지
%     \centering
%     \vspace{-45pt} % 표 위쪽 여백 조절
%     \caption{Hyperparameter bias strength $\beta$ analysis results.}
%     \label{tab:supp_bias}
%     % \vspace{3pt} % 캡션과 표 사이 간격 조절
%     % \resizebox{0.7\linewidth}{!}
%     \setlength{\tabcolsep}{5pt}
%     {%
%         \begin{tabular}{cccc}
%             \specialrule{0.8pt}{1pt}{1pt}
%                 $\beta$ & CLIP $\uparrow$ & MF $\uparrow$ & FTD $\downarrow$ \\
%                 \specialrule{1pt}{1pt}{1pt}
%                 \rowcolor{emerald!20} 0.1  &  -  & 75.0 & - \\
%                 0.3 &  - & 71.8 & - \\
%                 0.5 &  -  & 70.9 & -  \\
%                 \specialrule{0.8pt}{1pt}{1pt}
%         \end{tabular}
%     }
%     \vspace{-20pt} % 표 아래쪽 여백 조절
% \end{wraptable}
% , where a larger $\beta$ can over-constrain attention and reduce motion. 

\begin{algorithm}[t!]
\caption{\textit{HALO} inference process} \label{algori_1}
\begin{spacing}{1.05}
\begin{small}
\textbf{Input:} Reference video $x_{\text{ref}}$, text prompt $\mathcal{P}$, \\ 
\phantom{\textbf{Input:} } DiT model $\epsilon_{\theta}$, noise scheduler $\mathcal{N}$, encoder $\mathcal{E}$ decoder $\mathcal{G}$ \\ 
\textbf{Output:} Generated video $x_{0}$ \\
\textbf{Note:} Semantic extractor \textit{DIFT}, Attention head $h$ \\ 
\phantom{\textbf{Note:} } $\mathcal{T}, \mathcal{S}$ : Motion Heads, Structure Heads 
\begin{algorithmic}[1]
\State \text{Extract correspondence map:} $\mathcal{C}\xleftarrow{}\textit{DIFT}$ \Comment{cosine similarity}
\State \text{Compute attention:} $\{Q,K\}\xleftarrow{}\epsilon_{\theta}(z_{\text{ref}},\emptyset,0)$ \Comment{$z_{\text{ref}}=\mathcal{E}(x_{\text{ref}})$}
\State Head Classification for $h\in\mathcal{T}$ \Comment{Temporal or Spatial head}
\For{each $(i, j)$ where $i, j \in [1, F]$}
    \If{$h\in\mathcal{T}$}
    \State Calculate cross-frame attention $A_{i, j}$
    \State Construct displacement matrix $\mathcal{D}_{i, j}$
    \EndIf
\EndFor
\State \text{Construct reference displacement:} $\mathcal{D}_{\text{ref}}\xleftarrow{} \text{SCR}(\mathcal{D}_{i,j}, \mathcal{C}_{i,j})$   
\State \text{Initialize} ${z_{T}} \sim \mathcal{N}(0, I)$
\For{ $t=T$ to $0$} 
    \If{$t>T_{\text{opt}}$} 
        \For{
        \text{optimization step} $k=0$ \text{to} $N_{\text{opt}}$}
        \State Head Classification for $h\in\mathcal{T}$ 
        \State Entropy Calculation for $h\in \mathcal{S}$
        \State $z'_{t}\xleftarrow{}\mathcal{N}(z_{\text{ref}},t)$ \Comment{Noise Forward}
        \If{$h\in\mathcal{S}$} \Comment{Structure Heads}
        \State Store Value: $V'\xleftarrow{}\epsilon_\theta(z'_{t},\mathcal{P},t)$
        \State Inject Value: $V'\xleftarrow{}V$
        \EndIf
        \If{$h\in\mathcal{T}$} \Comment{Motion Heads}
        \For{each $(i, j)$ where $i, j \in [1, F]$}
        \State Calculate cross-frame attention $A_{i, j}$
        \State Reweight $\tilde{A}_{i, j}$: $\tilde{A}_{i, j}\xleftarrow{}\text{SRW}(A_{i,j}, \mathcal{C}_{i,j})$
        \State Construct displacement matrix $\mathcal{D}_{i, j}$
        \EndFor
        \State $\mathcal{L}_{SM}\xleftarrow{} $$
        \left\|\mathcal{D}_{\text{ref}} - \mathcal{D}_{t}\right\|_{2}^{2}
        $$ $
        \State Update $z_t$ by minimizing $\mathcal{L}_{SM}$
        \EndIf
        \EndFor
    \EndIf
    \State $z_{t-1}=f(z_{t},\epsilon_\theta(z_t,\mathcal{P},t))$
\EndFor

\State \textbf{Return} $x_0=\mathcal{G}(z_0)$
\end{algorithmic}
\end{small}
\end{spacing}
\end{algorithm}

\subsection{Algorithm}
\label{algorithm}
\namealg{}~\ref{algori_1} outlines the inference pipeline of \namemethod{}, encompassing both the motion-optimization and the structural feature-injection steps.
Notably, this entire procedure is training-free, requiring no additional parameters or model fine-tuning.

\subsection{Metric Details for \textit{Motion-Specific Heads}}
To verify that our head-selection strategy isolates motion-specific behavior, we compare displacement fields from temporal, spatial, and all-head configurations against ground-truth optical flow (RAFT~\cite{raft}) using Directional Alignment (DA) and Correlation (Corr).
Both metrics operate on displacement maps of shape $(T, H, W, 2)$, where $T$ is the number of frame pairs used for displacement estimation, $(H, W)$ is the latent spatial resolution, and the final dimension corresponds to the $(x, y)$ flow components.

\vspace{1pt}
\noindent\textbf{Directional Alignment (DA).}
DA evaluates how well the predicted displacement directions align with the reference optical flow, independent of magnitude. 
Given the reference flow $f_{\text{ref}}$ and generated displacement $f_{\text{gen}}$, DA is defined as:
\begin{align*}
\text{DA} = \mathrm{mean}\Big\langle 
\frac{f_{\text{ref}}}{\|f_{\text{ref}}\|},
\frac{f_{\text{gen}}}{\|f_{\text{gen}}\|}
\Big\rangle,
\tag{1}
\end{align*}
where $\|\cdot\|$ denotes the L2 norm and $\langle\cdot,\cdot\rangle$ is the dot product.
The computation proceeds as follows:  
1) Normalize $f_{\text{ref}}$ and $f_{\text{gen}}$ at each pixel.  
2) Compute per-pixel cosine similarity between normalized vectors.  
3) Average the directional similarity across all spatial locations and time steps.

% \subsection{Structure-Specialized Heads}
% \label{sec:entropy}
\vspace{1pt}
\noindent\textbf{Correlation (Corr).}
Corr measures the linear relationship between the predicted and reference flows, capturing both directional and magnitude consistency. It is computed using the Pearson correlation coefficient:
\begin{align*}
\text{Corr} =
\frac{\mathrm{Cov}(f_{\text{ref}},, f_{\text{gen}})}
{\sigma_{\text{ref}} , \sigma_{\text{gen}}},
\tag{2}
\end{align*}
where $\mathrm{Cov}$ denotes covariance and $\sigma$ represents the standard deviation.
To compute Corr, we flatten each displacement field into a one-dimensional array of length $T \times H \times W \times 2$ and apply the standard Pearson correlation formula.

Whereas DA focuses solely on directional agreement, Corr reflects overall linear consistency, enabling complementary assessment of both direction- and magnitude-level motion alignment.
This multi-dimensional analysis serves as a robust proof of concept for \textit{Motion-Specific head}, proving that \namemethod{} achieves superior motion-aware guidance through a well-validated selection process.

\begin{figure*}[t!]
    \centering
    \includegraphics[width=\textwidth]{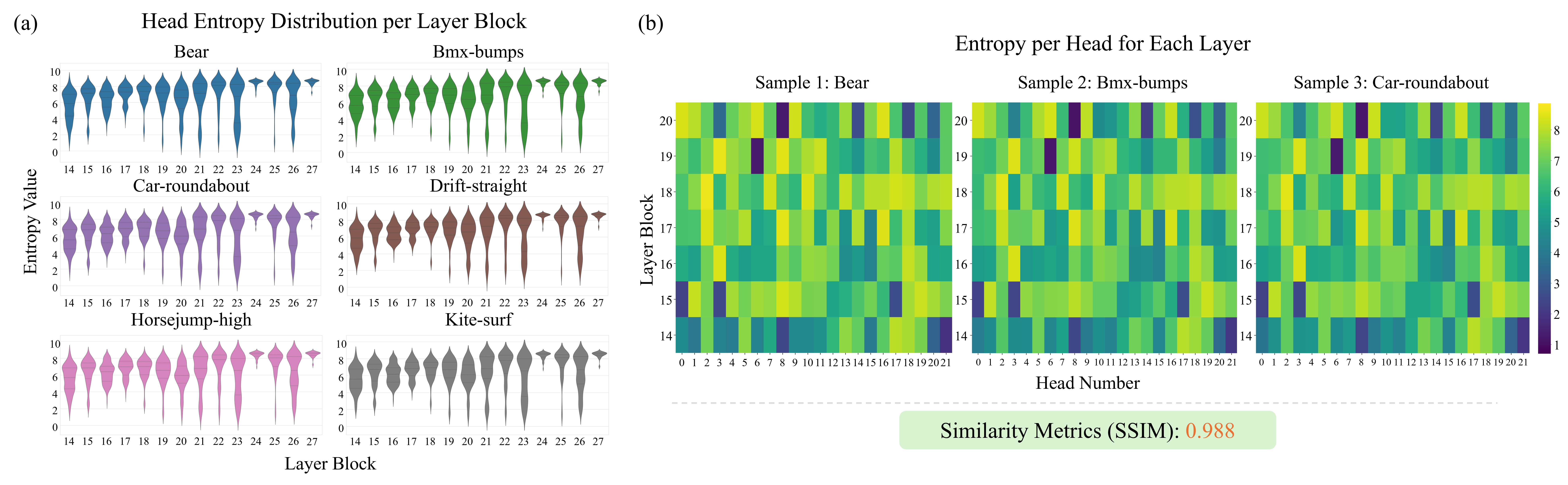}
\caption{
(a) Attention Entropy Distribution: Distribution of attention-map entropy values across samples.
(b) Layer-wise Head Entropy Consistency: Distribution of head-level entropy values across layers, showing consistent patterns across different samples.
Similarity Metric (SSIM): A heatmap is used to visualize pairwise similarity between samples.
}
    % \vspace{-8pt}
    \label{fig:entropy_analysis}
\end{figure*}

\subsection{Entropy Validation in \textit{Structure-Specialized Heads}}
% \vspace{1pt}
\noindent\textbf{Entropy Consistency Across Samples.}
We investigate the consistency of head entropy distributions across diverse video inputs. 
\namefig{}~\ref{fig:entropy_analysis}(a) shows the entropy patterns remain consistent across different samples, particularly within layers $14$ to $27$, which are utilized for structural feature injection. 
To quantify this consistency, we compute similarity metrics using heat maps that visualize entropy values across the entire benchmark (\namefig{}~\ref{fig:entropy_analysis}(b)). 
The resulting SSIM score of $0.988$ confirms a highly stable and invariant entropy pattern regardless of the input.
This high degree of similarity indicates that the structural information encoded by the low-entropy heads is inherently consistent.

\begin{figure}[t!]
    \centering
    \includegraphics[width=\columnwidth]{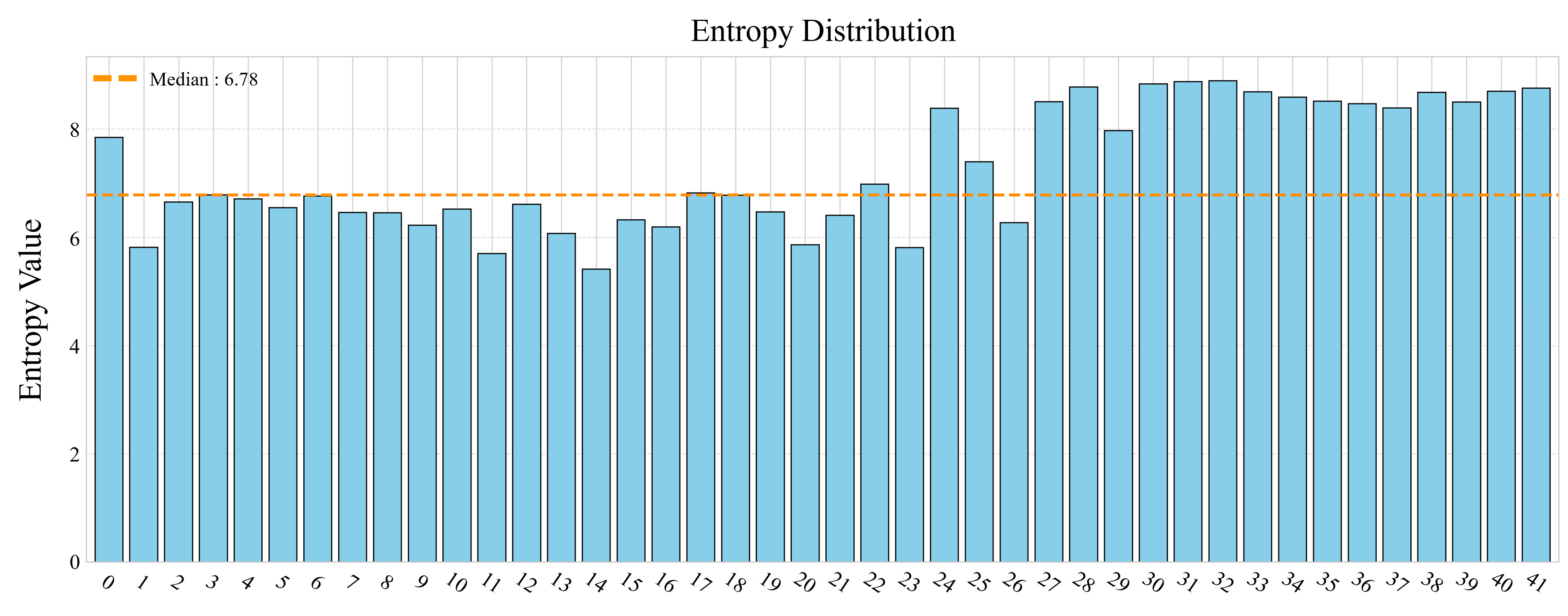}
\caption{
Entropy distribution across all layers, computed by averaging the entropy of all attention heads within each layer.
The orange dashed line denotes the median entropy value across layers, which is $6.78$.
}
    % \vspace{-8pt}
    \label{fig:Entropy_medianvalue}
\end{figure}

\vspace{1pt}
\noindent\textbf{Entropy Threshold.}
For structural-head selection, we set the entropy threshold to $\tau < 7$. This choice is motivated by our empirical observation that the median entropy of attention maps across layers is approximately 7, as shown in \namefig{}\ref{fig:Entropy_medianvalue}.
Intuitively, lower-entropy heads exhibit more concentrated attention distributions, which aligns with spatially localized, structure-preserving behavior.
To derive this distribution, we compute the entropy of attention maps across all 90 samples, 42 layers, and 50 denoising steps.
\namefig{}\ref{fig:Entropy_medianvalue} reports the mean entropy aggregated over all samples, layers, and steps, yielding a median value of roughly 7. This provides a principled basis for setting the threshold at $\tau < 7$.

\begin{figure}[t!]
    \centering
    \includegraphics[width=0.8\columnwidth]{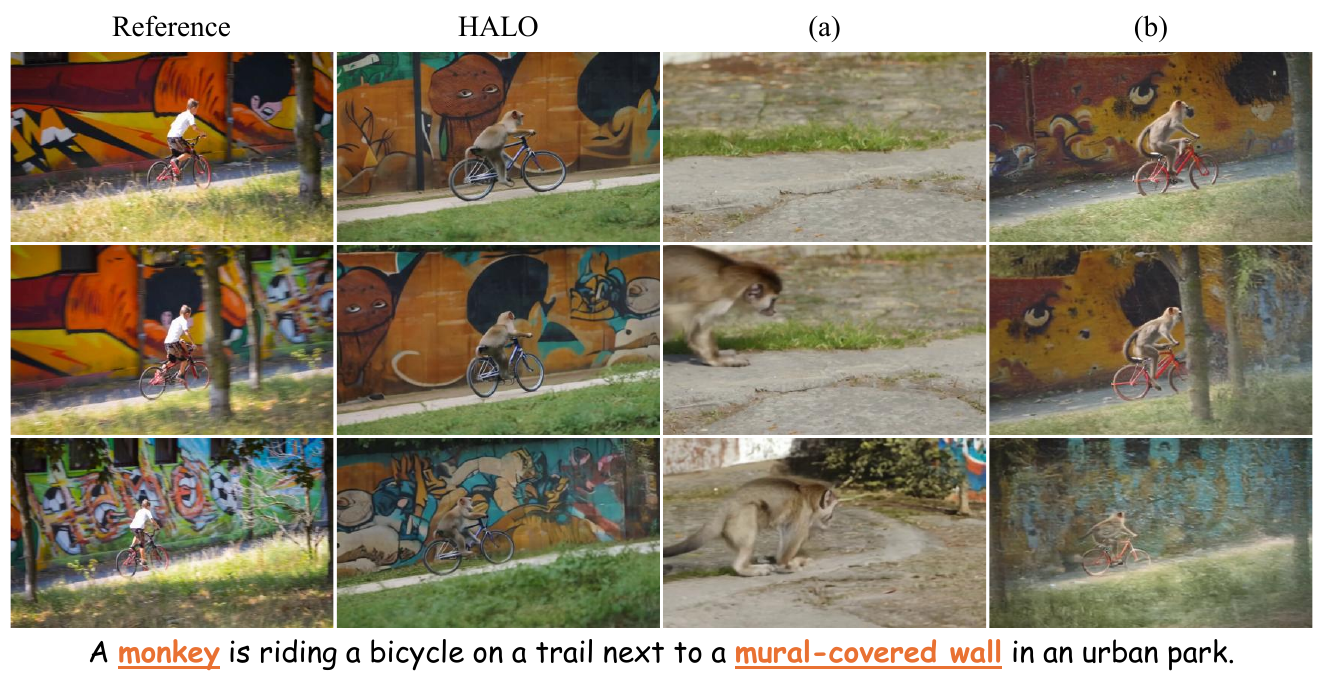}
    \caption{Qualitative results under different entropy thresholds.
    (a) Injecting features only from heads with entropy greater than 7. (b) Injecting features from all heads.}
    \vspace{-10pt}
    \label{fig:entropy_qual}
\end{figure}

% \vspace{1pt}
% \noindent\textbf{Entropy Threshold Comparison.}
We evaluate various entropy ranges to confirm that our selected entropy range, $\tau$, consistently identifies structural-specialized heads. 
% For comparison, we examine two alternative settings: injecting high-entropy heads ($\tau>7$) and an all-head baseline without thresholding.
\namefig{}~\ref{fig:entropy_qual}(a), high-entropy heads ($\tau>7$) produce diffuse attention patterns, resulting in layouts misaligned with the reference.
Meanwhile, injecting all heads (b) causes visual artifacts due to the accumulation of disparate spatial features.
These results highlight that selective structural-head injection is essential for providing clean structural guidance, allowing \namemethod{} to maintain a coherent spatial layout without artifacts.
% Attention maps in this range exhibit highly diffuse patterns, providing insufficient structure cues. %Thus, their features provide weaker structural cues for guiding generation.
% In (b), injecting all head features results in noticeable visual artifacts. 
% This stems from the unselective accumulation of disparate spatial features, which corrupts the synthesized output. 
% As features carrying various types of spatial information merge during generation, the accumulation results in outputs with artifacts. 

\subsection{Validation of \textit{Motion-Specific Heads}}
\label{sec:head}

\begin{figure*}[t!]
    \centering
    \includegraphics[width=\textwidth]{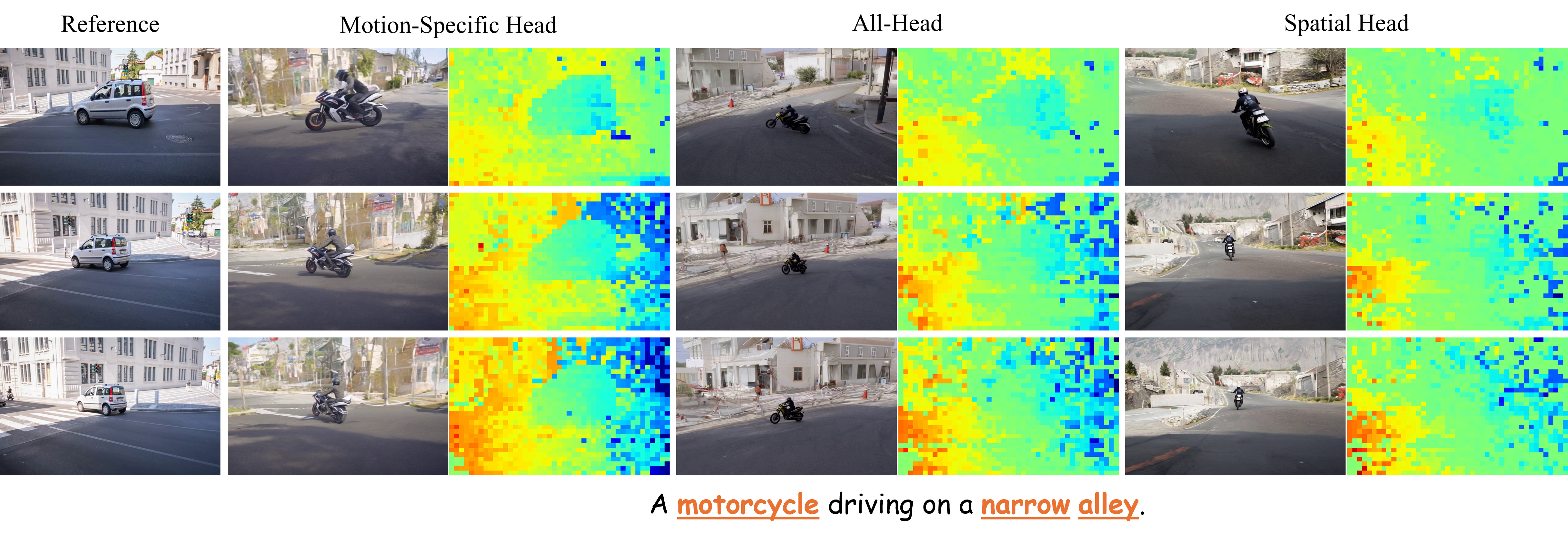}
    \caption{
    Qualitative results and reference displacement maps $\mathcal{D}_{\text{ref}}$ across various attention head configurations.
    }
    % \vspace{-8pt}
    \label{fig:head_qualitative}
\end{figure*}

\namefig{}~\ref{fig:head_qualitative} illustrates how the head analysis translates to the motion transfer results.
Motion-Specific Head demonstrates results that faithfully track the motion of the reference video while effectively capturing the movement of the car in the reference displacement map. 
In contrast, All-Head and Spatial Head fail to properly capture the car in the displacement map, leading to misaligned motion in the generated videos. 
Consequently, the results validate that \textit{Motion-Specific} Head plays a crucial role in effectively processing motion information.

% \subsection{Algorithm}
% \namealg{}\ref{algori_1} provides an overview of the inference pipeline of \namemethod{}, encompassing both the motion-optimization procedure and the structural feature–injection steps.

%\section{Additional Analysis of \namemethod{}}
%We present additional analyses of \namemethod{}.
%\namesec{}~\ref{sec:head} visualizes the head configuration displacement maps and corresponding qualitative results.
%\namesec{}~\ref{sec:entropy} provides further entropy-based analysis for structural-head selection.
%\namesec{}~\ref{sec:wan} details the application of our method to another video diffusion model, Wan~\cite{wan}, and reports the experimental results.

\section{Application to the Video DiT Model: Wan}
\label{sec:wan}

To demonstrate the generalizability of \namemethod{}, we extend it to Wan~\cite{wan} video DiT model. Notably, we utilize the identical hyperparameter configurations as those used for CogVideoX~\cite{cogvideox}, achieving consistent performance without any model-specific tuning. 
Unlike CogVideoX~\cite{cogvideox}, Wan employs a dedicated self-attention module that exclusively processes visual tokens. Accordingly, our analysis is conducted on this specialized attention mechanism.

\begin{figure}[t!]
\centering
\includegraphics[width=0.65\columnwidth]{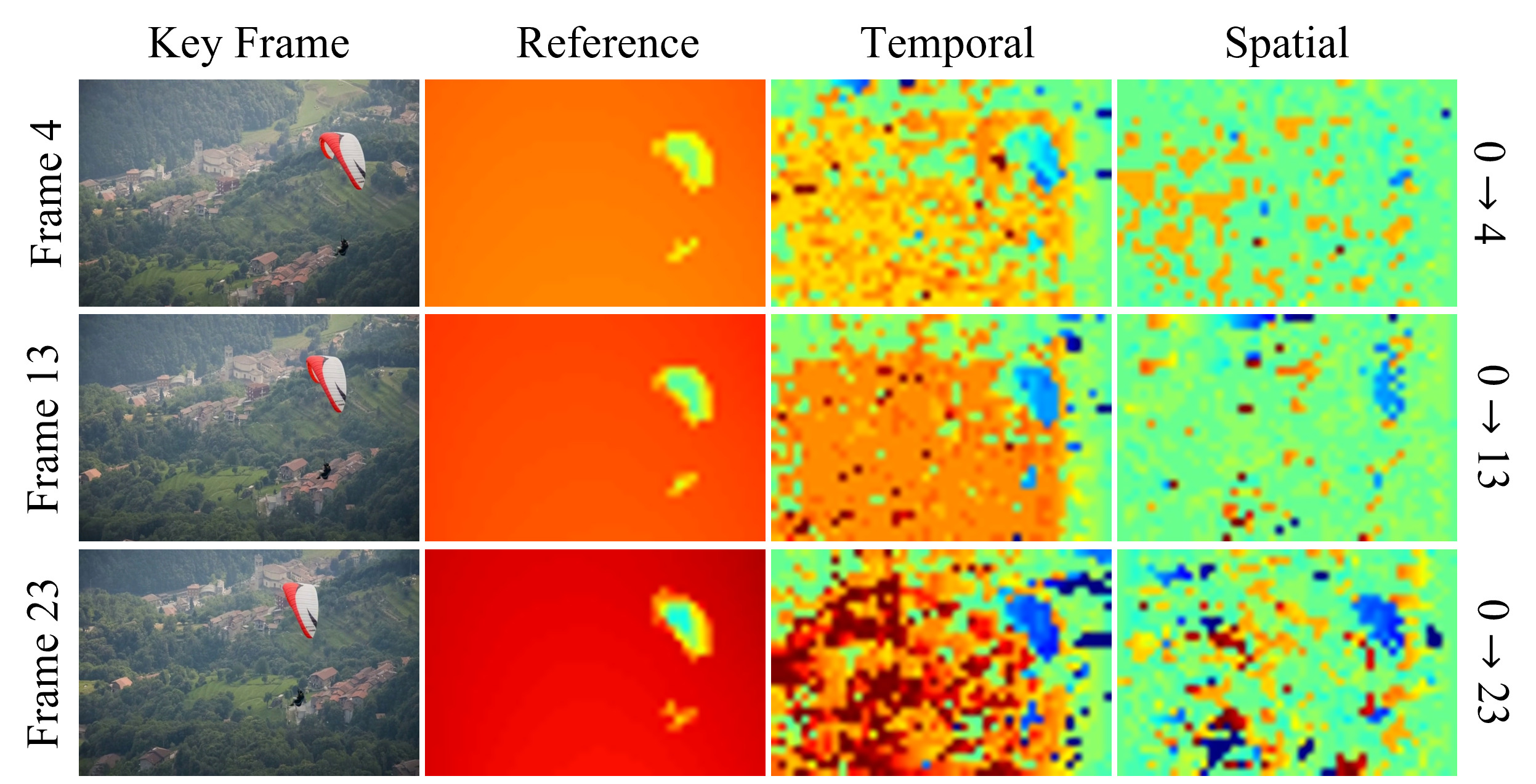}
\caption{
Displacement-field comparison between temporal and spatial heads in Wan. Temporal heads better capture reference motion, showing clearer separation of background and object motion.
}
\vspace{-10pt}
\label{fig:wan_displacement1}
\end{figure}

\begin{figure}[t!]
    \centering
    % --- 왼쪽: Histogram (Fig 5) ---
    \begin{minipage}[t]{0.49\columnwidth}
        \centering
        \includegraphics[width=\linewidth]{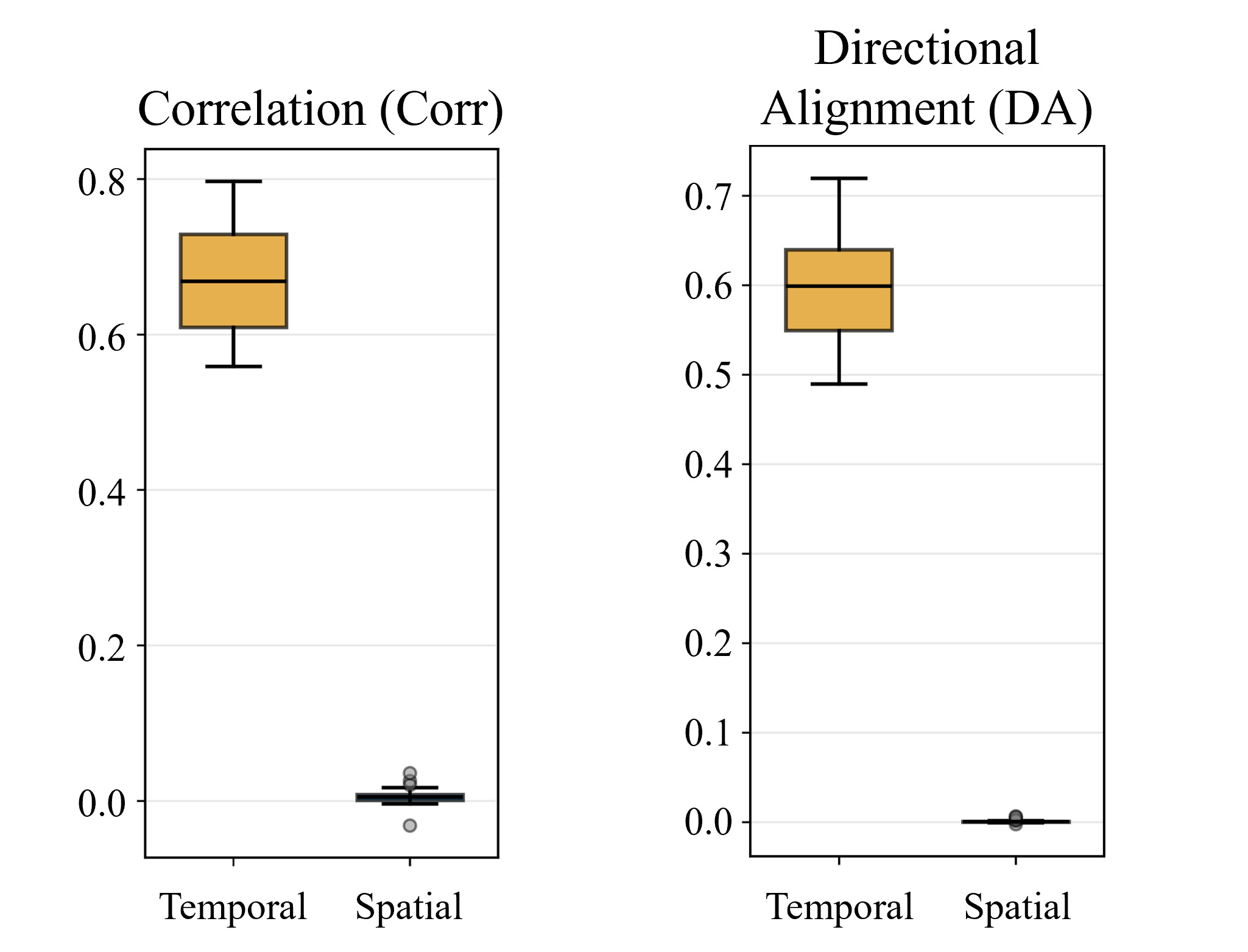}
        \caption{
        Quantitative comparison of each head in Wan using Directional Alignment (DA) and Correlation (Corr) with ground-truth optical flow.
        }
        \label{fig:wan_displacement2}
    \end{minipage}
    \hfill % 두 이미지 사이 간격 확보
    % --- 오른쪽: Saliency/Entropy (Fig 7) ---
    \begin{minipage}[t]{0.48\columnwidth}
        \centering
        \includegraphics[width=\linewidth]{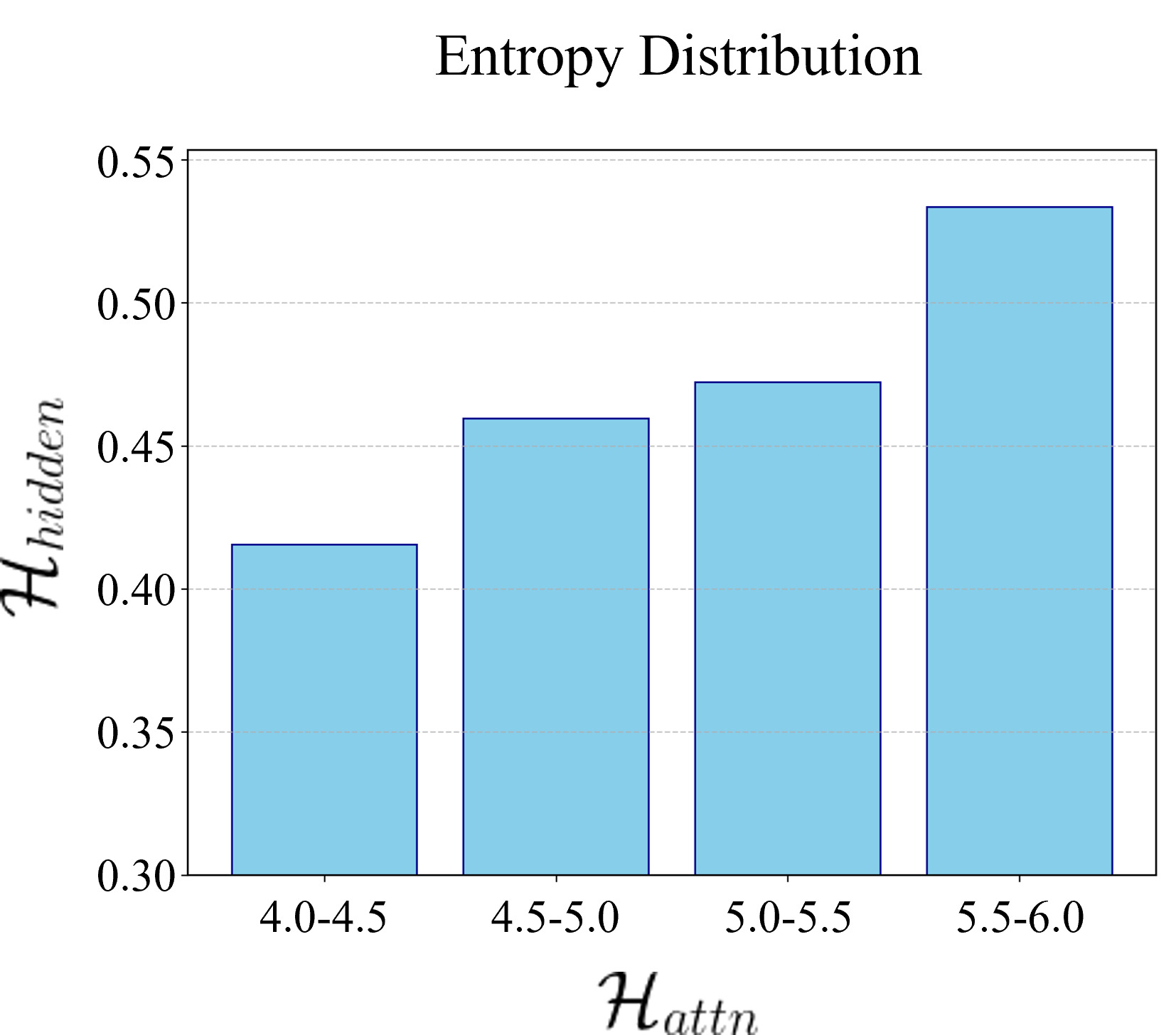}
        \caption{
        Relationship between attention-map entropy and spatial entropy in Wan, demonstrating that low-entropy heads correspond to structure-specialized heads.
        }
        \label{fig:wan_entropy}
    \end{minipage}
    \vspace{-10pt} % 하단 본문과의 간격 조절
\end{figure}

\noindent\textbf{Head Analysis in Wan.}
Using the same procedure as in the main model, we identify both Motion-Specific and Structure-Specialized heads within Wan. To analyze motion behavior, we compute displacement maps separately for the temporal and spatial heads. 
As shown in \namefig{}~\ref{fig:wan_displacement1}, temporal heads more accurately capture the reference motion than spatial heads, exhibiting clear distinctions between background motion (red) and object motion (blue).
We further compute Correlation (Corr) and Directional Alignment (DA) between RAFT-based optical flow~\cite{raft} and the displacement maps of each head. \namefig{}~\ref{fig:wan_displacement2} shows that temporal heads consistently achieve higher Corr and DA scores than spatial heads, confirming that temporal heads in Wan correspond to motion-specific heads.

%----------------------------
%----------------------------

We also compute attention-map entropy~\cite{attentropy} and compare it with the spatial entropy~\cite{spatialentropy, spatialentropy2} derived from corresponding attention features. As shown in \namefig{}~\ref{fig:wan_entropy}, attention-map entropy is strongly correlated with spatial entropy, consistent with the findings in the main paper. This confirms that low-entropy heads in Wan encode structural information.
% Furthermore, the diagonal patterns observed in the attention between visual tokens reveal structural relationships among patches. 
Together, these results demonstrate that our analysis methodology generalizes effectively across different video DiT architectures.

\begin{figure*}[t!]
\centering
\includegraphics[width=\textwidth]{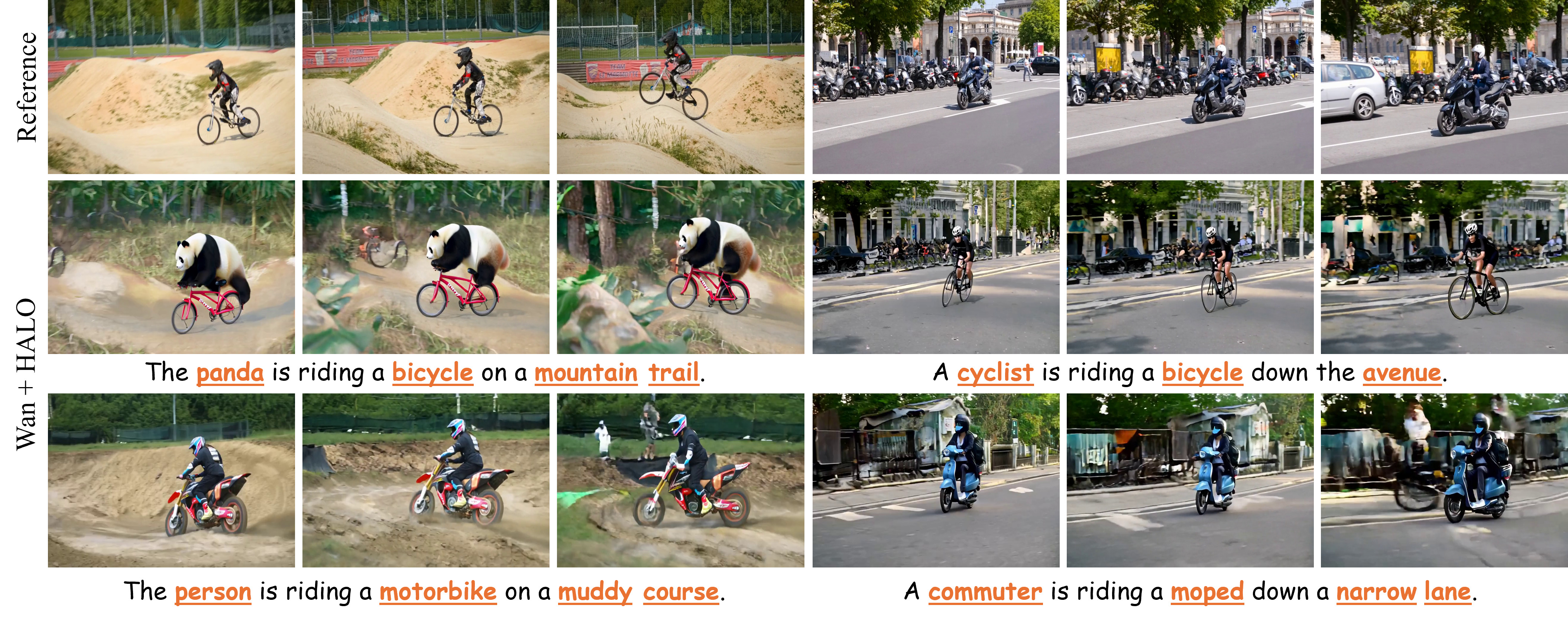}
\caption{
Motion transfer results of \namemethod{} applied to Wan.
}
\label{fig:wan_qualitative}
\end{figure*}

\vspace{1pt}
\noindent\textbf{Experimental Results.}
We conduct experiments on the same dataset~\cite{det} 
\begin{wraptable}{r}{0.5\textwidth} % r: 우측 배치, 0.5: 본문 너비의 절반 차지
    \centering
    \vspace{-10pt} % 표 위쪽 여백 조절
    \caption{Quantitative results of Wan.}
    \label{tab:supp_wan}
    \vspace{3pt} % 캡션과 표 사이 간격 조절
    \resizebox{\linewidth}{!}{%
        \begin{tabular}{l cccc}
        \specialrule{0.8pt}{1pt}{1pt}
        {Model} &  CLIP $\uparrow$ & TC $\uparrow$ & MF $\uparrow$ & FTD $\downarrow$\\
        \specialrule{1pt}{1pt}{1pt}
        WAN + \namemethod{} & 30.2 & 90.1 & 71.6 & 18.3 \\
        \specialrule{0.8pt}{1pt}{1pt}
        \end{tabular}
    }
    \vspace{-20pt} % 표 아래쪽 여백 조절
\end{wraptable}
used for the primary evaluations in the main paper. For faster experimentation, we generate 21-frame videos using 25-frame reference sequences.
\nametab{}~\ref{tab:supp_wan} shows that when applied to Wan, \namemethod{} produces videos that accurately reflect both the target prompt and the reference motion, achieving comparable motion fidelity to the results demonstrated in the main model. Qualitative motion transfer examples are shown in \namefig{}~\ref{fig:wan_qualitative}.

\begin{figure*}[t!]
    \centering
    \includegraphics[width=\textwidth]{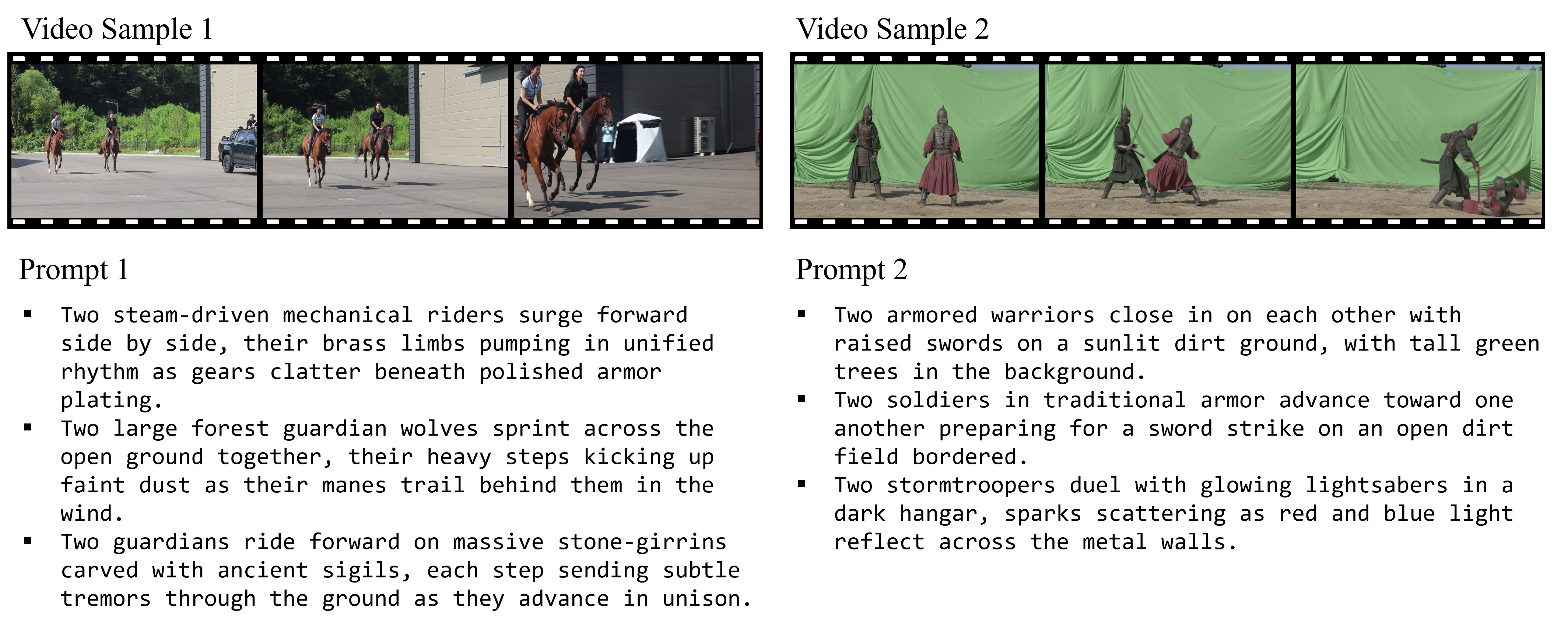}
    \caption{
    Overview of our movie scene dataset: video samples and corresponding prompts. 
    }
    \vspace{-8pt}
    \label{fig:dataset_overview}
\end{figure*}

\noindent\textbf{Cross-Architecture Robustness.}
While the main paper primarily reports results using CogVideoX~\cite{cogvideox}, \namemethod{} is not tied to this specific architecture.
To examine whether our entropy-based head selection depends on a particular backbone, we compare attention-entropy statistics between CogVideoX~\cite{cogvideox} and Wan~\cite{wan}.
We observe highly aligned timestep-wise entropy trends ($r{=}0.97$) and strong agreement in mean and median entropy ($0.92/0.96$), indicating that low-entropy structural heads emerge consistently across both video DiTs.
This supports entropy-based head selection as a backbone-robust criterion rather than architecture-specific tuning.

\section{Movie Scene Dataset}
\label{sec:supp-movie}
This section discusses practical applications of motion transfer in \namesec{}~\ref{sec:application}, provides the details of the movie-scene dataset in \namesec{}~\ref{sec:movie_detail}, and presents additional results of both our method and baseline models in \namesec{}~\ref{sec:movie_result}.

\subsection{Application Potential of Motion Transfer}
\label{sec:application}
Existing visual effects (VFX) and computer graphics (CG) productions involve time-consuming and labor-intensive pipelines, requiring specialized skills across stages, such as modeling, animation, and rendering~\cite{du2021diffpd}.
With recent advances in video generation accelerating content creation~\cite{ma2025controllable, li2025vfxmasterunlockingdynamicvisual}, motion transfer has emerged as a practical, controllable video generation with direct utility in filmmaking and game production.
The practical demand for such flexible content creation necessitates specialized datasets tailored to these domains.
To address this need, we introduce a movie scene dataset that serves as a foundational resource for advancing and benchmarking motion transfer in professional production contexts.
% To demonstrate its applicability in real-world scenarios, we additionally evaluate our method on the movie-scene dataset presented in the supplementary material.
% requiring only a reference video and a text prompt that specifies the desired motion and scenario.
% \input{table/supple/movie}

\subsection{Movie Scene Dataset Details}
\label{sec:movie_detail}
To evaluate our method in more realistic scenarios, we construct a movie scene dataset reflecting real-world video production environments, as shown in \namefig{}~\ref{fig:dataset_overview}.
While existing benchmarks~\cite{motionclone, det} already utilize real videos, we incorporate this dataset to further assess the generalizability and robustness of our approach in production-level settings.
The dataset consists of 20 videos, each paired with five different text prompts, totaling 100 samples, and the prompts describe both the primary objects and their backgrounds.
We generate prompts using GPT-4o~\cite{gpt4o} to cover diverse cinematic scenarios.
Moreover, the dataset involves complex motion patterns that go beyond simple movements, such as sword fighting, horseback riding, and walking in crowds.
Such variety provides a robust testbed for verifying the model's ability to faithfully transfer motion in dynamic real-world environments.

% In addition to object and human motion, the dataset also contains various types of camera movement, introducing additional challenges for motion transfer.

\subsection{Movie Scene Qualitative Results}
\label{sec:movie_result}
\begin{figure*}[t!]
    \centering
    \includegraphics[width=\textwidth]{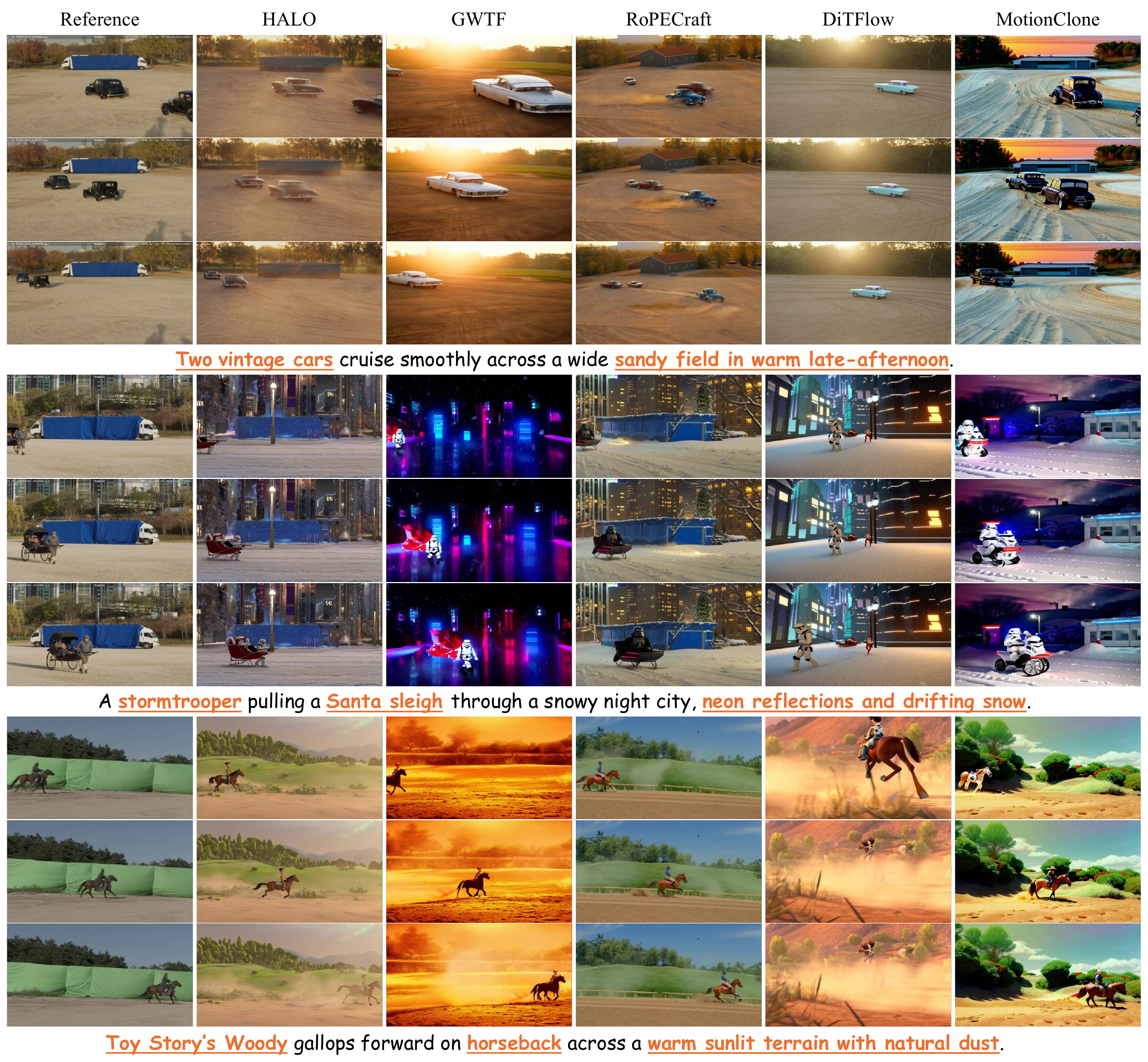}
    \caption{
    Qualitative comparison across diverse motion transfer tasks with the movie scene dataset.
    \namemethod{} is evaluated against U-Net- and DiT-based baselines across multiple subjects and motions.
    }
    % \vspace{-8pt}
    \label{fig:movie_baseline}
\end{figure*}

We present more qualitative comparisons on the movie scene dataset in \namefig{}~\ref{fig:movie_baseline}, evaluating multiple motion transfer methods. 
\namemethod{} achieves the most consistent motion transfer, exhibiting high motion fidelity and coherent structural alignment. 
ConMO \cite{conmo} is excluded from this comparison because it requires an input mask for generation, which is not available for our movie scene dataset.

\section{Additional Experiment Results}
\label{sec:supp-additional}

\subsection{More Qualitative Results}
\begin{figure*}[t!]
    \centering
    \includegraphics[width=\textwidth]{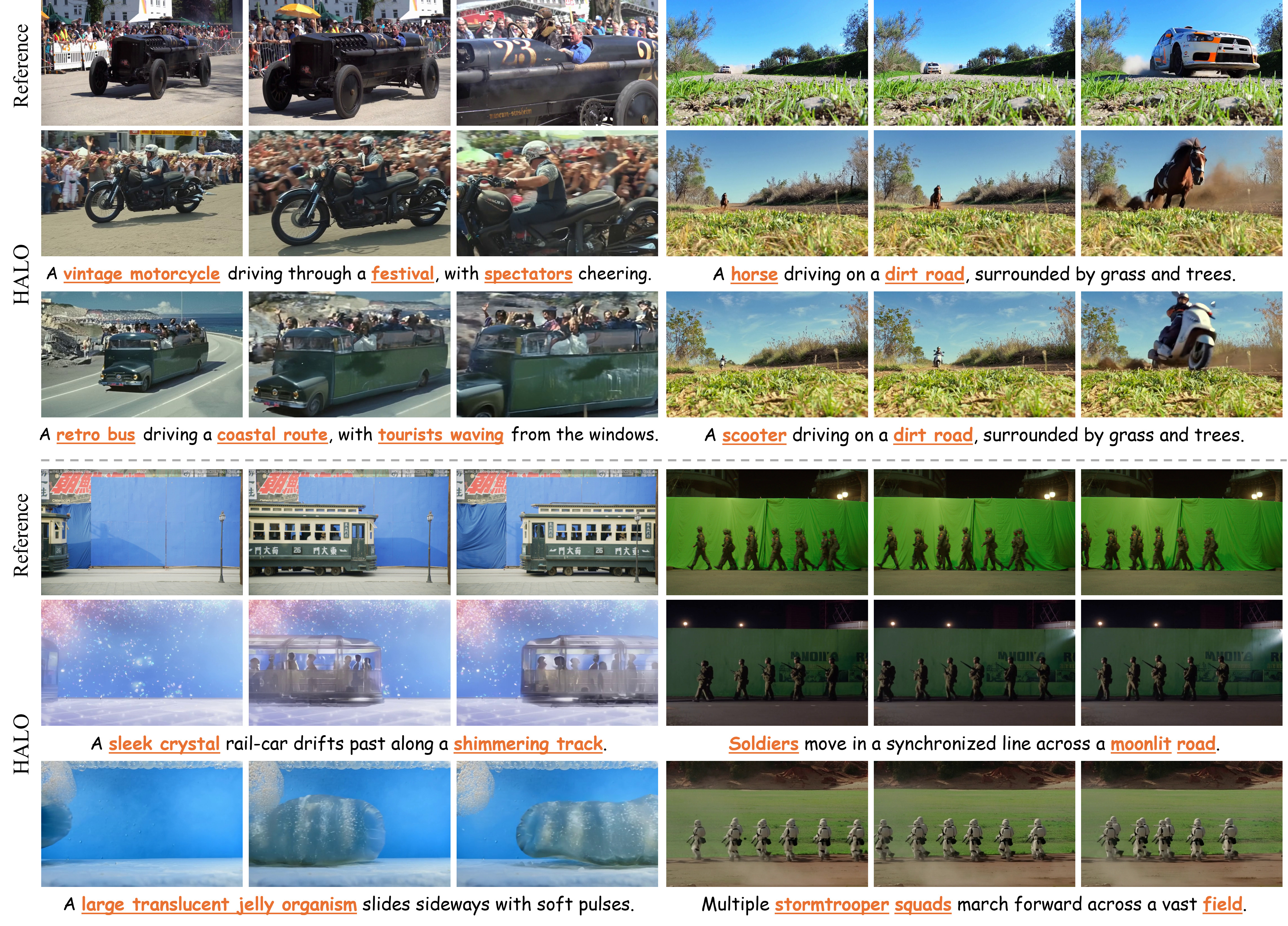}
    \caption{
    Qualitative results of \namemethod{} on benchmark and movie-scene datasets.
    The top rows illustrate performance on the motion transfer benchmark, while the bottom rows show generalization to movie scene videos.
    }
    % \vspace{-8pt}
    \label{fig:ours_quali}
\end{figure*}

\begin{figure*}[t!]
    \centering
    \includegraphics[width=\textwidth]{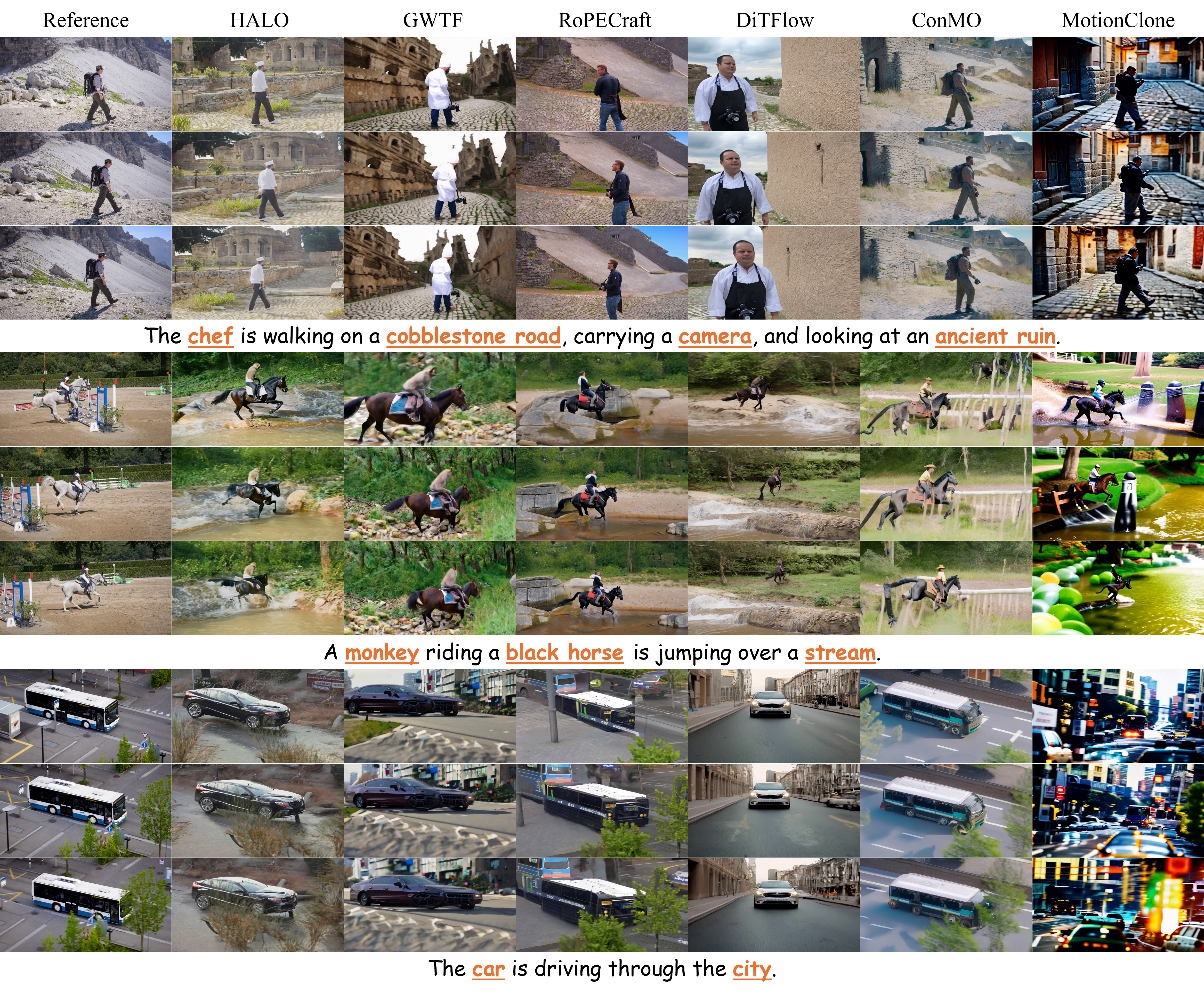}
    \caption{
    Qualitative comparison across diverse motion transfer tasks.
    \namemethod{} is evaluated against U-Net- and DiT-based baselines across multiple subjects and motions.
    }
    % \vspace{-8pt}
    \label{fig:baseline}
\end{figure*}

We provide additional qualitative results in \namefig{}~\ref{fig:ours_quali} across both the motion transfer benchmark (top) and our movie-scene dataset (bottom). 
Extended qualitative results in \namefig{}~\ref{fig:baseline} further show the robustness of \namemethod{} across varying motion scales, from routine activities (e.g., walking, bus motion) to challenging high-dynamic scenes (e.g., horse jumping). 
The results highlight our method's superior ability to maintain motion fidelity even as the complexity of the scene increases.
% See \texttt{video\_results.zip} for more video results. 

\clearpage

\subsection{Quantitative Trade-off and metric Analysis}
\begin{figure}[t!]
    \centering
    \vspace{-1pt}
    \includegraphics[width=0.65\columnwidth]{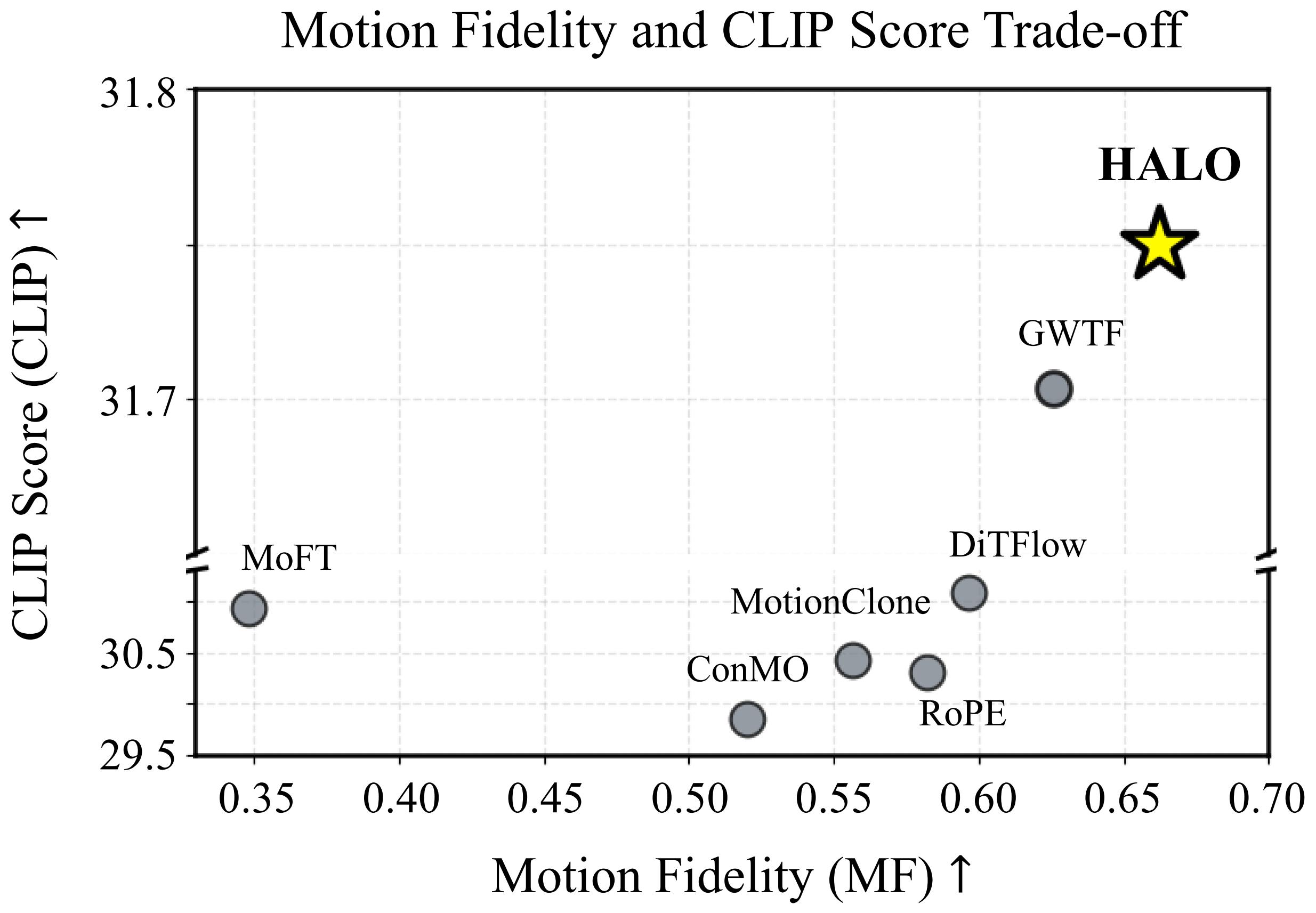}
    \vspace{-2pt}
    \caption{Objective comparison across motion-transfer methods in terms of Motion Fidelity (MF) and CLIP Score (CLIP). Baseline methods exhibit a trade-off between MF and CLIP, whereas \namemethod{} (Ours) achieves balanced performance.}
    \vspace{-7pt}
    \label{fig:quantitative_analysis}
\end{figure}

\textbf{CLIP vs. MF trade-off}
Given that motion transfer necessitates a delicate balance between preserving target semantics and following reference motion, CLIP and MF serve as complementary metrics that should be interpreted jointly.
While baseline methods exhibit a pronounced CLIP--MF trade-off--consistent with prior studies~\cite{det,smm}--\namemethod{} achieves a superior balance, attaining high scores in both metrics simultaneously (\namefig{}~\ref{fig:quantitative_analysis}).
These results suggest that our head-analysis approach effectively transfers motion while reducing irrelevant semantic interference.

\noindent\textbf{TC vs. Motion Dynamics analysis}
We further analyze temporal consistency by measuring the average optical-flow magnitude across generated videos to quantify motion dynamics.
As shown in \nametab{}~\ref{tab:supp_TC_tradeoff}, methods with higher automated TC scores can exhibit lower motion dynamics, suggesting that TC may favor relatively static outputs.
Therefore, TC should be interpreted jointly with reference-motion alignment metrics such as MF. Although \namemethod{} shows a slightly lower TC score than DiTFlow~\cite{ditflow} and GWTF~\cite{gowithflow}, it achieves substantially higher MF in the main quantitative results.
Moreover, the user study ranks \namemethod{} highest in perceived temporal consistency, indicating that the lower automated TC score does not necessarily imply weaker temporal coherence.

\begin{table}[h]
    \centering
    \caption{Temporal consistency and motion dynamics analysis.}
        \label{tab:supp_TC_tradeoff}
        \vspace{-5pt}
        \setlength{\tabcolsep}{5pt}
        \resizebox{0.65\linewidth}{!}
        {
        \begin{tabular}{lcccc}
            \specialrule{0.7pt}{0pt}{0pt}
            & \namemethod & GWTF & DiTFlow  \\
            \specialrule{0.5pt}{0pt}{0pt}
            Motion Dynamics $\uparrow$ & 5.79 & 4.81 & 5.19   \\
            Temporal Consistency $\uparrow$ & 87.5 & 88.2 & 89.5 \\ 
            \specialrule{0.7pt}{0pt}{0pt}
            % \rule{0pt}{2.5ex} % 행 높이 맞춤용 (표 높이가 다를 때 사용)
        \vspace{-20pt}
        \end{tabular}
        }
\end{table}

\subsection{Computational cost}

\begin{table}[h] % [t]: 페이지 최상단 배치
    \centering
    \vspace{-20pt}
    % --- 첫 번째 표 (왼쪽): Runtime 및 Memory ---
    \begin{minipage}[t]{0.46\textwidth}
        \centering
        \caption{Runtime and Memory Report.}
        \label{tab:supp_computational}
        \vspace{-5pt} % 캡션과 표 사이 간격
        \resizebox{\linewidth}{!}{%
            \begin{tabular}{lcccc}
                \specialrule{0.8pt}{1pt}{1pt}
                & Base  & +Sem. & +Inj. & \textit{HALO} \\
                \specialrule{1pt}{1pt}{1pt}
                Time(s) & 1076 & 1088 & 1115 & 1125 \\
                Memory(GB) & 38.1 & 38.2 & 38.2 & 38.2 \\
                \specialrule{0.8pt}{1pt}{1pt}
            \end{tabular}
        }
    \end{minipage}
    \hfill % 두 표 사이의 간격을 벌림
    % --- 두 번째 표 (오른쪽): Hyperparameter 등 다른 분석 ---
    \begin{minipage}[t]{0.5\textwidth}
        \centering
        \caption{Comparison with Baselines}
        \label{tab:supp_computational_baseline}
        \vspace{-5pt}
        \resizebox{\linewidth}{!}{%
            \begin{tabular}{lcccc}
                \specialrule{0.8pt}{1pt}{1pt}
                & DiTFlow & RoPECraft & GWTF & \namemethod{} \\
                \specialrule{1pt}{1pt}{1pt}
                Time(s)  &  1076  & 8187 & 456$+\alpha$ & 1125  \\
                Memory(GB)& 38.1 &  35.7 & 14.8$+\alpha$ & 38.2 \\
                % 15 &  -  & 76.4 & -  \\
                % 18 &  -  & 76.5 & -  \\
                \specialrule{0.8pt}{1pt}{1pt}
            \end{tabular}
        }
    \end{minipage}
\end{table}
% \begin{wraptable}{r}{0.5\textwidth} % r: 우측 배치, 0.5: 본문 너비의 절반 차지
%     \centering
%     \vspace{-35pt} % 표 위쪽 여백 조절
%     \caption{Runtime and Memory Report.}
%     \label{tab:supp_computational}
%     \vspace{3pt} % 캡션과 표 사이 간격 조절
%     \resizebox{\linewidth}{!}{%
%         \begin{tabular}{lccccc}
%         \specialrule{0.8pt}{1pt}{1pt}
%         & Base  & +Semantic & +Injection & \textit{HALO} \\
%         \specialrule{1pt}{1pt}{1pt}
%         Time(s) & 1076 & 1088 & 1115 & 1125 \\
%         Memory(GB) & 38.1 & 38.2 & 38.2 & 38.2 \\
%         \specialrule{0.8pt}{1pt}{1pt}
%         \end{tabular}
%     }
%     \vspace{-20pt} % 표 아래쪽 여백 조절
% \end{wraptable}

We report the runtime and memory efficiency of \textit{HALO} on a single NVIDIA RTX A6000.
As shown in \nametab{}~\ref{tab:supp_computational}, incorporating semantic correspondence and head-wise injection incurs only a minor runtime overhead; compared to the base model, \textit{HALO} adds just $49s$ in inference time and $0.1$GB in peak memory.
We further compare \namemethod{}'s computational cost against other DiT-based methods in \nametab{}~\ref{tab:supp_computational_baseline}. 
A direct comparison with GWTF is challenging because its training requirements introduce significant additional overhead $\alpha$ beyond simple inference.
\namemethod{} demonstrates computational parity with leading training-free methods. 
Specifically, our approach remains highly competitive with DiTFlow and RopeCraft, achieving faithful motion transfer without sacrificing practical inference scalability.

\begin{figure*}[t!]
    \centering
    \includegraphics[width=\textwidth]{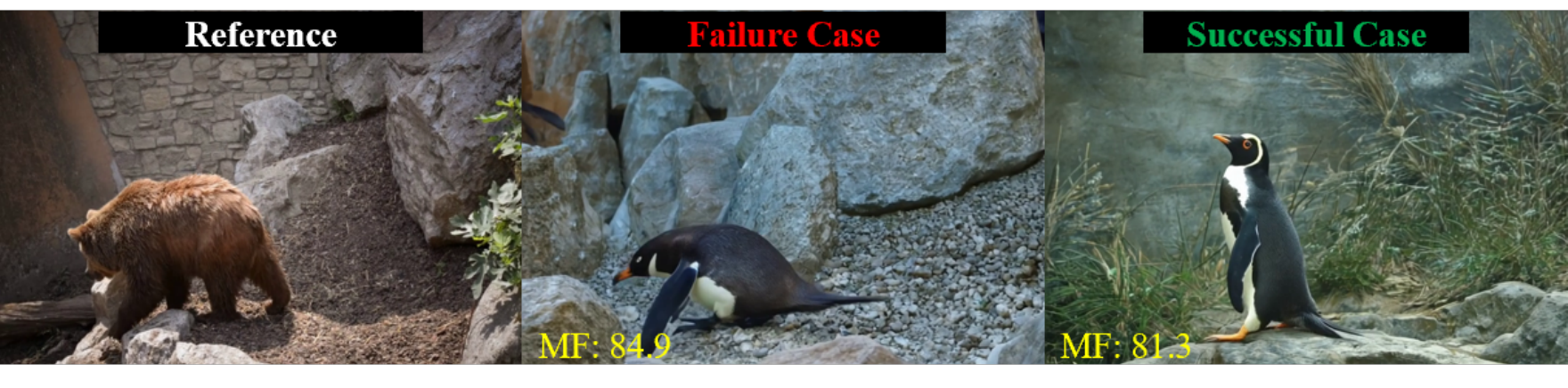}
    \caption{Failure cases of our method. In these examples, the output closely follows the reference motion but exhibits unnatural articulated movements (e.g., a penguin walking like a bear), which occurs when semantic correspondence over-aligns structurally incompatible articulations.}
    % \vspace{-8pt}
    \label{fig:supp_failure}
\end{figure*}

\subsection{Limitations}
In some cases, the generated motion closely follows the reference yet becomes unnatural (e.g., a penguin walking like a bear), as illustrated in \namefig{}~\ref{fig:supp_failure}. This failure arises when semantic correspondence over-aligns structurally incompatible articulations. Because our semantic correspondence module refines motion representations via fine-grained correspondences, it can occasionally enforce near one-to-one part alignment, leading to implausible kinematics. Interestingly, these cases can exhibit higher motion fidelity scores because they more strictly track the reference dynamics. 
This reveals an inherent trade-off between reference-motion following and kinematic plausibility, which is not fully captured by existing evaluations. 

\begin{figure*}[t]
    \centering
    \includegraphics[width=0.7\textwidth]{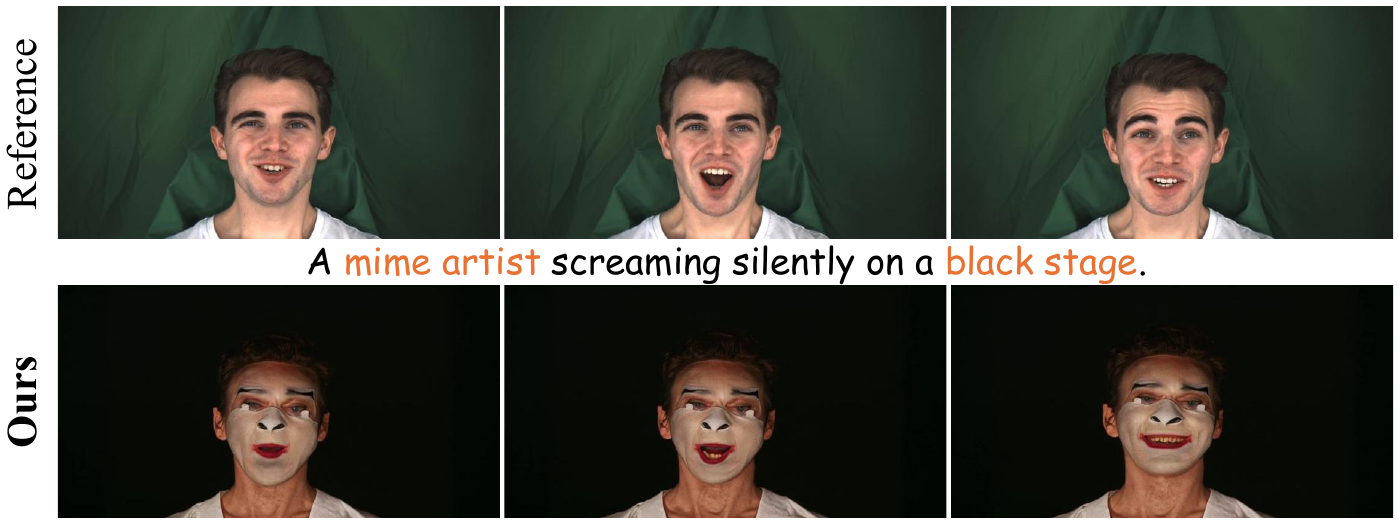}
    \caption{Fine-grained local motion transfer.
    \namemethod{} preserves coarse facial motion and head pose, but subtle non-rigid deformations, such as detailed facial expressions, remain challenging due to the patch-level displacement representation.}
    % \vspace{-8pt}
    \label{fig:supp_facial}
\end{figure*}

\vspace{1mm}
\noindent\textbf{Fine-Grained Local Motion}
We further analyze \namemethod{} under fine-grained local motion, such as facial expressions.
As shown in \namefig{}~\ref{fig:supp_facial}, \namemethod{} preserves the overall head pose and coarse facial motion of the reference, showing comparable motion alignment to existing methods.
However, subtle non-rigid deformations, such as detailed mouth shapes and fine facial expressions, remain challenging.
This limitation arises because \namemethod{} represents motion using patch-level displacements, which are effective for object-level and medium-scale motion but less precise for very local deformation.

\subsection{Evaluation of Structure-Specialized Heads for Video Editing}
\begin{figure*}[t!]
    \centering
    \includegraphics[width=\textwidth]{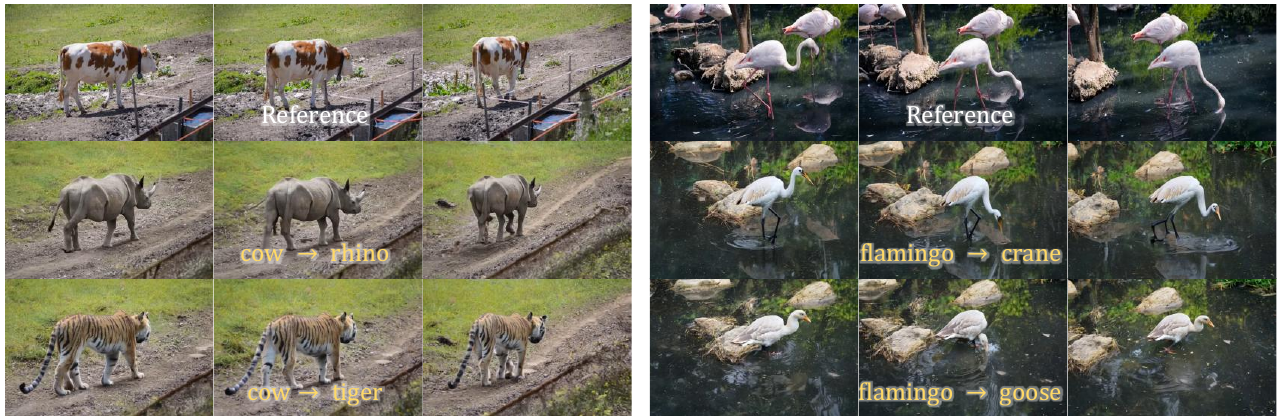}
    \caption{Application to video editing using the structure-specialized heads.}
    % \vspace{-8pt}
    \label{fig:supp_edit}
\end{figure*}

As shown in the main paper, we further extend our method to video editing. For quantitative evaluation, we report CLIP score (CLIP), Temporal Consistency (TC), Motion Fidelity (MF), masked PSNR (m.P), and masked LPIPS (m.L)~\cite{wang2025videodirector}. To compute m.P and m.L, we use the provided segmentation masks and evaluate reconstruction quality on the background region, which is expected to remain unchanged from the reference. As shown in \nametab{}~\ref{tab:supp_videoedit}, \namemethod{} achieves strong editing quality while preserving background structure, supporting the effectiveness of structure-specialized heads beyond motion transfer. Additional qualitative results are provided in \namefig{}~\ref{fig:supp_edit}.

\begin{table}[h]
    \centering
    \caption{Video editing results.}
        \label{tab:supp_videoedit}
        \vspace{-5pt}
        \setlength{\tabcolsep}{5pt}
        \resizebox{0.65\linewidth}{!}
        {
        \begin{tabular}{cccccc}
            \specialrule{0.8pt}{1pt}{1pt}
              & CLIP $\uparrow$ & TC $\uparrow$ & MF $\uparrow$ & m.P $\uparrow$ & m.L $\downarrow$\\
            \specialrule{1pt}{1pt}{1pt}
            \namemethod &  29.3 & 93.1 & 81.9 & 17.2 & 0.4 \\
            \specialrule{0.8pt}{1pt}{1pt}
            % \rule{0pt}{2.5ex} % 행 높이 맞춤용 (표 높이가 다를 때 사용)
        \vspace{-20pt}
        \end{tabular}
        }

\end{table}

\subsection{Structural Consistency in Long Videos}
\begin{figure*}[t]
    \centering
    \includegraphics[width=\textwidth]{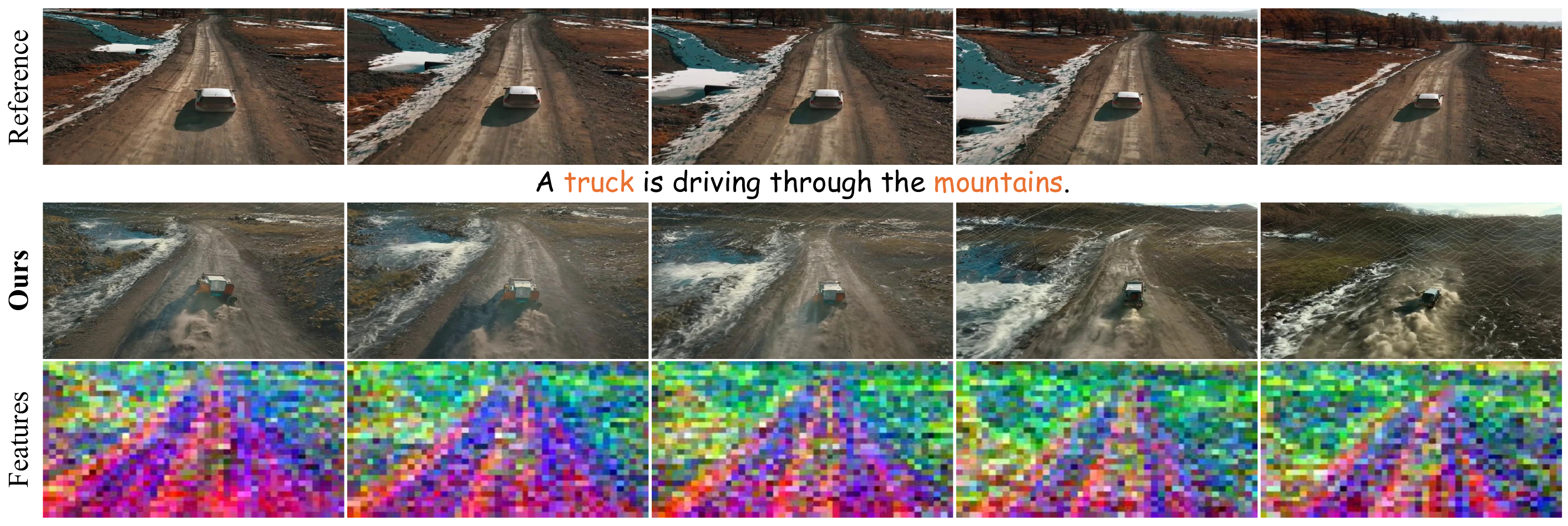}
    \caption{Structural consistency in long-video generation.
    We evaluate \namemethod{} on $112$-frame sequences and visualize the behavior of selected low-entropy heads.
    The selected heads maintain stable structural attention patterns over extended temporal horizons, supporting long-sequence structural preservation.}
    % \vspace{-8pt}
    \label{fig:supp_long_video}
\end{figure*}

To validate structural consistency in long videos, we evaluate \namemethod{} on $112$-frame sequences.
This setting is substantially longer than the standard generation length used in the main experiments, allowing us to examine whether the selected structure-specialized heads remain stable over extended temporal horizons.
As shown in \namefig{}~\ref{fig:supp_long_video}, low-entropy heads preserve consistent structural attention patterns across the sequence, without noticeable drift or degradation over time.
This indicates that entropy-based head selection is not limited to short clips but provides stable structural cues for long-horizon motion transfer.

\subsection{Scalability of SCR}
SCR operates entirely in the latent space, making its computational overhead moderate even at higher output resolutions.
As shown in \nametab{}~\ref{tab:supp_scalability}, increasing the resolution from $720$p to $1080$p increases the SCR runtime only from $13.2$s to $19.1$s, while the total inference time increases from $1125$s to $1404$s.
The memory usage of SCR also remains nearly unchanged across the two resolutions.
These results indicate that SCR does not introduce a major bottleneck and remains practical for high-resolution motion transfer.

\begin{table}[h]
    \centering
    \caption{Scalability analysis of SCR at higher resolutions.}
        \label{tab:supp_scalability}
        \vspace{-5pt}
        \setlength{\tabcolsep}{5pt}
        \resizebox{0.65\linewidth}{!}
        {
        \begin{tabular}{lccccc}
            \specialrule{0.7pt}{0pt}{0pt}
            & \namemethod{} 720p & SCR & \namemethod{} 1080p & SCR \\
            \specialrule{0.5pt}{0pt}{0pt}
            Time (s) & 1125 & 13.2 & 1404 & 19.1 \\
            Mem. (GB) & 38.2 & 18.2 & 38.2 & 18.4 \\
            \specialrule{0.7pt}{0pt}{0pt}
            % \rule{0pt}{2.5ex} % 행 높이 맞춤용 (표 높이가 다를 때 사용)
        \vspace{-20pt}
        \end{tabular}
        }
\end{table}

% \clearpage

% \clearpage

\section{User Study}
\label{sec:supp-discussion}
For human evaluation, we conduct a comprehensive user study involving 20 participants with expertise in computer vision. 
The evaluation is performed on a set of 10 representative video samples. 
Participants are asked to assess the results based on three key criteria: editing accuracy, temporal consistency, and motion fidelity.
The specific evaluation metrics are defined as follows:

\begin{itemize}
    \item \textbf{Edit Accuracy}: How well does the video content align with the provided text prompt? (e.g., whether the characters, actions, and attributes described in the prompt are accurately reflected in the visual output).

    \item \textbf{Temporal Consistency}: Do the motion, appearance, and background elements transition naturally throughout the video? (e.g., checking for the absence of flickers or abrupt changes in character identity or background details).

    \item \textbf{Motion Fidelity}: How closely does the overall motion in the generated video resemble the reference motion? (e.g., whether the pose, flow of action, speed, and timing align with the source motion).
\end{itemize}

\clearpage
\bibliographystyle{splncs04}
\bibliography{main}

@String(CVPR  = {IEEE Conf. Comput. Vis. Pattern Recog.})

@String(ICCV  = {Int. Conf. Comput. Vis.})

@String(ECCV  = {Eur. Conf. Comput. Vis.})

@String(NeurIPS = {Adv. Neural Inform. Process. Syst.})

@String(ICML  = {Int. Conf. Mach. Learn.})

@String(ICLR  = {Int. Conf. Learn. Represent.})

@String(AAAI  = {AAAI})

@String(TOG   = {ACM Trans. Graph.})

@String(CVPR  = {CVPR})

@String(ICCV  = {ICCV})

@String(ECCV  = {ECCV})

@String(NeurIPS = {NeurIPS})

@String(ICML  = {ICML})

@String(ICLR  = {ICLR})

@String(TOG   = {ACM TOG})

@article{ropecraft,
  title={RoPECraft: Training-Free Motion Transfer with Trajectory-Guided RoPE Optimization on Diffusion Transformers},
  author={Gokmen, Ahmet Berke and Ekin, Yigit and Bilecen, Bahri Batuhan and Dundar, Aysegul},
  journal={arXiv preprint arXiv:2505.13344},
  year={2025}
}

@inproceedings{ditflow,
  title={Video motion transfer with diffusion transformers},
  author={Pondaven, Alexander and Siarohin, Aliaksandr and Tulyakov, Sergey and Torr, Philip and Pizzati, Fabio},
  booktitle=cvpr,
  pages={22911--22921},
  year={2025}
}

@inproceedings{det,
  title={Decouple and track: Benchmarking and improving video diffusion transformers for motion transfer},
  author={Shi, Qingyu and Wu, Jianzong and Bai, Jinbin and Zhang, Jiangning and Qi, Lu and Tong, Yunhai and Li, Xiangtai},
  booktitle=iccv,
  pages={10995--11005},
  year={2025}
}

@inproceedings{gowithflow,
  title={Go-with-the-flow: Motion-controllable video diffusion models using real-time warped noise},
  author={Burgert, Ryan and Xu, Yuancheng and Xian, Wenqi and Pilarski, Oliver and Clausen, Pascal and He, Mingming and Ma, Li and Deng, Yitong and Li, Lingxiao and Mousavi, Mohsen and others},
  booktitle=cvpr,
  pages={13--23},
  year={2025}
}

@inproceedings{
cogvideox,
title={CogVideoX: Text-to-Video Diffusion Models with An Expert Transformer},
author={Zhuoyi Yang and Jiayan Teng and Wendi Zheng and Ming Ding and Shiyu Huang and Jiazheng Xu and Yuanming Yang and Wenyi Hong and Xiaohan Zhang and Guanyu Feng and Da Yin and Yuxuan.Zhang and Weihan Wang and Yean Cheng and Bin Xu and Xiaotao Gu and Yuxiao Dong and Jie Tang},
booktitle=iclr,
year={2025},
url={https://openreview.net/forum?id=LQzN6TRFg9}
}

@article{wan,
      title={Wan: Open and Advanced Large-Scale Video Generative Models}, 
      author={Team Wan and Ang Wang and Baole Ai and Bin Wen and Chaojie Mao and Chen-Wei Xie and Di Chen and Feiwu Yu and Haiming Zhao and Jianxiao Yang and Jianyuan Zeng and Jiayu Wang and Jingfeng Zhang and Jingren Zhou and Jinkai Wang and Jixuan Chen and Kai Zhu and Kang Zhao and Keyu Yan and Lianghua Huang and Mengyang Feng and Ningyi Zhang and Pandeng Li and Pingyu Wu and Ruihang Chu and Ruili Feng and Shiwei Zhang and Siyang Sun and Tao Fang and Tianxing Wang and Tianyi Gui and Tingyu Weng and Tong Shen and Wei Lin and Wei Wang and Wei Wang and Wenmeng Zhou and Wente Wang and Wenting Shen and Wenyuan Yu and Xianzhong Shi and Xiaoming Huang and Xin Xu and Yan Kou and Yangyu Lv and Yifei Li and Yijing Liu and Yiming Wang and Yingya Zhang and Yitong Huang and Yong Li and You Wu and Yu Liu and Yulin Pan and Yun Zheng and Yuntao Hong and Yupeng Shi and Yutong Feng and Zeyinzi Jiang and Zhen Han and Zhi-Fan Wu and Ziyu Liu},
      journal = {arXiv preprint arXiv:2503.20314},
      year={2025}
}

@inproceedings{videocrafter2,
  title={Videocrafter2: Overcoming data limitations for high-quality video diffusion models},
  author={Chen, Haoxin and Zhang, Yong and Cun, Xiaodong and Xia, Menghan and Wang, Xintao and Weng, Chao and Shan, Ying},
  booktitle=cvpr,
  pages={7310--7320},
  year={2024}
}

@article{sora,
  title={Sora: A review on background, technology, limitations, and opportunities of large vision models},
  author={Liu, Yixin and Zhang, Kai and Li, Yuan and Yan, Zhiling and Gao, Chujie and Chen, Ruoxi and Yuan, Zhengqing and Huang, Yue and Sun, Hanchi and Gao, Jianfeng and others},
  journal={arXiv preprint arXiv:2402.17177},
  year={2024}
}

@inproceedings{zeroscope,
  title={Text2video-zero: Text-to-image diffusion models are zero-shot video generators},
  author={Khachatryan, Levon and Movsisyan, Andranik and Tadevosyan, Vahram and Henschel, Roberto and Wang, Zhangyang and Navasardyan, Shant and Shi, Humphrey},
  booktitle=iccv,
  pages={15954--15964},
  year={2023}
}

@inproceedings{lumiere,
  title={Lumiere: A space-time diffusion model for video generation},
  author={Bar-Tal, Omer and Chefer, Hila and Tov, Omer and Herrmann, Charles and Paiss, Roni and Zada, Shiran and Ephrat, Ariel and Hur, Junhwa and Liu, Guanghui and Raj, Amit and others},
  booktitle={SIGGRAPH Asia 2024 Conference Papers},
  pages={1--11},
  year={2024}
}

@inproceedings{
animatediff,
title={AnimateDiff: Animate Your Personalized Text-to-Image Diffusion Models without Specific Tuning},
author={Yuwei Guo and Ceyuan Yang and Anyi Rao and Zhengyang Liang and Yaohui Wang and Yu Qiao and Maneesh Agrawala and Dahua Lin and Bo Dai},
booktitle=iclr,
year={2024},
url={https://openreview.net/forum?id=Fx2SbBgcte}
}

@article{svd,
  title={Stable video diffusion: Scaling latent video diffusion models to large datasets},
  author={Blattmann, Andreas and Dockhorn, Tim and Kulal, Sumith and Mendelevitch, Daniel and Kilian, Maciej and Lorenz, Dominik and Levi, Yam and English, Zion and Voleti, Vikram and Letts, Adam and others},
  journal={arXiv preprint arXiv:2311.15127},
  year={2023}
}

@article{ltx,
  title={Ltx-video: Realtime video latent diffusion},
  author={HaCohen, Yoav and Chiprut, Nisan and Brazowski, Benny and Shalem, Daniel and Moshe, Dudu and Richardson, Eitan and Levin, Eran and Shiran, Guy and Zabari, Nir and Gordon, Ori and others},
  journal={arXiv preprint arXiv:2501.00103},
  year={2024}
}

@article{hunyuanvideo,
  title={Hunyuanvideo: A systematic framework for large video
generative models},
  author={Kong, Weijie and Tian, Qi and Zhang, Zijian and Min, Rox and Dai, Zuozhuo and Zhou, Jin and Xiong, Jiangfeng and Li, Xin and Wu, Bo and Zhang, Jianwei and others},
  journal={arXiv preprint arXiv:2412.03603},
  year={2024}
}

@article{moft,
  title={Video diffusion models are training-free motion interpreter and controller},
  author={Xiao, Zeqi and Zhou, Yifan and Yang, Shuai and Pan, Xingang},
  journal=neurips,
  volume={37},
  pages={76115--76138},
  year={2024}
}

@inproceedings{vmc,
        title={VMC: Video Motion Customization using Temporal Attention Adaption for Text-to-Video Diffusion Models}, 
        author={Jeong, Hyeonho and Park, Geon Yeong and Ye, Jong Chul},
    booktitle=cvpr,
        year={2023},
}

@article{motionclone,
  title={Motionclone: Training-free motion cloning for controllable video generation},
  author={Ling, Pengyang and Bu, Jiazi and Zhang, Pan and Dong, Xiaoyi and Zang, Yuhang and Wu, Tong and Chen, Huaian and Wang, Jiaqi and Jin, Yi},
  journal={arXiv preprint arXiv:2406.05338},
  year={2024}
}

@inproceedings{motioninv,
  title={Motion inversion for video customization},
  author={Wang, Luozhou and Mai, Ziyang and Shen, Guibao and Liang, Yixun and Tao, Xin and Wan, Pengfei and Zhang, Di and Li, Yijun and Chen, Ying-Cong},
  booktitle={Proceedings of the Special Interest Group on Computer Graphics and Interactive Techniques Conference Conference Papers},
  pages={1--12},
  year={2025}
}

@inproceedings{motiondirector,
  title={Motiondirector: Motion customization of text-to-video diffusion models},
  author={Zhao, Rui and Gu, Yuchao and Wu, Jay Zhangjie and Zhang, David Junhao and Liu, Jia-Wei and Wu, Weijia and Keppo, Jussi and Shou, Mike Zheng},
  booktitle=eccv,
  pages={273--290},
  year={2024},
  organization={Springer}
}

@article{understanding,
  title={Understanding attention mechanism in video diffusion models},
  author={Liu, Bingyan and Wang, Chengyu and Su, Tongtong and Ten, Huan and Huang, Jun and Guo, Kailing and Jia, Kui},
  journal={arXiv preprint arXiv:2504.12027},
  year={2025}
}

@inproceedings{transfer1,
  title={Animating arbitrary objects via deep motion transfer},
  author={Siarohin, Aliaksandr and Lathuili{\`e}re, St{\'e}phane and Tulyakov, Sergey and Ricci, Elisa and Sebe, Nicu},
  booktitle=cvpr,
  pages={2377--2386},
  year={2019}
}

@inproceedings{smm,
  title={Space-time diffusion features for zero-shot text-driven motion transfer},
  author={Yatim, Danah and Fridman, Rafail and Bar-Tal, Omer and Kasten, Yoni and Dekel, Tali},
  booktitle=cvpr,
  pages={8466--8476},
  year={2024}
}

@article{difftrack,
  title={Emergent Temporal Correspondences from Video Diffusion Transformers},
  author={Nam, Jisu and Son, Soowon and Chung, Dahyun and Kim, Jiyoung and Jin, Siyoon and Hur, Junhwa and Kim, Seungryong},
  journal={arXiv preprint arXiv:2506.17220},
  year={2025}
}

@article{sparsevideogen,
  title={Sparse videogen: Accelerating video diffusion transformers with spatial-temporal sparsity},
  author={Xi, Haocheng and Yang, Shuo and Zhao, Yilong and Xu, Chenfeng and Li, Muyang and Li, Xiuyu and Lin, Yujun and Cai, Han and Zhang, Jintao and Li, Dacheng and others},
  journal={arXiv preprint arXiv:2502.01776},
  year={2025}
}

@article{dift,
  title={Emergent correspondence from image diffusion},
  author={Tang, Luming and Jia, Menglin and Wang, Qianqian and Phoo, Cheng Perng and Hariharan, Bharath},
  journal=neurips,
  volume={36},
  pages={1363--1389},
  year={2023}
}

@inproceedings{raft,
  title={Raft: Recurrent all-pairs field transforms for optical flow},
  author={Teed, Zachary and Deng, Jia},
  booktitle=eccv,
  pages={402--419},
  year={2020},
  organization={Springer}
}

@inproceedings{conmo,
  title={Conmo: Controllable motion disentanglement and recomposition for zero-shot motion transfer},
  author={Gao, Jiayi and Yin, Zijin and Hua, Changcheng and Peng, Yuxin and Liang, Kongming and Ma, Zhanyu and Guo, Jun and Liu, Yang},
  booktitle=cvpr,
  pages={7191--7200},
  year={2025}
}

@article{cove,
  title={Cove: Unleashing the diffusion feature correspondence for consistent video editing},
  author={Wang, Jiangshan and Ma, Yue and Guo, Jiayi and Xiao, Yicheng and Huang, Gao and Li, Xiu},
  journal=neurips,
  volume={37},
  pages={96541--96565},
  year={2024}
}

@article{motionbyqueries,
  title={Motion by Queries: Identity-Motion Trade-offs in Text-to-Video Generation},
  author={Atzmon, Yuval and Gal, Rinon and Tewel, Yoad and Kasten, Yoni and Chechik, Gal},
  journal={arXiv preprint arXiv:2412.07750},
  year={2024}
}

@inproceedings{ditctrl,
  title={Ditctrl: Exploring attention control in multi-modal diffusion transformer for tuning-free multi-prompt longer video generation},
  author={Cai, Minghong and Cun, Xiaodong and Li, Xiaoyu and Liu, Wenze and Zhang, Zhaoyang and Zhang, Yong and Shan, Ying and Yue, Xiangyu},
  booktitle=cvpr,
  pages={7763--7772},
  year={2025}
}

@inproceedings{stableflow,
  title={Stable flow: Vital layers for training-free image editing},
  author={Avrahami, Omri and Patashnik, Or and Fried, Ohad and Nemchinov, Egor and Aberman, Kfir and Lischinski, Dani and Cohen-Or, Daniel},
  booktitle=cvpr,
  pages={7877--7888},
  year={2025}
}

@article{unsupervised,
  title={Unsupervised semantic correspondence using stable diffusion},
  author={Hedlin, Eric and Sharma, Gopal and Mahajan, Shweta and Isack, Hossam and Kar, Abhishek and Tagliasacchi, Andrea and Yi, Kwang Moo},
  journal=neurips,
  volume={36},
  pages={8266--8279},
  year={2023}
}

@article{spatialentropy,
  title={Spatial entropy},
  author={Batty, Michael},
  journal={Geographical analysis},
  volume={6},
  number={1},
  pages={1--31},
  year={1974},
  publisher={Wiley Online Library}
}

@article{spatialentropy2,
  title={Spatial entropy as an inductive bias for vision transformers},
  author={Peruzzo, Elia and Sangineto, Enver and Liu, Yahui and De Nadai, Marco and Bi, Wei and Lepri, Bruno and Sebe, Nicu},
  journal={Machine Learning},
  volume={113},
  number={9},
  pages={6945--6975},
  year={2024},
  publisher={Springer}
}

@inproceedings{Kang,
  title={Your large vision-language model only needs a few attention heads for visual grounding},
  author={Kang, Seil and Kim, Jinyeong and Kim, Junhyeok and Hwang, Seong Jae},
  booktitle=cvpr,
  pages={9339--9350},
  year={2025}
}

@article{attentropy,
  title={Active visual exploration based on attention-map entropy},
  author={Pardyl, Adam and Rype{\'s}{\'c}, Grzegorz and Kurzejamski, Grzegorz and Zieli{\'n}ski, Bartosz and Trzci{\'n}ski, Tomasz},
  journal={arXiv preprint arXiv:2303.06457},
  year={2023}
}

@inproceedings{scaling,
  title={Scaling rectified flow transformers for high-resolution image synthesis},
  author={Esser, Patrick and Kulal, Sumith and Blattmann, Andreas and Entezari, Rahim and M{\"u}ller, Jonas and Saini, Harry and Levi, Yam and Lorenz, Dominik and Sauer, Axel and Boesel, Frederic and others},
  booktitle=icml,
  year={2024}
}

@inproceedings{sma,
  title={Spectral motion alignment for video motion transfer using diffusion models},
  author={Park, Geon Yeong and Jeong, Hyeonho and Lee, Sang Wan and Ye, Jong Chul},
  booktitle={Proceedings of the AAAI Conference on Artificial Intelligence},
  volume={39},
  number={6},
  pages={6398--6405},
  year={2025}
}

@inproceedings{clipscore,
  title={Clipscore: A reference-free evaluation metric for image captioning},
  author={Hessel, Jack and Holtzman, Ari and Forbes, Maxwell and Le Bras, Ronan and Choi, Yejin},
  booktitle={Proceedings of the 2021 conference on empirical methods in natural language processing},
  pages={7514--7528},
  year={2021}
}

@inproceedings{dino,
  title={DINO: DETR with Improved DeNoising Anchor Boxes for End-to-End Object Detection},
  author={Zhang, Hao and Li, Feng and Liu, Shilong and Zhang, Lei and Su, Hang and Zhu, Jun and Ni, Lionel and Shum, Heung-Yeung},
  booktitle={The Eleventh International Conference on Learning Representations},
  year={2023}

}

@inproceedings{gentron,
  title={Gentron: Diffusion transformers for image and video generation},
  author={Chen, Shoufa and Xu, Mengmeng and Ren, Jiawei and Cong, Yuren and He, Sen and Xie, Yanping and Sinha, Animesh and Luo, Ping and Xiang, Tao and Perez-Rua, Juan-Manuel},
  booktitle={Proceedings of the IEEE/CVF Conference on Computer Vision and Pattern Recognition},
  pages={6441--6451},
  year={2024}
}

@article{xing2024make,
  title={Make-your-video: Customized video generation using textual and structural guidance},
  author={Xing, Jinbo and Xia, Menghan and Liu, Yuxin and Zhang, Yuechen and Zhang, Yong and He, Yingqing and Liu, Hanyuan and Chen, Haoxin and Cun, Xiaodong and Wang, Xintao and others},
  journal={IEEE Transactions on Visualization and Computer Graphics},
  volume={31},
  number={2},
  pages={1526--1541},
  year={2024},
  publisher={IEEE}
}

@article{du2021diffpd,
  title={Diffpd: Differentiable projective dynamics},
  author={Du, Tao and Wu, Kui and Ma, Pingchuan and Wah, Sebastien and Spielberg, Andrew and Rus, Daniela and Matusik, Wojciech},
  journal={ACM Transactions on Graphics (ToG)},
  volume={41},
  number={2},
  pages={1--21},
  year={2021},
  publisher={ACM New York, NY}
}

@article{ma2025controllable,
  title={Controllable video generation: A survey},
  author={Ma, Yue and Feng, Kunyu and Hu, Zhongyuan and Wang, Xinyu and Wang, Yucheng and Zheng, Mingzhe and He, Xuanhua and Zhu, Chenyang and Liu, Hongyu and He, Yingqing and others},
  journal={arXiv preprint arXiv:2507.16869},
  year={2025}
}

@misc{li2025vfxmasterunlockingdynamicvisual,
      title={VFXMaster: Unlocking Dynamic Visual Effect Generation via In-Context Learning}, 
      author={Baolu Li and Yiming Zhang and Qinghe Wang and Liqian Ma and Xiaoyu Shi and Xintao Wang and Pengfei Wan and Zhenfei Yin and Yunzhi Zhuge and Huchuan Lu and Xu Jia},
      year={2025},
      eprint={2510.25772},
      archivePrefix={arXiv},
      primaryClass={cs.CV},
      url={https://arxiv.org/abs/2510.25772}, 
}

@InProceedings{vbench,
    title={{VBench}: Comprehensive Benchmark Suite for Video Generative Models},
    author={Huang, Ziqi and He, Yinan and Yu, Jiashuo and Zhang, Fan and Si, Chenyang and Jiang, Yuming and Zhang, Yuanhan and Wu, Tianxing and Jin, Qingyang and Chanpaisit, Nattapol and Wang, Yaohui and Chen, Xinyuan and Wang, Limin and Lin, Dahua and Qiao, Yu and Liu, Ziwei},
    booktitle={Proceedings of the IEEE/CVF Conference on Computer Vision and Pattern Recognition},
    year={2024}
}

@article{kim2025seg4diff,
  title={Seg4diff: Unveiling open-vocabulary segmentation in text-to-image diffusion transformers},
  author={Kim, Chaehyun and Shin, Heeseong and Hong, Eunbeen and Yoon, Heeji and Arnab, Anurag and Seo, Paul Hongsuck and Hong, Sunghwan and Kim, Seungryong},
  journal={arXiv preprint arXiv:2509.18096},
  year={2025}
}

@article{ahn2025fine,
  title={Fine-grained perturbation guidance via attention head selection},
  author={Ahn, Donghoon and Kang, Jiwon and Lee, Sanghyun and Kim, Minjae and Min, Jaewon and Jang, Wooseok and Lee, Saungwu and Paul, Sayak and Hong, Susung and Kim, Seungryong},
  journal={arXiv e-prints},
  pages={arXiv--2506},
  year={2025}
}

@inproceedings{wang2025videodirector,
  title={Videodirector: Precise video editing via text-to-video models},
  author={Wang, Yukun and Wang, Longguang and Ma, Zhiyuan and Hu, Qibin and Xu, Kai and Guo, Yulan},
  booktitle={Proceedings of the IEEE/CVF Conference on Computer Vision and Pattern Recognition},
  pages={2589--2598},
  year={2025}
}

@article{gpt4o,
  title={Gpt-4o system card},
  author={Hurst, Aaron and Lerer, Adam and Goucher, Adam P and Perelman, Adam and Ramesh, Aditya and Clark, Aidan and Ostrow, AJ and Welihinda, Akila and Hayes, Alan and Radford, Alec and others},
  journal={arXiv preprint arXiv:2410.21276},
  year={2024}
}

\end{document}